\documentclass[twoside]{article}

\usepackage[accepted]{aistats2024}
\usepackage[sort]{natbib} 
\usepackage{enumitem}
\usepackage{amsmath,amsfonts,bm}

\def\1{\bm{1}}

\DeclareMathAlphabet{\mathsfit}{\encodingdefault}{\sfdefault}{m}{sl}
\SetMathAlphabet{\mathsfit}{bold}{\encodingdefault}{\sfdefault}{bx}{n}

\DeclareMathOperator*{\argmax}{arg\,max}
\DeclareMathOperator*{\argmin}{arg\,min}

\usepackage{amsthm}
\newtheorem{thm}{Theorem}[section]

\newtheorem{rmk}[thm]{Remark}

\newcommand{\mbf}[1]{\mathbf{#1}}

\usepackage{mathtools}

\usepackage{bbm} 
\usepackage{bbold} 

\usepackage{mdframed}
\newmdtheoremenv[topline=false, bottomline=false, leftline=false, rightline=false, backgroundcolor=aliceblue,%
innertopmargin=\topskip, splittopskip=\topskip, skipbelow=\baselineskip, skipabove=\baselineskip]{boxthm}{Theorem}[section]
\newmdtheoremenv[topline=false, bottomline=false, leftline=false, rightline=false, backgroundcolor=aliceblue,%
innertopmargin=\topskip, splittopskip=\topskip, skipbelow=\baselineskip, skipabove=\baselineskip]{boxprop}[boxthm]{Proposition}
\newmdtheoremenv[topline=false, bottomline=false, leftline=false, rightline=false, backgroundcolor=aliceblue,%
innertopmargin=\topskip, splittopskip=\topskip, skipbelow=\baselineskip, skipabove=\baselineskip]{boxcor}[boxthm]{Corollary}
\newmdtheoremenv[topline=false, bottomline=false, leftline=false, rightline=false, backgroundcolor=aliceblue,%
innertopmargin=\topskip, splittopskip=\topskip, skipbelow=\baselineskip, skipabove=\baselineskip]{boxlem}[boxthm]{Lemma}
\newmdtheoremenv[topline=false, bottomline=false, leftline=false, rightline=false, backgroundcolor=aliceblue,%
innertopmargin=\topskip, splittopskip=\topskip, skipbelow=\baselineskip, skipabove=\baselineskip]{boxdef}[boxthm]{Definition}

\usepackage{stmaryrd}

\usepackage[table]{xcolor}   
\definecolor{rightblue}{RGB}{76,114,176} 
\definecolor{rightorange}{RGB}{221,132,82} 
\definecolor{aliceblue}{rgb}{0.94, 0.97, 1.0} 
\definecolor{darkcerulean}{rgb}{0.03, 0.27, 0.49} 
\definecolor{iris}{rgb}{0.35, 0.31, 0.81} 
\definecolor{carmine}{rgb}{0.59, 0.0, 0.09} 
\definecolor{green(munsell)}{rgb}{0.0, 0.66, 0.47} 
\definecolor{celadon}{rgb}{0.67, 0.88, 0.69} 
\definecolor{darklavender}{rgb}{0.45, 0.31, 0.59}
\definecolor{darkmagenta}{rgb}{0.55, 0.0, 0.55}
\definecolor{darkorchid}{rgb}{0.6, 0.2, 0.8}

\definecolor{color1}{HTML}{D8D5F2}
\definecolor{color2}{HTML}{84A2C6}
\definecolor{color3}{HTML}{3A7376}
\definecolor{color4}{HTML}{0F3222}

\definecolor{orangeintro}{HTML}{EC5800}
\definecolor{blueintro}{HTML}{4682B4}

\usepackage{booktabs}
\usepackage{multirow}
\usepackage{graphicx}
\usepackage{caption} 
\usepackage{hyperref}
  \hypersetup{
    final=true,
    plainpages=false,
    pdfstartview=FitV,
    pdftoolbar=true,
    pdfmenubar=true,
    bookmarksopen=true,
    bookmarksnumbered=true,
    breaklinks=true,
    linktocpage=true,
    colorlinks=true,
    linkcolor=red, 
    urlcolor=iris, 
    citecolor=darkcerulean, 
    anchorcolor=black
  }

\def\hypref#1{
  \hyperref[#1]{~\cref{#1}}
}

\usepackage{subfig}

\usepackage{algorithm2e}
\RestyleAlgo{ruled}

\usepackage{varwidth}

\usepackage{dirtytalk}

\usepackage{tikz}
\usepackage{pgfplots}
\usetikzlibrary{arrows}
\usetikzlibrary{patterns}

\usepackage{algpseudocode} 

\usepackage{wrapfig}

\usepackage{etoolbox}
\newlength{\radius}
\newrobustcmd{\mytriangle}[1]{\tikz{\draw[line width=0.2mm, fill=#1] (90:\radius) -- (210:\radius) -- (330:\radius) -- cycle;}}

\newrobustcmd{\mysquare}[1]{\tikz{\draw[line width=0.2mm, fill=#1] (45:\radius) -- (135:\radius) -- (225:\radius) -- (315:\radius) -- cycle;}}

\newrobustcmd{\mycircle}[1]{\tikz{\draw[line width=0.2mm, fill=#1] (0,0)  circle (0.2\radius);}}

\makeatletter
\newcommand{\manuallabel}[2]{\def\@currentlabel{#2}\label{#1}}
\makeatother

\renewcommand*{\proofname}{Proof Sketch}

\begin{document}

\addtocontents{toc}{\protect\setcounter{tocdepth}{0}}

%

%
\runningauthor{Ambroise Odonnat, Vasilii Feofanov, Ievgen Redko}

\twocolumn[
\def\affiliation{Huawei Noah's Ark Lab}
\aistatstitle{Leveraging Ensemble Diversity for Robust Self-Training in the Presence of Sample Selection Bias}
\aistatsauthor{ Ambroise Odonnat\textsuperscript{*} \And Vasilii Feofanov \And  Ievgen Redko}
\aistatsaddress{\affiliation \\ \'Ecole des Ponts ParisTech \\ ENS Paris-Saclay \\ \And \affiliation \And \affiliation} 
]

\begin{abstract}
Self-training is a well-known approach for semi-supervised learning. It consists of iteratively assigning pseudo-labels to unlabeled data for which the model is confident and treating them as labeled examples. For neural networks, \texttt{softmax} prediction probabilities are often used as a confidence measure, although they are known to be overconfident, even for wrong predictions. This phenomenon is particularly intensified in the presence of sample selection bias, i.e., when data labeling is subject to some constraints. To address this issue, we propose a novel confidence measure, called $\mathcal{T}$-similarity, built upon the prediction diversity of an ensemble of linear classifiers. We provide the theoretical analysis of our approach by studying stationary points and describing the relationship between the diversity of the individual members and their performance. We empirically demonstrate the benefit of our confidence measure for three different pseudo-labeling policies on classification datasets of various data modalities. The code is available at \href{https://github.com/ambroiseodt/tsim}{\texttt{https://github.com/ambroiseodt/tsim}}.
\end{abstract}

\section{Introduction}
Deep learning has been remarkably successful in the past decade when large amounts of labeled data became available \citep{he2016residual, goodfellow2014generative, dosovitskiy2021an}. However, in many real-world applications data annotation is costly and time-consuming \citep{9356271}, while data acquisition is cheaper and may result in an abundance of unlabeled examples \citep{NIPS2009_1651cf0d}. In this context, semi-supervised learning \citep[denoted by \texttt{SSL}]{Chapelle:2006} has emerged as a powerful approach to exploit both labeled and unlabeled data \citep{vanEngelen2020}. Among existing \texttt{SSL} techniques, self-training~\citep{amini2023selftraining} received a lot of interest in recent years \citep{lee_pl, Sohn:2020:FixMatch, zhang2021flexmatch, DST}. The main idea behind the self-training approach is to use the predictions of a classifier learned on available labeled data to pseudo-label unlabeled data and progressively include them in the labeled set during the training. Traditionally, at each iteration of self-training, we select for pseudo-labeling the unlabeled examples that have a prediction confidence above a certain threshold. The latter can be fixed \citep{Yarowsky:1995}, dynamic along the training \citep{cascante2021curriculum}, or optimized \citep{Feofanov:2019}.
\begin{figure}[t!]
   \centering
   \includegraphics[width=0.5\textwidth]{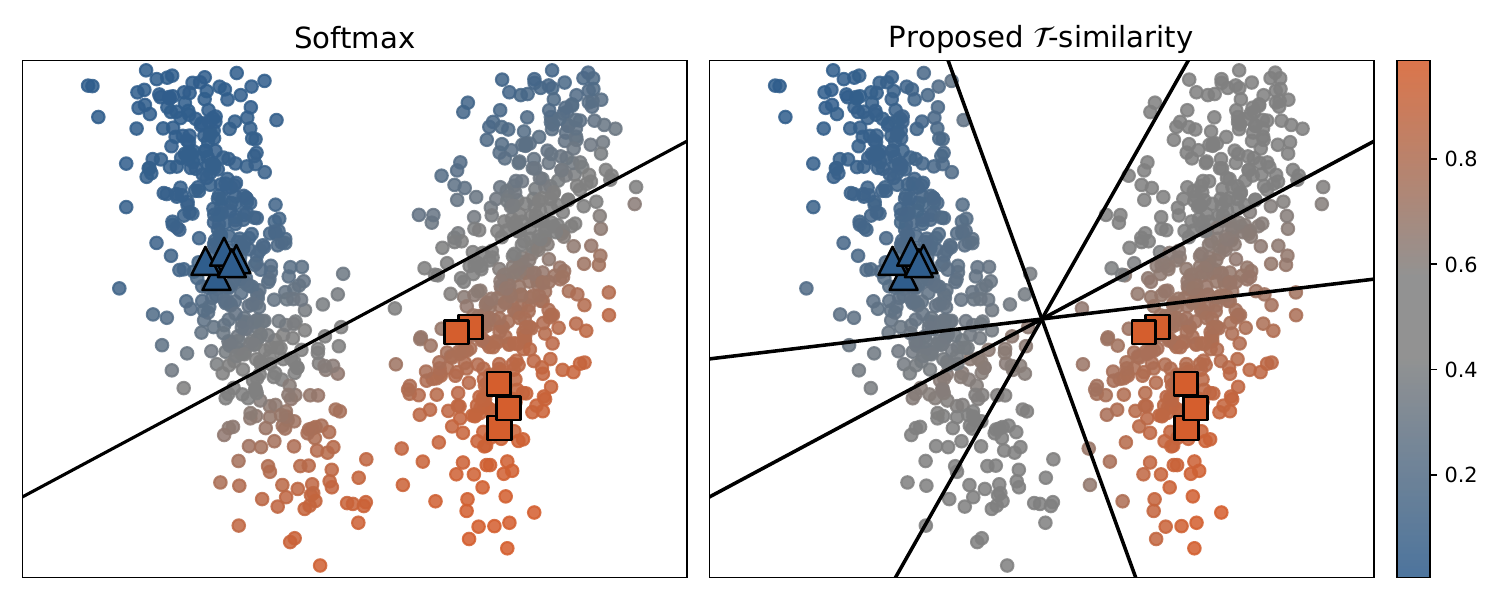}
   \caption{Unlabeled data (circles) colored by the confidence value of being from the \textcolor{orangeintro}{orange} class (right point cloud), from \textcolor{blueintro}{blue} to \textcolor{orangeintro}{orange} as it increases. \textbf{Left}: Given a model trained on few labeled examples (\mytriangle{blueintro} \mysquare{orangeintro}), \texttt{softmax} may provide wrong confidence estimates for unlabeled data. \textbf{Right}: Our method averages confidence estimates of a diverse set of classifiers leading to a well-calibrated model robust to distribution shift.}
   \label{fig:illustration_intro}
\end{figure}
When it comes to analyzing the performance of self-training algorithms, two fundamental questions primarily arise: \textbf{(1)} How to rank unlabeled data to reflect their difficulty for classification? \textbf{(2)} How to select the threshold for pseudo-labeling at each iteration? While the second question is related to the choice of a pseudo-labeling policy and is already well addressed in the literature \citep{Feofanov:2019,cascante2021curriculum,zhang2021flexmatch}, the first question is related to confidence estimation and remains an open problem as the most conventional choice for such a ranking -- the \texttt{softmax} prediction probability -- is known to suffer from overconfidence \citep{pmlr-v162-wei22d}. 

In this work, we propose a reliable ranking measure for pseudo-labeling and place ourselves in a challenging, yet realistic, scenario of \texttt{SSL} with a distribution shift. For the latter, we consider the sample selection bias (\texttt{SSB}) setup \citep{Heckman:1974} where the annotation of training data is subject to certain constraints. Selection bias is known to occur in survey design \citep{10.5555/1462129}, in medical research studies during the creation of cohorts and control groups \citep{ALVES2006205,AHERN2018109,ARIAS2023191}, or in industry due to privacy or security reasons. We denote this challenging scenario by \texttt{SSL} + \texttt{SSB}. 

\paragraph{Summary of our contributions.} We summarize our contributions as follows:
\begin{enumerate}
    \item We propose a novel confidence measure, illustrated in Figure~\ref{fig:illustration_intro}, that builds upon the diversity of an ensemble of classifiers. Such a measure is easy to implement into any popular \texttt{SSL} method using neural networks as a backbone;
    \item We provide a thorough theoretical analysis of our method by studying stationary points and showing the connection between diversity and the performance of the individual classifiers;
    \item We experimentally demonstrate the superiority of our approach for self-training on various \texttt{SSL} datasets under \texttt{SSB}. Additionally, we show that \texttt{SSB} degrades the performance of other popular methods when not dealt with properly.
\end{enumerate}

\section{Related Work}
\label{sec:rel-work}

\paragraph{Self-training.} In \texttt{SSL}, a classical strategy to incorporate unlabeled data into the learning process is to use the predictions of a supervised model on it, either for regularization \citep{Grandvalet:2004:EntropyMin,Feofanov:2023} or for self-training by iteratively including most confident pseudo-labeled data to the labeled set. The latter approach is widespread in computer vision, where self-training is often combined with consistency regularization to encourage similar predictions for different augmentations of the same unlabeled image \citep{Sohn:2020:FixMatch,zhang2021flexmatch,DST}. Correctly choosing the confidence threshold for unlabeled data is key to the success of self-training. Instead of using a fixed threshold, several works propose to select the threshold at each iteration via curriculum learning to control the number of pseudo-labeled examples \citep{cascante2021curriculum,zhang2021flexmatch}, while \citet{Feofanov:2019} finds the threshold as a balance between the upper-bounded transductive error on the pseudo-labeled examples and their number. All of these methods strongly depend on the confidence measure of the base classifier, and thus are not well suited when the base classifier is biased towards the labeled set, or when labeled and unlabeled data follow different probability distributions. In this work, we aim to fill this gap and propose a model- and application-agnostic confidence estimation approach that is robust to such distribution mismatch.  

\paragraph{Sample selection bias.} \texttt{SSB} describes the situation where the distribution mismatch between labeled and unlabeled data is due to some unknown sample selection process, i.e., when data labeling is subject to some constraints. Formalized by 
\citet{Heckman:1974}, this framework received a lot of attention in the 1980s in the case of linear regression from the econometrics community \citep{cf860a73-19e3-3e55-8fdd-0a3e0f31a64e, 0316ac5c-b340-35fa-ab28-c927e20316b7}. In the context of classification, \texttt{SSB} was properly defined by \citet{Zadrozny:2004}, and most of the methods address it via importance sampling by estimating biased densities or selection probabilities \citep{Zadrozny:2004, NIPS2005_a36b0dcd, Shimodaira2000} or by using prior knowledge about the class distributions \citep{Lin2002}. Alternatively, the resampling weights can be inferred by kernel mean matching
\citep{NIPS2006_a2186aa7}. All these methods heavily rely on density estimation and thus are not well suited in \texttt{SSL} where labeled data is scarce. In our work, we propose to turn this curse of scarcity of labeled data into a blessing by exploiting the diversity of a set of classifiers that can be fit to a handful of available labeled points. 
\paragraph{Ensemble diversity.} It is well known that an ensemble of learners \citep{HansenSalamon:1990} is efficient when its members are \textit{\say{diverse}} in a certain sense \citep{Dietterich:2000,Kuncheva:2004, Lu:2010:AccDivTradeoff}.
Over the last decades, generating diversity has been done in many ways, including bagging \citep{Breiman2001}, boosting \citep{freund1997decision,friedman2001greedy}, and random subspace training \citep{ho1998random}. These methods, however, are based on implicit diversity criteria, calling for new approaches where the ensemble diversity can be defined explicitly. To this end, \citet{Liu:1999:NegCorLoss} introduced a mixture of experts that are diversified through the negative correlation loss that forces a trade-off between specialization and cooperation of the experts. \citet{buschjäger2020generalized} derive a bias-variance decomposition that encompasses many existing diversity methods, particularly showing that the negative correlation loss is linked to the prediction variance of the ensemble members. \citet{pmlr-v151-ortega22a} derive an upper bound over the generalization error of the majority vote classifier showing that the performance of the ensemble depends on the error variance of the individual classifiers. Some recent works rely on ensemble diversity to estimate accuracy on a given test set, namely, \citet{jiang2022assessing} use the disagreement rate of two independently trained neural networks, while \citet{chen2021detecting} evaluate the disagreement between a deep ensemble and a given pre-trained model. The closest method to our work is that of \citet{Zhang2013} which learns an ensemble classifier by imposing diversity in predictions on unlabeled data. In this paper, we extend their binary setting to multi-class classification and push their idea further by showing the benefits of using diversity for calibration and confidence estimation in self-training under distribution shift. In addition, we provide a theoretical explanation of why the diversity imposed in such a way works in practice.

\section{Our Contributions}
\label{sec:contributions}
\paragraph{Notations.} Scalar values are denoted by regular letters (e.g., parameter $\lambda$), vectors are represented in bold lowercase letters (e.g., vector $\mathbf{x}$) and matrices are represented by bold capital letters (e.g., matrix $\mathbf{A}$). The $i$-th row of the matrix $\mathbf{A}$ is denoted by $\mathbf{A}_{i}$ and its $j$-th column is denoted by $\mathbf{A}_{\cdot, j}$. The trace of a matrix $\mathbf{A}$ is denoted by $\mathrm{Tr}(\mathbf{A})$ and its transpose by $\mathbf{A}^\top$. The identity matrix of size $n$ is denoted by $\mathbf{I}_n \in \mathbb{R}^{n \times n}$. 
We denote by $\lambda_{\mathrm{min}}(\mathbf{A})$ and $\lambda_{\mathrm{max}}(\mathbf{A})$ the minimum and maximum eigenvalues of a matrix $\mathbf{A}$, respectively.

\subsection{Problem setup} 
\paragraph{Semi-supervised learning.} Consider the classification problem with input space $\mathcal{X}$ and label space $\mathcal{Y}\!=\!\{1, \dots, C\}$. Let $(\mathbf{X}_\ell, \mathbf{y}_\ell)\!=\!(\mathbf{x}_i, y_i)_{i=1}^{n_\ell} \in \left(\mathcal{X} \times \mathcal{Y} \right)^{n_\ell}$ be the set of labeled training examples. Let $\mathbf{X}_u = (\mathbf{x}_i)_{i=n_\ell+1}^{n_\ell + n_u}\!\in\!\mathcal{X}^{n_u}$ be the set of unlabeled training examples. The hypothesis space is denoted by  $\mathcal{H} = \{h \colon \mathcal{X}\!\to\!\Delta_C \}$, where $\Delta_C\!=\!\{p \in [0, 1]^C | \sum_{c=1}^C p_c = 1\}$ is the probability simplex. For an input $\mathbf{x}\!\in\!\mathcal{X}$, a learning model $h\!\in\!\mathcal{H}$, $h(\mathbf{x})$ is a probability measure on $\mathcal{Y}$, and the predicted label is defined as $\hat{y} = \argmax h(\mathbf{x}) $. 

\paragraph{Sample selection bias.} We model \texttt{SSB} for labeled data following \citet[chap. 3]{10.5555/1462129} and introduce a random binary selection variable $s$, where $s=1$ means that the training point is labeled, while $s=0$ implies that it remains unlabeled. Assuming the true stationary distribution of data $P$ on $\mathcal{X} \times \mathcal{Y}$, we consider that labeled training examples are i.i.d. drawn from $P_\mathrm{lab}$, while unlabeled training and test examples are from $P_\mathrm{unlab}$ and $P_\mathrm{test}$ respectively, with $P_\mathrm{lab}$, $P_\mathrm{unlab}$ and $P_\mathrm{test}$ defined as follows:
\begin{align*}
    P_\mathrm{lab}(\mathbf{x}, y) &= P(\mathbf{x}, y | s=1) = \frac{P(s=1|\mathbf{x}, y)}{P(s=1)}P(\mathbf{x}, y),\\
    P_\mathrm{unlab}(\mathbf{x}, y) &= P_\mathrm{test}(\mathbf{x}, y) = P(\mathbf{x}, y).
\end{align*}

\paragraph{Self-training.} 
Most commonly, a self-training algorithm is first initialized by the base classifier trained using only labeled data $(\mathbf{X}_l, \mathbf{y}_\ell)$. Then, at each iteration $i$, the algorithm measures the prediction confidence for unlabeled points from $\mathbf{X}_u$, typically, through prediction probabilities like \texttt{softmax}. Based on these confidence estimates, a pseudo-labeling policy determines the unlabeled examples that are pseudo-labeled by the corresponding model's predictions. These pseudo-labeled data are moved from $\mathbf{X}_u$ to $(\mathbf{X}_\ell, \mathbf{y}_\ell)$ and the classifier is re-trained. The same procedure is repeated for several iterations until stop criteria are satisfied. Algorithm~\ref{alg:self_training_algorithm}, given in Appendix~\ref{app:self_training_algorithm}, outlines the pseudo-code of this learning method.

\subsection{Proposed approach}
\paragraph{Similarity as a surrogate of confidence.}
Our main idea is to use the similarity between ensemble predictions on a given unlabeled example to estimate the prediction's confidence, instead of using the usual \texttt{softmax} prediction probability. The underlying intuition is to say that if the individual, but diverse, classifiers agree on a point, then the associated prediction can be trusted with high confidence. Conversely, if we find many ways to disagree on a given point then it is likely a difficult point for our model, so a low confidence is attributed. Below, we formalize the proposed confidence measure:
\begin{boxdef}[$\mathcal{T}$-similarity]
\label{def:t_similarity}
Consider an unlabeled data point $\mathbf{x}\!\,\in\!\,\mathbf{X}_u$ and an ensemble of classifiers $\mathcal{T}\!=\!\{h_m\!\,\in\!\,\mathcal{H}| 1\!\,\leq\!\,m\!\,\leq\!\,M \}$. The $\mathcal{T}$-similarity of $\mathbf{x}$ is defined by:
\begin{equation*}
    s_\mathcal{T}(\mathbf{x}) = \frac{1}{M(M-1)} \sum_{m \neq k} h_m(\mathbf{x})^\top h_k(\mathbf{x}).
\end{equation*}
\end{boxdef}
We now present a simple property of the proposed confidence measure $s_\mathcal{T}$ highlighting that it can be used as a drop-in replacement of the \texttt{softmax} prediction probabilities for confidence estimation when the output of the individual classifiers lies in the probability simplex $\Delta_C$. The proof is given in Appendix~\ref{app:similarity_0_1}.
\begin{boxprop}[Property of $s_\mathcal{T}$]
\label{prop:similarity_0_1}
Let $\mathcal{T}$ be an ensemble of probabilistic classifiers. Then, for any input $x \in \mathcal{X}$, we have:
\begin{equation*}
    0 \leq s_\mathcal{T}(\mathbf{x}) \leq 1.
\end{equation*}
\end{boxprop}
\begin{proof}
    As each classifier outputs a probability measure $h_m(\mathbf{x})$, $s_\mathcal{T}(\mathbf{x})$ is nonnegative and we can upper-bound each element of $h_m(\mathbf{x})$ by $1$ to obtain the desired upper bound. 
\end{proof}
\begin{figure}[!t]
\centering
\includegraphics[width=0.9\linewidth, trim={0cm 1cm 0cm 1cm},clip=true]{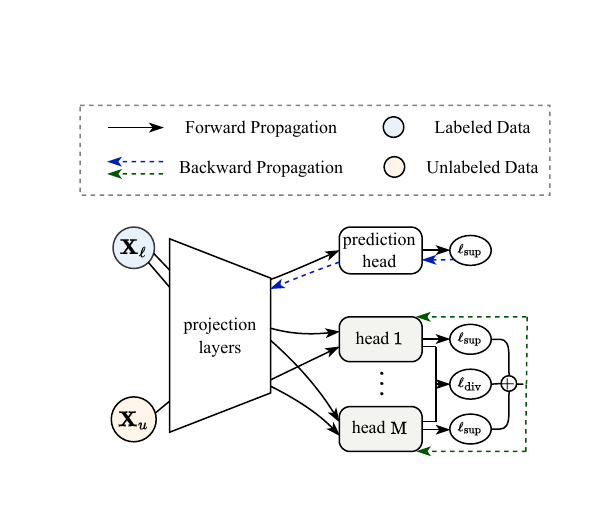}
\caption{\small Architecture of the model.}
\label{fig:architecture}
\end{figure}

\paragraph{Learning with the $\mathcal{T}$-similarity.} Following \citet{Zhang2013}, we train an ensemble to fit well the labeled set while having the most diverse possible individual predictions on the unlabeled set. We achieve this by minimizing the following loss function:
\begin{align}
\label{eq:diverse_ensemble_generic_loss}
    & \mathcal{L}_{\text{conf}\,}(\mathcal{T}, (\mathbf{X}_\ell, \mathbf{y}_\ell), \mathbf{X}_u)  
    \notag \\
     &=  \underbrace{(1/M) \sum_{m=1}^M \ell_{\text{sup}}(h_m, \mathbf{X}_\ell, \mathbf{y}_\ell)}_{\text{label fidelity term }}  \  - \ \underbrace{ 
     \vphantom{\sum_{m=1}^M}\gamma\,\ell_{\mathrm{div}}(\mathcal{T}, \mathbf{X}_u)}_{\text{diversity term }},
\end{align}
where $\ell_{\text{sup}}$ is a supervised loss evaluated on the labeled examples, typically the cross-entropy loss, $\ell_{\mathrm{div}}$ corresponds to the diversity loss of the ensemble $\mathcal{T}$, and $\gamma \geq 0$ is a hyperparameter that controls the strength of the imposed diversity. As maximizing the diversity amounts to minimizing the similarity, we consider
\begin{equation}
\label{eq:l_div}
\ell_{\mathrm{div}}(\mathcal{T}, \mathbf{X}_u) = - \frac{1}{n_u}\sum_{\mathbf{x} \in \mathbf{X}_u} s_\mathcal{T}(\mathbf{x}). 
\end{equation}
As labeled data is scarce, the range of classifiers that can fit it -- while being diverse on unlabeled data -- is large. Intuitively, such classifiers will disagree on examples far from labeled data, i.e., in \emph{unsafe} regions, but will strongly agree on samples close to it, i.e., in \emph{safe} regions. Eq.~\eqref{eq:diverse_ensemble_generic_loss} can be seen as a proxy to access this intractable range, while $s_\mathcal{T} (\mbf{x})$ characterizes the agreement on each input $\mbf{x}$.

\paragraph{Practical implementation.} 
To combine confidence estimation and prediction, we introduce the neural network described in Figure~\ref{fig:architecture}. First, input data is projected on a high-dimensional feature space. The projection layers are learned together with a classification head that is also used for predicting pseudo-labels. This prediction head, denoted by $h_{\text{pred}}$, is updated via backpropagation of the supervised loss $\mathcal{L}_{\text{sup}\,}(h_{\text{pred}}, (\mathbf{X}_\ell, \mathbf{y}_\ell))\!=\!\ell_{\text{sup}}(h_{\text{pred}}, (\mathbf{X}_\ell, \mathbf{y}_\ell))$.
Another part of the network is responsible for confidence estimation. After projecting inputs to the hidden space, an ensemble of $M$ heads follows and is optimized using Eq.~\eqref{eq:diverse_ensemble_generic_loss}. The important thing is that backpropagation of Eq.~\eqref{eq:diverse_ensemble_generic_loss} only influences the ensemble heads, not the projection layers, since we want to estimate prediction confidence given a fixed representation. Then, the total loss for the network is:
\begin{equation*}
\mathcal{L}_{\text{sup}\,}(h_{\text{pred}}, (\mathbf{X}_\ell, \mathbf{y}_\ell)) + \mathcal{L}_{\text{conf}\,}(\mathcal{T}, (\mathbf{X}_\ell, \mathbf{y}_\ell), \mathbf{X}_u).
\end{equation*}
In our experiments, we consider $M\!=\!5$ linear heads. It results in fast training and no significant computational burden. The ease of implementation enables the combination of the proposed $\mathcal{T}$-similarity with any \texttt{SSL} method that uses a neural network as a backbone.

\subsection{Theoretical analysis}
We now provide the theoretical guarantees of the proposed learning framework. For the sake of clarity, we formulate the problem in the binary classification case and consider only the confidence estimation part (the ensemble of heads) for a fixed representation space. We show that, under a mild assumption, the 
solution of the optimization problem (minimization of Eq.~\eqref{eq:diverse_ensemble_generic_loss}) is unique. In addition, we establish a lower bound on the diversity of the optimal ensemble that gives theoretical insights into the relationship between diversity and the performance of the individual classifiers from the considered ensemble. Finally, we provide some understanding of the role of representation learning on diversity.  

\paragraph{Problem formulation.} Assuming $\mathcal{Y}\!=\!\{-1, +1\}$ and centered training data, we parameterize a linear head $h_m$, $m\!\in\!\llbracket 1, M \rrbracket$, by a separating hyperplane $\boldsymbol{\omega}_m\!\in\!\mathbb{R}^d$ and $h_m(\mathbf{x})\!=\!\boldsymbol{\omega}_m^\top \mathbf{x}\!\in\!\mathbb{R}$ so that Proposition~\ref{prop:similarity_0_1} no longer holds. For a given example $\mathbf{x} \in \mathbb{R}^d$, the classifier $h_m$ predicts the label by $\text{sign}(h_m(\mathbf{x}))$. We denote by $\mathbf{W} \in \mathbb{R}^{d \times M}$ the matrix whose columns are the separating hyperplanes $\boldsymbol{\omega}_m$, i.e., $\forall m \in \llbracket 1,M \rrbracket, \mathbf{W}_{\cdot, m} = \boldsymbol{\omega}_m \in \mathbb{R}^d$. In the rest of this section, we refer to $\mathbf{W}$ as the ensemble of classifiers instead of using the notation $\mathcal{T}$. For practical considerations from the theoretical point of view, we consider ridge (also known as LS-SVM \citep{suykens1999least}) classifiers that minimize the least-square loss with Tikhonov regularization. Then, following this setup, we can re-write Eq.~\eqref{eq:diverse_ensemble_generic_loss}, and formulate the optimization problem as:
\begin{align} 
\label{eq:neg_corr_linear_ensemble_with_reg}
     &\argmin_{\mathbf{W} \in \mathbb{R}^{d \times M}} \mathcal{L}(\mathbf{W}) \coloneqq \notag \\ 
     & \underbrace{\frac{1}{Mn_\ell} \sum_{m=1}^M \sum_{i=1}^{n_\ell} \left (y_i - \boldsymbol{\omega}_m^\top \mathbf{x}_i\right)^ 2}_{\text{label fidelity term}}  + \underbrace{\frac{1}{M} \sum_{m=1}^M\lambda_m \lVert \boldsymbol{\omega}_m \rVert_2^2}_{\text{regularization}} \notag \\
    & \qquad + \underbrace{\frac{\gamma}{n_uM(M-1)} \sum_{m \neq k} \sum_{i=n_\ell+1}^{n_\ell +n_u} \bm{\omega}_m^\top\mathbf{x}_i \bm{\omega}_k^\top\mathbf{x}_i}_{\text{agreement term}}, \tag{\textbf{P}}
\end{align}
where the agreement term corresponds to the opposite of the diversity term, i.e., to $-\gamma\ell_{\mathrm{div}}(\mathbf{W}, \mathbf{X}_u)$. In the binary setting, the similarity measure does not lie in the $[0,1]$ interval anymore and can take any real value. However, with a reasonable choice of $\gamma$, the binarized objective should still lead to diverse ensembles. In the next paragraph, we show that, under a mild assumption on the $\lambda_m$, Problem~\eqref{eq:neg_corr_linear_ensemble_with_reg} can be solved efficiently.

\paragraph{Convergence to a stationary point.} In practice, as $\mathcal{L}$ is differentiable, the learning problem is solved via gradient descent, which aims at finding local minimizers of $\mathcal{L}$. Such minimizers are stationary points, i.e., solutions of the Euler equation
\begin{equation}
    \label{eq:stationary_point_euler}
    \nabla \mathcal{L}(\mathbf{W}) = 0.
\end{equation}
We provide a closed-form expression of the gradient $\nabla \mathcal{L}(\mathbf{W})$ in Proposition~\ref{prop:closed_form_gradient}, which we defer to Appendix~\ref{app:closed_form_gradient}, and show in Proposition~\ref{prop:sol_P_linear_problem}, deferred to Appendix~\ref{app:stationary_point_euler}, that Eq.~\eqref{eq:stationary_point_euler} is equivalent to a linear problem in $\mathbf{W}$. We now make the following assumption on the parameters $\lambda_m$: 
\begin{enumerate}[label=\textbf{A.}, wide, labelwidth=!, labelindent=0pt]
    \item $\forall m \in \llbracket 1,M \rrbracket, \lambda_m > \frac{\gamma(M+1)}{n_u(M-1)}\lambda_{\mathrm{max}}(\mathbf{X}_u^\top \mathbf{X}_u).$ \manuallabel{assumption:pd}{A}
\end{enumerate}
Assumption~\ref{assumption:pd} ensures that $\lambda_m \mathbf{I}_d - \frac{\gamma(M+1)}{n_u(M-1)} \mathbf{X}_u^\top \mathbf{X}_u$ is positive definite, and thereby that $\lambda_m > 0$, for all $m$. The next proposition establishes the convergence of Problem~\eqref{eq:neg_corr_linear_ensemble_with_reg} towards a unique solution. The fact that the loss function $\mathcal{L}$ includes cross-terms between distinct $\boldsymbol{\omega}_m, \boldsymbol{\omega}_k$ makes the proof of its convergence to a stationary point somewhat involved and we defer it to Appendix~\ref{app:prop_loss_func}.
\begin{boxprop}[Convergence of Problem~\eqref{eq:neg_corr_linear_ensemble_with_reg}]
\label{prop:loss_func_properties}
    Under Assumption~\ref{assumption:pd}, $\mathcal{L}$ is strictly convex and coercive on $\mathbb{R}^{d \times M}$. Hence, Problem~\eqref{eq:neg_corr_linear_ensemble_with_reg} admits a unique solution $\mathbf{W}^*$ that verifies Eq.~\eqref{eq:stationary_point_euler}.
\end{boxprop}
\begin{proof}
We first reformulate $\mathcal{L}$ into three terms easier to analyze and demonstrate separately their differentiability, strict convexity, and coercivity using classical matrix analysis tools. It leads to $\mathcal{L}$ being differentiable (thus continuous), strictly convex, and coercive on $\mathbb{R}^{d \times M}$. The convergence of \eqref{eq:neg_corr_linear_ensemble_with_reg} to a unique global minimizer solution of Eq.~\eqref{eq:stationary_point_euler}, follows from those properties. 
\end{proof}
Proposition~\ref{prop:loss_func_properties} highlights the fact that, under assumption~\ref{assumption:pd}, solving Problem~\eqref{eq:neg_corr_linear_ensemble_with_reg} amounts to solving Eq.~\eqref{eq:stationary_point_euler}. In the rest of this section, we study a measure of diversity for the stationary points of $\mathcal{L}$, i.e., solutions of Eq.~\eqref{eq:stationary_point_euler}. 

\paragraph{Relationship between diversity and individual performance.}
We proceed with the analysis of the diversity loss $\ell_{\mathrm{div}}$ on the unlabeled set re-written as:
\begin{equation}
    \label{def:diversity_ensemble}
    \ell_{\mathrm{div}}(\mathbf{W}, \mathbf{X}_u) \!=\! - \frac{1}{n_uM(M-1)} \sum_{m \neq k} \boldsymbol{\omega}_m ^\top \mathbf{X}_u^\top \mathbf{X}_u \boldsymbol{\omega}_k.
\end{equation}
We now want to develop our intuition about when $\ell_{\mathrm{div}}$ achieves its maximum value. We note that $\mathbf{X}_u^\top \mathbf{X}_u$ is positive semi-definite, thus $\mathbf{X}_u^\top\mathbf{X}_u\boldsymbol{\omega}_m$ remains in the same half-space as $\boldsymbol{\omega}_m$. Positive values of $\ell_{\mathrm{div}}$ are then achieved when the angles between the $\boldsymbol{\omega}_m$ and the $\boldsymbol{\omega}_k$ are between $90$ and $180$ degrees. The next theorem characterizes the diversity of the stationary points of $\mathcal{L}$, i.e., the solutions of Eq.~\eqref{eq:stationary_point_euler}. The detailed proof is given in Appendix~\ref{app:diversity_characterization}.
\begin{boxthm}[A lower bound on the diversity]
\label{thm:diversity_characterization}
Let $\tilde{\mathbf{W}}$ be a stationary point of $\mathcal{L}$, i.e., solution of Eq.~\eqref{eq:stationary_point_euler}. 
We denote by $\tilde{\boldsymbol{\omega}}_m$ its $m$-th column and assume that $\frac{1}{M} \sum_{m=1}^M \lambda_m \lVert \tilde{\boldsymbol{\omega}}_m \rVert_2^2 \geq 1$.
Then, we have
\begin{equation*}
\begin{split}
    &\gamma\ell_{\mathrm{div}}(\tilde{\mathbf{W}}, \mathbf{X}_u) \geq \frac{1}{2n_\ell M} \sum_{m=1}^M \lVert \mathbf{y}_\ell - \mathbf{X}_\ell \tilde{\boldsymbol{\omega}}_m \rVert_2^2 \\
    & \qquad + \frac{1}{2M} \sum_{m=1}^M \tilde{\boldsymbol{\omega}}_m^\top \left( \lambda_m \mathbf{I}_d + \frac{\mathbf{X}_\ell^\top \mathbf{X}_\ell}{n_\ell} \right) \tilde{\boldsymbol{\omega}}_m.
\end{split}
\end{equation*}
\end{boxthm}
\begin{proof}
    Using classical matrix analysis tools, we derive a closed-form expression of the stationary points of $\mathcal{L}$ in Proposition~\ref{prop:sol_P_linear_problem}, deferred to Appendix~\ref{app:stationary_point_euler}. It enables us to formalize $\gamma\ell_{\mathrm{div}}(\tilde{\mathbf{W}}, \mathbf{X}_u)$ in closed form, which leads to the desired lower bound.
\end{proof}
From Theorem~\ref{thm:diversity_characterization}, we obtain that the diversity loss of stationary points of $\mathcal{L}$ is non-negative, although, as discussed above, $\ell_{\mathrm{div}}(\mathbf{W}, \mathbf{X}_u)$ may be negative. This implies that the diversity term encourages opposite predictions between classifiers, while the labeled loss term minimized explicitly by our approach likely prevents completely colinear solutions that would degrade the supervised loss too much. We also note that the second term, which we do not optimize explicitly, can be decomposed into a weighted sum of the norms of the individual classifiers $\frac{1}{2M}\sum_{m=1}^M  \lambda_m  ||\tilde{\boldsymbol{\omega}}_m||_2^2$ and a margin term $\frac{1}{2M {n_\ell}}\sum_{m=1}^M \left( \tilde{\boldsymbol{\omega}}_m ^\top \mathbf{X}_\ell^\top \mathbf{X}_\ell \tilde{\boldsymbol{\omega}}_m \right)$. Assuming for simplicity that the $\tilde{\boldsymbol{\omega}}_m$ are orthogonal, it implies that high diversity is achieved by finding predictors of the largest possible margin so that they also span the $M$ directions of the largest variance of the labeled data. This insight is quite important as we do not explicitly consider the spectral properties of the labeled data in our approach, yet we implicitly exploit them by using a very simple and lightweight approach. 

\paragraph{The role of representation learning.} So far, we assumed that the labeled data representation is fixed. In this case, we showed that the diversity is high when the individual predictors cover the directions of large variance in the data. Below, we show that the direction of the smallest variance in $\mathbf{X}_\ell$ is also important for diversity, suggesting that labeled data covering the input space evenly might be beneficial to our approach. The proof is given in Appendix~\ref{app:contrastive_learning}.
\begin{boxcor}[The role of representation]
\label{cor:contrastive_learning}
Let $\tilde{\mathbf{W}}$ be a stationary point of $\mathcal{L}$, i.e., solution of Eq.~\eqref{eq:stationary_point_euler}. We denote by $\tilde{\boldsymbol{\omega}}_m$ its $m$-th column. Assuming that the condition of Theorem~\ref{thm:diversity_characterization} holds and that all $\lambda_m$ are equal to some $\lambda$, we have: 
\begin{equation*}
    \gamma \ell_{\mathrm{div}}(\tilde{\mathbf{W}}, \mathbf{X}_u) \geq \frac{1}{2M}\!\left( \lambda + \frac{1}{n_\ell}\lambda_{\mathrm{min}}\left(\mathbf{X}_\ell^\top \mathbf{X}_\ell\right)\right)\!\lVert\tilde{\mathbf{W}} \rVert ^2_\mathrm{F}.
\end{equation*}
\end{boxcor}
\begin{proof}
    The proof follows from the nonnegativity of the first term of the lower bound in Theorem~\ref{thm:diversity_characterization} and classical matrix analysis tools.
\end{proof}
As $\lambda_{\mathrm{min}}(\mathbf{X}_\ell^\top\mathbf{X}_\ell)$ represents the magnitude of the spread in the direction of less variance, we want to push $\lambda_{\mathrm{min}}(\mathbf{X}_\ell^\top\mathbf{X}_\ell)$ away from $0$ to span the whole space as evenly as possible. When $\mathbf{X}_\ell$ is the output of an embedding layer of a neural network, this idea is reminiscent of contrastive learning methods that learn an embedding space having a uniform distribution on the sphere \citep{pmlr-v119-wang20k}. Although our method is not directly linked to contrastive learning, Corollary~\ref{cor:contrastive_learning} gives some insights into how representation learning could help achieve higher diversity. We believe this direction bears great potential for future work.

\section{Experiments}
In this section, we first showcase the failure of self-training in the \texttt{SSB} setting when the \texttt{softmax} is used as a confidence measure. Then, we empirically demonstrate the effectiveness of the $\mathcal{T}$-similarity for confidence estimation and for self-training on common classification datasets with different data modalities. The implementation of the labeling procedure and the $\mathcal{T}$-similarity is open-sourced at \href{https://github.com/ambroiseodt/tsim}{\texttt{https://github.com/ambroiseodt/tsim}}.

\paragraph{Datasets.} 
We consider $13$ publicly available \texttt{SSL} datasets with various data modalities: biological data for \texttt{Cod-RNA} \citep{chang2011libsvm}, \texttt{DNA} \citep{chang2011libsvm}, \texttt{Protein} \citep{dua_2019}, \texttt{Splice} \citep{dua_2019}; images for \texttt{COIL-20} \citep{Nene1996}, \texttt{Digits} \citep{scikit-learn}, \texttt{Mnist} \citep{726791}; tabular data for \texttt{DryBean}  \citep{dua_2019}, \texttt{Mushrooms} \citep{dua_2019}, \texttt{Phishing} \citep{chang2011libsvm}, \texttt{Rice} \citep{dua_2019}, \texttt{Svmguide1} \citep{chang2011libsvm}; time series for \texttt{HAR} \citep{dua_2019}. More details can be found in Appendix~\ref{app:datasets}. All experimental results reported below are obtained over 9 different seeds. 

\paragraph{Labeling procedure.} To generate \texttt{SSL} data, we consider the two following labeling strategies and compare the studied baselines in both cases (see more details in Appendix~\ref{app:labeling_procedure}):

1. \texttt{IID}: this is the case usually considered in classification tasks and that verifies the i.i.d. assumption. The selection variable $s$ is completely random, i.e., independent of $\mathbf{x}$ and $y$. In this case, we have $P(s=1|\mathbf{x}, y) = P(s=1)$ and thus $P_\mathrm{lab}(\mathbf{x}, y) = P(\mathbf{x}, y)$.

2. \texttt{SSB}: in this case, $s$ is dependent of $\mathbf{x}$ and $y$. For each class $c$ and each data $\mathbf{x}$ with label $y=c$, we impose:
\begin{equation*}
        P(s=1|\mathbf{x}, y=c) = \frac{1}{\beta} \exp(r \times \lvert \mathrm{proj}_1(\mathbf{x}) \rvert),
\end{equation*}
where $r > 0$ is a hyperparameter, $\mathrm{proj}_1(\mathbf{x})$ is the projection value of $\mathbf{x}$ on the first principal component of the training data in class $c$ and $\beta = \sum_\mathbf{x} \exp(r \times \lvert \mathrm{proj}_1(\mathbf{x}) \rvert)$ is a normalizing constant to ensure that $P(s=1 | \mathbf{x} \in \mathbf{X}_\ell, y=c) = 1$. The impact of the strength of the bias is studied in Appendix~\ref{app:app_ssb_stength}.

\paragraph{Baselines.} We conduct experiments with various pseudo-labeling policies: \texttt{ERM}, the supervised baseline that does not pseudo-label and is trained using labeled data only; $\texttt{PL}_{\theta=0.8}$ with the fixed threshold $\theta\!=\!0.8$ \citep{lee_pl}; $\texttt{CSTA}_{\Delta=0.4}$ with the curriculum step $\Delta\!=\!0.4$ \citep{cascante2021curriculum} and \texttt{MSTA} \citep{Feofanov:2019}. More details about the policies and their implementation can be found in Appendix~\ref{app:policies} and \ref{app:baselines} respectively.
Each baseline is evaluated with both the usual \texttt{softmax} prediction probability and our proposed $\mathcal{T}$-similarity. For the sake of simplicity, we use \texttt{softmax} and $\mathcal{T}$-similarity in the result tables to distinguish those methods.

\begin{figure}[!t]
    \centering
    \includegraphics[width=0.5\textwidth]{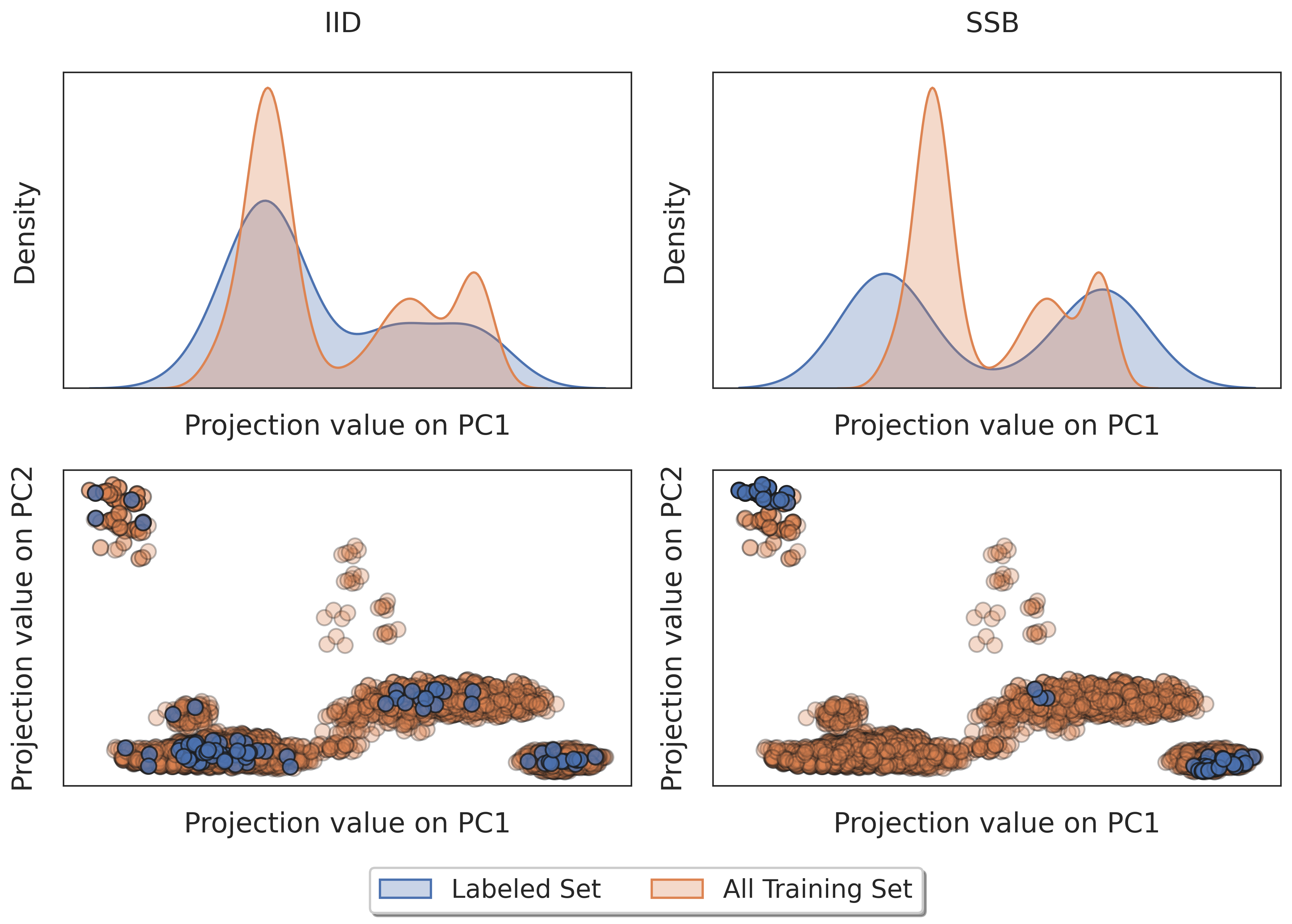}
    \caption{\small Visualization of sample selection bias on \texttt{Mushrooms}. \textbf{First row:} Distribution of the projection values on the first principal component (PC1). \textbf{Second row:} Visualization of the projection values on the PC1 and the PC2.}
    \label{fig:visu_bias_mushrooms}
\end{figure}

\paragraph{Architecture and training parameters.} In all of our experiments, we train a $3$-layer MLP, with the Adam optimizer \citep{KingBa15} and a learning rate of $0.001$. Unless specified otherwise, we consider $M\!=\!5$ linear heads and a diversity strength parameter $\gamma\!=\!1$. In our experiments, we take $\ell_{\mathrm{sup}}$ as the cross-entropy loss. Training is performed during $5$ epochs with $100$ training iterations per epoch. We evaluate the model on a test set of size $25\%$ of the dataset. For each dataset, we run our experiments with $9$ different seeds and display the average and the standard deviation of the test accuracy (both in $\%$) over the $9$ trials. 

\subsection{Visualization of sample selection bias.} As a picture is worth a thousand words, we illustrate the sample selection bias on the \texttt{Mushrooms} dataset \citep{dua_2019}. We select $80$ labeled examples out of the $6093$ training examples with the \texttt{IID} and the \texttt{SSB} procedures. Inspired by \citet{adapt}, we plot in the first row of Figure~\ref{fig:visu_bias_mushrooms} the distribution of the projection values on the first principal component (PC1) of the labeled training set (\textcolor{rightblue}{blue}) and of the whole training set (\textcolor{rightorange}{orange}). Contrary to the \texttt{IID} sampling, we can see that the \texttt{SSB} sampling injects a clear bias between distributions. We also visualize in the second row of Figure~\ref{fig:visu_bias_mushrooms} the projection values on the first two principal components (PC1 and PC2) of the labeled training examples (\textcolor{rightblue}{blue}) and of all the training samples (\textcolor{rightorange}{orange}).  We can see that the \texttt{IID} procedure samples in all regions of the space while the \texttt{SSB} concentrates in specific areas. 

\subsection{Failure of self-training with the \texttt{softmax}}
We start by empirically illustrating that the performance of self-training is heavily dependent on the initial performance of the base classifier when the \texttt{softmax} prediction probabilities are used as a confidence measure.
In Figure~\ref{fig:drop_ssb}, we compare the performance of the base classifier \texttt{ERM} and the self-training methods under the two considered labeling procedures. One can see that all self-training methods together with their base classifier exhibit a drop in performance (in some cases, up to 30\%) indicating that \texttt{softmax} predictions are not robust to the distribution shift. Of particular interest here is \texttt{Mushrooms} data set, where unlabeled data degrades the performance even further by 6.5\% for \texttt{CSTA}$_{\Delta=0.4}$ and by 15.4\% for \texttt{PL}$_{\theta=0.8}$ when we compare them to \texttt{ERM}. We also show a similar failure on the model selection task in Appendix~\ref{app:app_model_selection}. In what follows, we show that our proposal tackles these drawbacks inherent to other baselines.
\def\figlength{0.55}
\def\resizelegend{2.8cm}

\begin{figure}[!h]
\subfloat{
\resizebox{!}{\resizelegend}{%
\begin{tikzpicture}
\begin{axis}[
    ybar,axis on top, font=\Large,
    legend style={font=\fontsize{6}{2}\selectfont,cells={align=left}},width = \figlength\textwidth,
height = 5cm,
    ymin = 58, ymax = 103,
    enlargelimits=0.27,
    enlarge x limits=0.15,
    tick align=inside,
    title={\texttt{ERM}},
    major grid style={draw=white},
    bar width=0.35cm,
    legend style={at={(0.5,-0.4)},
     anchor=north},
    ylabel={Test Accuracy},
    y label style={at={(0.01,0.5)}},
    symbolic x coords={Coil20, DNA, HAR, Mushrooms, Protein},
    xtick=data,
    nodes near coords,
    every node near coord/.append style={font=\footnotesize},
    xticklabel style = {rotate=45, text opacity=0},
    nodes near coords align={vertical},
    nodes near coords style = {rotate=90, anchor=west},
    ]
\addplot [draw=none, fill=color2] coordinates {(Coil20,93.2) (DNA,81.3) (HAR,91.2) (Mushrooms,96.5) (Protein,73.7) };
\addplot [draw=none, fill=color3] coordinates {(Coil20,84.5) (DNA,78.8) (HAR,82.6) (Mushrooms,69.5) (Protein,57.6) };
\end{axis}
\end{tikzpicture}
}
}
\subfloat{
\resizebox{!}{\resizelegend}{%
\begin{tikzpicture}
\begin{axis}[
    ybar,axis on top, font=\Large,
    legend style={font=\fontsize{6}{2}\selectfont,cells={align=left}},width = \figlength\textwidth,
height = 5cm,
    ymin = 58, ymax = 103,
    enlargelimits=0.27,
    enlarge x limits=0.15,
    tick align=inside,
    title={\texttt{PL}$_{\theta=0.8}$},
    major grid style={draw=white},
    bar width=0.35cm,
    legend style={at={(1.15,0.65)},
     anchor=north},
    symbolic x coords={Coil20, DNA, HAR, Mushrooms, Protein},
    xtick=data,
    nodes near coords,
    every node near coord/.append style={font=\footnotesize},
    xticklabel style = {rotate=45, text opacity=0},
    nodes near coords align={vertical},
    nodes near coords style = {rotate=90, anchor=west},
    ]
\addplot [draw=none, fill=color2] coordinates {(Coil20,94.3) (DNA,83.5) (HAR,91.8) (Mushrooms,96.9) (Protein,75.5) };
\addplot [draw=none, fill=color3] coordinates {(Coil20,84.6) (DNA,82.3) (HAR,84.0) (Mushrooms,54.1) (Protein,56.9) };
\end{axis}
\end{tikzpicture}
}
}

\vspace{-1.05cm}
\subfloat{
\resizebox{!}{\resizelegend}{%
\begin{tikzpicture}
\begin{axis}[
    ybar,axis on top, font=\Large,
    legend style={font=\fontsize{6}{2}\selectfont,cells={align=left}},width = \figlength\textwidth,
height = 5cm,
    ymin = 58, ymax = 103,
    enlargelimits=0.27,
    enlarge x limits=0.15,
    tick align=inside,
    title={\texttt{CSTA}$_{\Delta=0.4}$},
    major grid style={draw=white},
    bar width=0.35cm,
    legend style={at={(0.5,-0.4)},
     anchor=north},
    ylabel={Test Accuracy},
    y label style={at={(0.01,0.5)}},
    symbolic x coords={Coil20, DNA, HAR, Mushrooms, Protein},
    xtick=data,
    nodes near coords,
    every node near coord/.append style={font=\footnotesize},
    xticklabel style = {rotate=45},
    nodes near coords align={vertical},
    nodes near coords style = {rotate=90, anchor=west},
    ]
\addplot [draw=none, fill=color2] coordinates {(Coil20,93.1) (DNA,81.5) (HAR,91.4) (Mushrooms,96.3) (Protein,74.7) };
\addplot [draw=none, fill=color3] coordinates {(Coil20,84.4) (DNA,80.1) (HAR,82.2) (Mushrooms,63.0) (Protein,56.1) };
\end{axis}
\end{tikzpicture}
}
}
\subfloat{
\resizebox{!}{\resizelegend}{%
\begin{tikzpicture}
\begin{axis}[
    ybar,axis on top, font=\Large,
    legend style={font=\fontsize{6}{2}\selectfont,cells={align=left}},width = \figlength\textwidth,
height = 5cm,
    ymin = 58, ymax = 103,
    enlargelimits=0.27,
    enlarge x limits=0.15,
    tick align=inside,
    title={\texttt{MSTA}},
    major grid style={draw=white},
    bar width=0.35cm,
    legend style={at={(1.15,0.65)},
     anchor=north},
    symbolic x coords={Coil20, DNA, HAR, Mushrooms, Protein},
    xtick=data,
    nodes near coords,
    every node near coord/.append style={font=\footnotesize},
    xticklabel style = {rotate=45},
    nodes near coords align={vertical},
    nodes near coords style = {rotate=90, anchor=west},
    ]
\addplot [draw=none, fill=color2] coordinates {(Coil20,93.2) (DNA,83.2) (HAR,89.3) (Mushrooms,96.7) (Protein,74.0) };
\addplot [draw=none, fill=color3] coordinates {(Coil20,84.3) (DNA,80.9) (HAR,81.4) (Mushrooms,72.2) (Protein,58.8) };
\end{axis}
\end{tikzpicture}
}
}

\vspace{-0.25cm}
\subfloat{
\hspace{3.1cm}
  \begin{tikzpicture}
    \draw[rounded corners=1pt] (-0.15,-0.15) rectangle (1.7,0.35);
    \draw[fill=color2] (0.0,0.0) rectangle (0.08,0.21);
    \node at (0.45,0.09) {\scalebox{0.7}{\texttt{IID}}};
    \draw[fill=color3] (0.9,0.0) rectangle (0.98,0.21);
    \node at (1.35,0.09) {\scalebox{0.75}{\texttt{SSB}}};
  \end{tikzpicture}
}
\caption{Test accuracies of the different baselines on $5$ datasets. Full results are in Appendix~\ref{app:app_failure_self}.}
\label{fig:drop_ssb}
\end{figure}
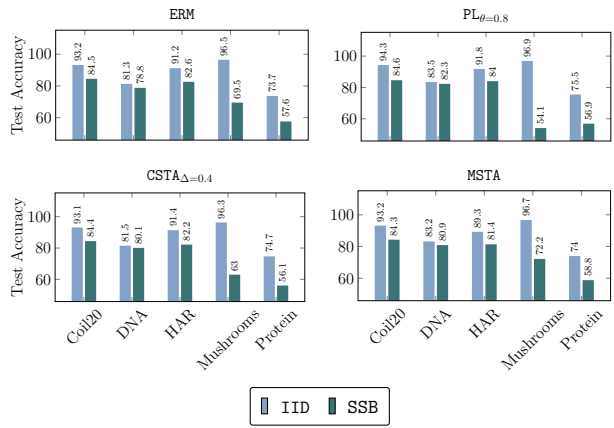

\begin{figure*}[!t]
    \centering
    \includegraphics[width=\textwidth]{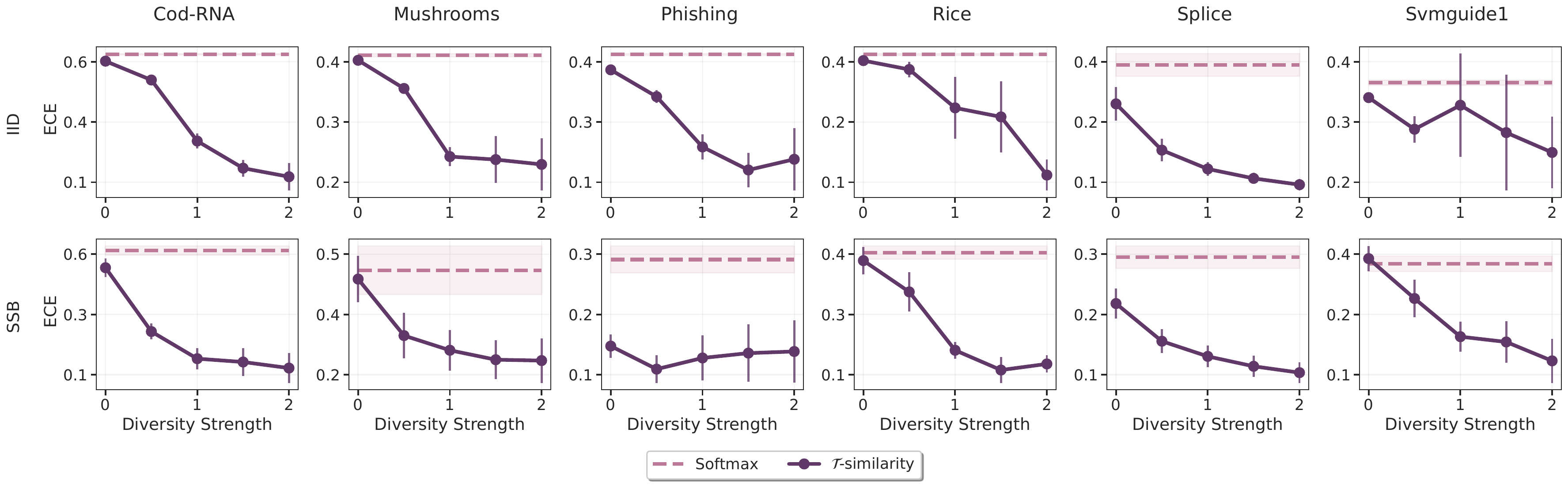}
    \caption{Increasing the diversity leads to a better-calibrated classifier in both \texttt{IID} and \texttt{SSB} settings.}
    \label{fig:ECE}
\end{figure*}

\subsection{Ensemble diversity provides a calibrated confidence measure} 
In this section, we train the model on the labeled set and then compute $\mathcal{T}$-similarity on the unlabeled data to verify its advertised behavior both in confidence estimation and in calibration of the final model.

\paragraph{Correcting the overconfidence of the $\texttt{softmax}$.} We plot the distributions of the confidence values for both \texttt{softmax} and $\mathcal{T}$-similarity on accurate (Correct Prediction) and incorrect predictions (Wrong Prediction). We display the plots obtained on \texttt{Mnist} in Figure~\ref{fig:calibrated_uncertainty_mnist} (results for other datasets given in Appendix~\ref{app:calibration}). Our conclusion here is two-fold: \textbf{(1)} \texttt{softmax} is overconfident both in \texttt{IID} and \texttt{SSB} settings (high confidence leads to the highest error rate), while $\mathcal{T}$-similarity confidence is low when the model makes most of its mistakes and is confident when it reduces them to 0; \textbf{(2)} \texttt{SSB} setting degrades the confidence estimation with \texttt{softmax} degrades even further while $\mathcal{T}$-similarity remains robust to such a distribution shift. This vividly highlights that $\mathcal{T}$-similarity possesses the desired properties and behaves as expected.

\def\figlength{0.48}
\begin{figure}[!h]
\centering
\subfloat[\texttt{IID}]{\includegraphics[width=\figlength\textwidth]{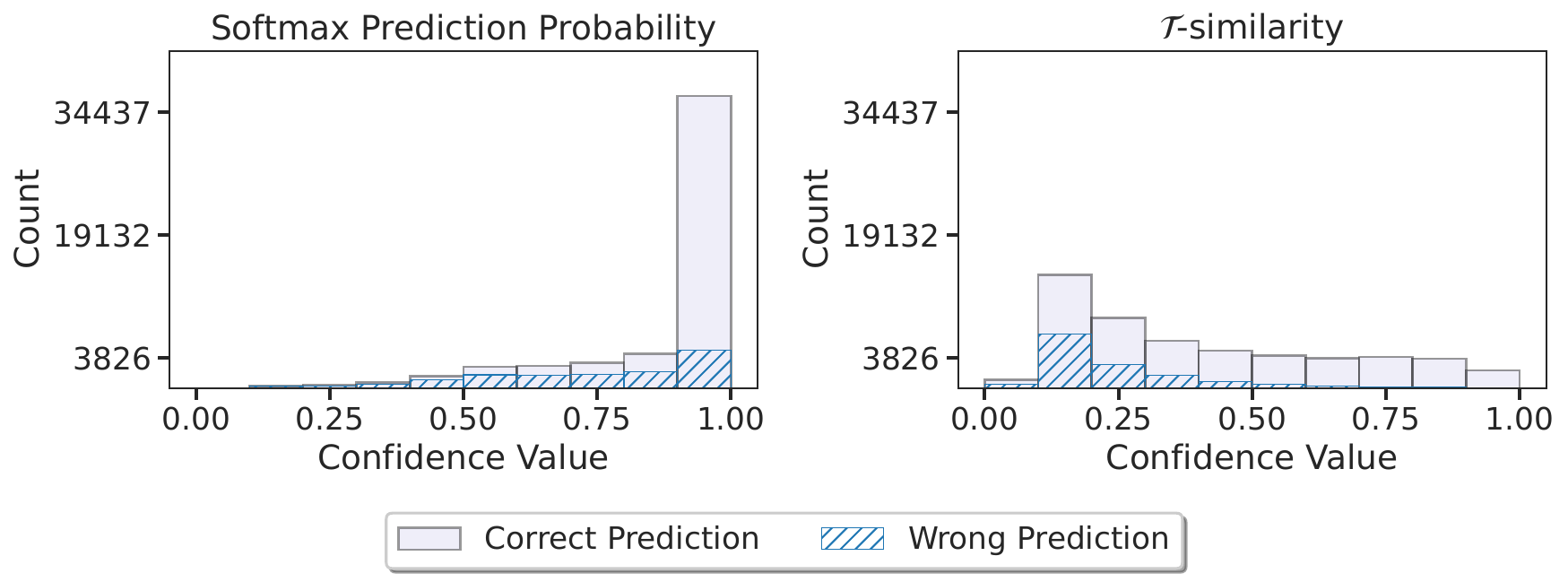}} 
\label{fig:calibrated_similarity_mnist_balanced} \qquad
\subfloat[\texttt{SSB}]{\includegraphics[width=\figlength\textwidth]{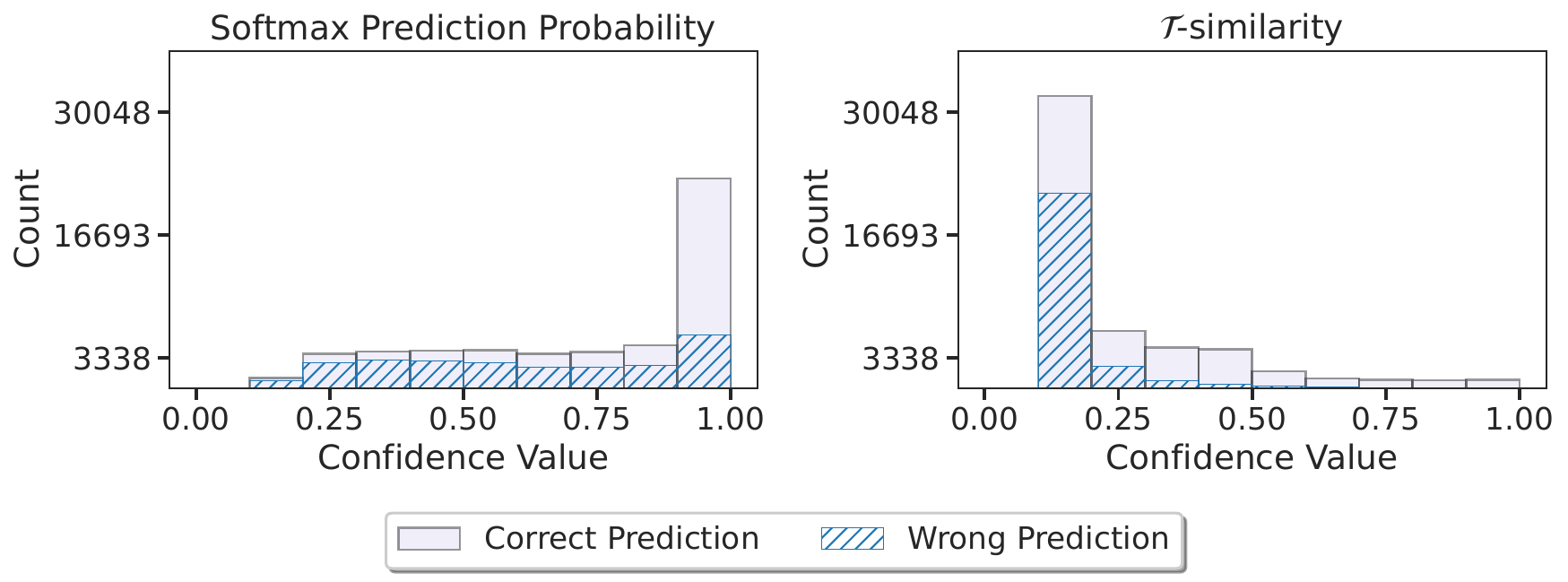}}
\label{fig:calibrated_similarity_mnist_pca} 
\caption{Softmax classifier is overconfident for both correctly and wrongly predicted samples. $s_\mathcal{T}$ assigns lower confidence to samples that are likely to be misclassified both in \texttt{IID} and \texttt{SSB} settings.}
 \label{fig:calibrated_uncertainty_mnist}
\end{figure}

\paragraph{Higher diversity improves calibration.} In Figure~\ref{fig:ECE}, we compare the Expected Calibration Error \citep{ece} obtained on the unlabeled set with the \texttt{softmax} and $\mathcal{T}$-similarity as a function of a varying regularization strength $\gamma$. We can see that when no diversity is imposed, i.e., $\gamma\!=\!0$, the ECE obtained with both confidence measures is comparable. However, for any positive value of $\gamma$, the calibration error becomes smaller and decreases with $\gamma$ in most of the cases considered. This is equally true for both \texttt{IID} and \texttt{SSB} settings and backs up our claim about the robustness of $\mathcal{T}$-similarity to the distribution shift. 

We now move to the evaluation of $\mathcal{T}$-similarity when used in self-training within the \texttt{SSL+SSB} paradigm.

\setlength{\tabcolsep}{0.4em}
\begin{table*}[!t] 
\centering
\caption{Classification performance of the different baselines on the datasets described in Table~\ref{tab:dataset_description} when labeling is done with \texttt{SSB}. We display the average and the standard deviation of the test accuracy (both in \%). The \texttt{softmax} corresponds to the usual self-training which uses the \texttt{softmax} prediction probability as a confidence estimate while the $\mathcal{T}$-similarity corresponds to our proposed method in Algorithm~\ref{alg:self_training_algorithm}. For each baseline, the best result between \texttt{softmax} and $\mathcal{T}$-similarity is in \textbf{bold}.}
\label{tab:comparative_performance_pca_bias}
\scalebox{0.85}{
\begin{tabular}{l||c||cc||cc||cc}
\toprule
\multicolumn{1}{c}{\multirow{2}{*}{Dataset}} & \multicolumn{1}{c}{\multirow{2}{*}{\texttt{ERM}}} & \multicolumn{2}{c}{$\texttt{PL}_{\theta=0.8}$} & \multicolumn{2}{c}{$\texttt{CSTA}_{\Delta=0.4}$}  & \multicolumn{2}{c}{\texttt{MSTA}}\\ 
\cmidrule(r{10pt}l{5pt}){3-4} \cmidrule(r{10pt}l{5pt}){5-6} \cmidrule(r{10pt}l{5pt}){7-8}
\multicolumn{1}{c}{} & \multicolumn{1}{c}{} & \multicolumn{1}{c}{\texttt{softmax}} & \multicolumn{1}{c}{$\mathcal{T}$-similarity} & \multicolumn{1}{c}{\texttt{softmax}} & \multicolumn{1}{c}{$\mathcal{T}$-similarity} & \multicolumn{1}{c}{\texttt{softmax}} & \multicolumn{1}{c}{$\mathcal{T}$-similarity} \\
\midrule
\texttt{Cod-RNA} & 	$74.51 \pm 8.86$ & $74.75 \pm 8.14$ & $\mathbf{80.06 \pm 3.55}$ & $73.39 \pm 7.36$ & $\mathbf{78.39 \pm 4.66}$ & $75.28 \pm 8.79$ & $\mathbf{76.88 \pm 7.67}$ \\
\texttt{COIL-20} & $84.54 \pm 2.19$ & $\mathbf{84.69 \pm 3.56}$ & $84.57 \pm 2.85$ & $84.38 \pm 3.05$ & $\mathbf{84.57 \pm 3.16}$ & $\mathbf{84.32 \pm 2.34}$ & $84.07 \pm 2.85$ \\
\texttt{Digits} & $75.68 \pm 4.59$ & $\mathbf{80.47 \pm 3.8}$ & $78.2 \pm 3.34$ & $78.4 \pm 3.28$ & $\mathbf{79.14 \pm 3.5}$ & $78.02 \pm 5.15$ & $\mathbf{79.8 \pm 5.92}$ \\
\texttt{DNA} & $78.82 \pm 2.31$ & $\mathbf{80.29 \pm 2.24}$ & $79.06 \pm 2.31$ & $80.12 \pm 2.08$ & $\mathbf{80.76 \pm 2.24}$ & $80.89 \pm 2.64$ & $\mathbf{84.09 \pm 1.7}$ \\
\texttt{DryBean} & $64.6 \pm 3.89$ & $\mathbf{65.6 \pm 4.18}$ & $61.55 \pm 4.91$ & $\mathbf{64.91 \pm 3.72}$ & $64.6 \pm 3.53$ & $66.24 \pm 4.31$ & $\mathbf{67.0 \pm 3.96}$ \\
\texttt{HAR} & 	$82.57 \pm 1.96$ & $82.87 \pm 3.02$ & $\mathbf{83.12 \pm 2.27}$ & $82.19 \pm 2.61$ & $\mathbf{83.53 \pm 3.77}$ & $\mathbf{81.35 \pm 2.54}$ & $81.16 \pm 1.63$ \\
\texttt{Mnist} & $50.74 \pm 2.25$ & $51.08 \pm 2.55$ & $\mathbf{52.69 \pm 2.42}$ & $51.7 \pm 3.52$ & $\mathbf{54.26 \pm 1.82}$ & $51.6 \pm 2.58$ & $\mathbf{54.18 \pm 2.34}$ \\
\texttt{Mushrooms} & $69.45 \pm 7.29$ & $59.53 \pm 10.46$ & $\mathbf{71.36 \pm 6.63}$ & $62.98 \pm 7.25$ & $\mathbf{77.55 \pm 7.65}$ & $72.16 \pm 7.59$ & $\mathbf{76.16 \pm 13.04}$ \\
\texttt{Phishing} & $67.42 \pm 3.55$ & 	$66.08 \pm 5.66$ & $\mathbf{77.41 \pm 3.93}$ & $66.88 \pm 5.64$ & $\mathbf{76.17 \pm 8.58}$ & $69.48 \pm 4.37$ & $\mathbf{75.83 \pm 7.52}$ \\
\texttt{Protein} & $57.57 \pm 6.33$ & 	$57.45 \pm 6.36$& $\mathbf{57.61 \pm 6.23}$ & $56.09 \pm 5.61$ & $\mathbf{57.74 \pm 7.8}$ & $58.81 \pm 6.54$ & $\mathbf{59.88 \pm 6.29}$ \\
\texttt{Rice} & $79.19 \pm 5.12$ & $80.54 \pm 4.31$ & $\mathbf{81.1 \pm 4.28}$ & $79.88 \pm 4.48$ & $\mathbf{81.56 \pm 3.61}$ & $80.35 \pm 4.89$ & $\mathbf{82.63 \pm 5.63}$ \\
\texttt{Splice} & $66.13 \pm 4.47$ & $67.14 \pm 2.62$ & $\mathbf{67.45 \pm 2.53}$ & $67.28 \pm 2.07$ & $\mathbf{68.05 \pm 2.17}$ & $66.08 \pm 4.98$ & $\mathbf{66.32 \pm 4.73}$ \\
\texttt{Svmguide1} & $70.89 \pm 10.98$ & $70.35 \pm 11.74$ & $\mathbf{81.07 \pm 5.39}$ & $69.84 \pm 11.06$ & $\mathbf{74.46 \pm 7.23}$ & $71.04 \pm 11.11$ & $\mathbf{73.13 \pm 8.82}$ \\
\bottomrule
\end{tabular}
}
\end{table*}
\subsection{Robust self-training with the $\mathcal{T}$-similarity}
\paragraph{Improved performance under \texttt{SSB}.} We now compare the \texttt{softmax} prediction probability and our $\mathcal{T}$-similarity under \texttt{SSB}. For each dataset, we display the average and the standard deviation of the test accuracy (both in $\%$) in Table~\ref{tab:comparative_performance_pca_bias}. For $\texttt{CSTA}_{\Delta=0.4}$ and \texttt{MSTA}, the $\mathcal{T}$-similarity leads to a substantial improvement on $11$ of the $13$ datasets. For $\texttt{PL}_{\theta=0.8}$, it improves the baseline on $8$ of the $13$ datasets. The obtained improvement is significant on \texttt{Mushrooms} and \texttt{Phishing}, where the \texttt{softmax} performs worse than \texttt{ERM}, the supervised baseline. For a fair evaluation of our method, we perform the same experiment as above when the labeling is done with \texttt{IID} in Appendix~\ref{app:comparative_performance_no_bias}. The obtained performance of our method is close to that of \texttt{softmax} suggesting that diversity can be safely promoted without any assumptions on the statistical relationship between the labeled and unlabeled data.

\paragraph{Sensitivity analysis.}  In Table~\ref{tab:comparative_performance_pca_bias} and  Table~\ref{tab:comparative_performance_no_bias}, we display the results of the pseudo-labeling policy $\texttt{PL}_{\theta}$ with $\theta=0.8$. To show that our obtained improvements are robust to the choice of confidence threshold $\theta$, we study the performance of self-training with the \texttt{softmax} and the $\mathcal{T}$-similarity on \texttt{Mushrooms}, \texttt{Phishing} and \texttt{Svmguide1}, under both the \texttt{IID} and \texttt{SSB}  settings, when $\theta$ varies in $\{0.7, 0.8, 0.9, 0.95\}$. We present the results in Figure~\ref{fig:ablation_study_conf_threshold} and observe that for \texttt{IID} setting, the choice of the confidence level is not very important for both baselines considered. However, in \texttt{SSB} setting, it appears safer to choose a lower confidence level for pseudo-labeling as in most cases it leads to the best performance. Finally, and in accordance with Table~\ref{tab:comparative_performance_pca_bias}, we note that our approach behaves much better under distribution shift compared to \texttt{softmax}.

\def\figlength{0.48}
\begin{figure}[!ht]
\centering
\subfloat[\texttt{IID}.]{\includegraphics[width=\figlength\textwidth]{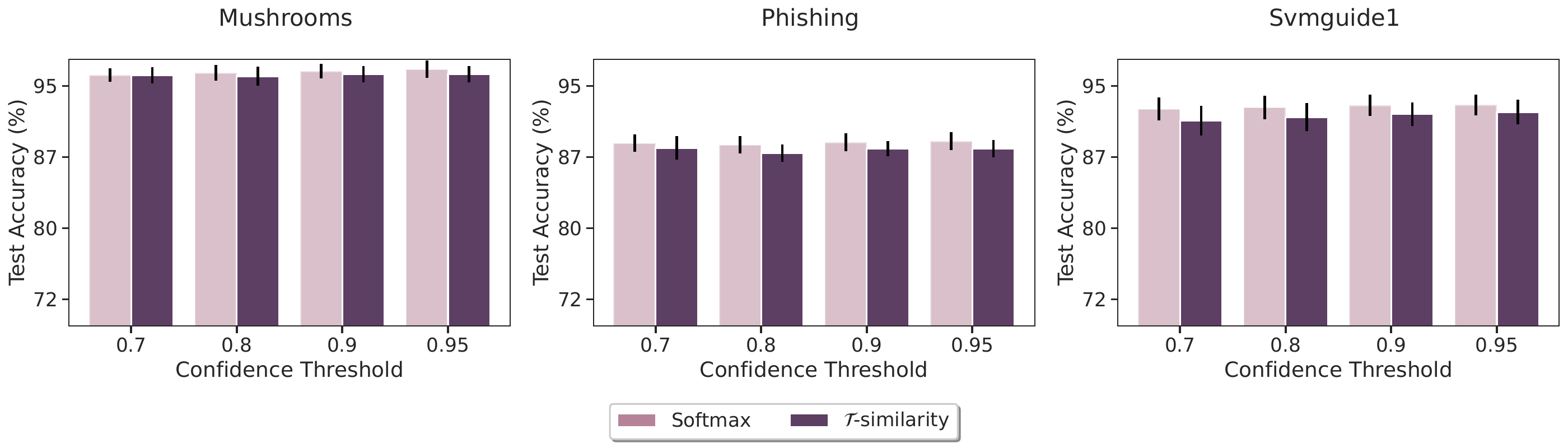}} 
\label{fig:ablation_study_conf_threshold_no_bias} \qquad
\subfloat[\texttt{SSB}.]{\includegraphics[width=\figlength\textwidth]{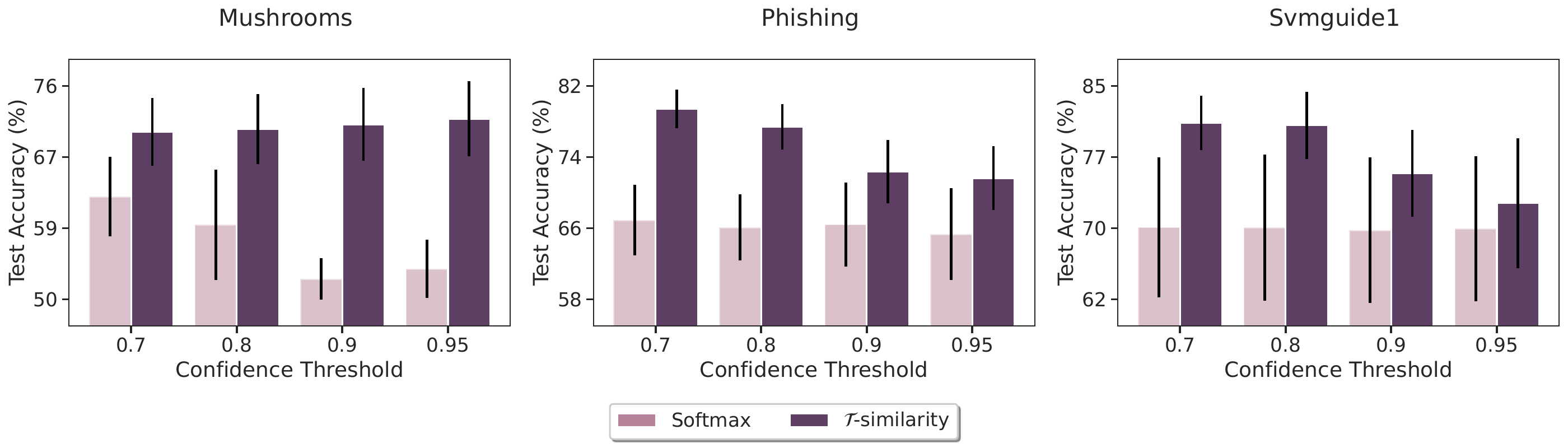}} 
\label{fig:ablation_study_conf_threshold_pca_bias} 
\caption{Ablation study on the confidence threshold on \texttt{Mushrooms} and \texttt{Phishing} and \texttt{Svmguide1}. We display the average test accuracy (\%) with a $95\%$ confidence interval.}
 \label{fig:ablation_study_conf_threshold}
\end{figure}

\paragraph{Additional results.} Due to the space constraints, we provide a study on the impact of the number of labeled examples $n_\ell$ in Appendix~\ref{app:ablation_study}. We also highlight the sensitivity of our method as a function of $\gamma$ with respect to the pseudo-labeling strategies, and the number of classifiers $M$ in Appendix~\ref{app:sensitivity_analysis}. 

\section{Conclusion}
In this paper, we studied the effect of the sample selection bias on the performance of self-training methods in the semi-supervised learning framework. We showed that the conventional choice of \texttt{softmax} as a confidence measure degrades their performance regardless of the choice of the pseudo-labeling policy. To overcome this problem, we proposed a new $\mathcal{T}$-similarity measure that assigns high confidence to those unlabeled examples for which an ensemble of diverse linear classifiers $\mathcal{T}$ agrees in its predictions. Firstly, we empirically showed that the proposed confidence measure improves all the considered self-training methods in the case of a biased labeling procedure. Secondly, we performed a theoretical analysis of the proposed similarity and found that the representation space plays an important role in the utility of the ensemble diversity. This suggests another direction of future work, where the representation could be jointly learned with the diverse ensemble as a part of a batch self-training architecture \citep{DST}. 
\clearpage
\section*{Aknowledgments}
The authors would like to thank Gabriel Peyr\'e for his insightful comments on early drafts of this paper, as well as Malik Tiomoko, and Aladin Virmaux for the fruitful discussions that led to this work. The authors thank the anonymous reviewers and meta-reviewers for their time and constructive feedback. This work was enabled thanks to open-source software such as Python~\citep{van1995python}, PyTorch~\citep{pytorch}, Scikit-learn~\citep{scikit-learn} and Matplotlib~\citep{hunter2007matplotlib}.

\bibliography{bibliography}
\bibliographystyle{apalike}

\section*{Checklist}
 \begin{enumerate}

 \item For all models and algorithms presented, check if you include:
 \begin{enumerate}
   \item A clear description of the mathematical setting, assumptions, algorithm, and/or model. [Yes]
   \item An analysis of the properties and complexity (time, space, sample size) of any algorithm. [Not Applicable]
   \item (Optional) Anonymized source code, with specification of all dependencies, including external libraries. [No]
 \end{enumerate}

 \item For any theoretical claim, check if you include:
 \begin{enumerate}
   \item Statements of the full set of assumptions of all theoretical results. [Yes]
   \item Complete proofs of all theoretical results. [Yes]
   \item Clear explanations of any assumptions. [Yes]     
 \end{enumerate}

 \item For all figures and tables that present empirical results, check if you include:
 \begin{enumerate}
   \item The code, data, and instructions needed to reproduce the main experimental results (either in the supplemental material or as a URL). [Yes]
   \item All the training details (e.g., data splits, hyperparameters, how they were chosen). [Yes]
         \item A clear definition of the specific measure or statistics and error bars (e.g., with respect to the random seed after running experiments multiple times). [Yes]
         \item A description of the computing infrastructure used. (e.g., type of GPUs, internal cluster, or cloud provider). [No]
 \end{enumerate}

 \item If you are using existing assets (e.g., code, data, models) or curating/releasing new assets, check if you include:
 \begin{enumerate}
   \item Citations of the creator If your work uses existing assets. [Yes]
   \item The license information of the assets, if applicable. [Not Applicable]
   \item New assets either in the supplemental material or as a URL, if applicable. [Not Applicable]
   \item Information about consent from data providers/curators. [Not Applicable]
   \item Discussion of sensible content if applicable, e.g., personally identifiable information or offensive content. [Not Applicable]
 \end{enumerate}

 \item If you used crowdsourcing or conducted research with human subjects, check if you include:
 \begin{enumerate}
   \item The full text of instructions given to participants and screenshots. [Not Applicable]
   \item Descriptions of potential participant risks, with links to Institutional Review Board (IRB) approvals if applicable. [Not Applicable]
   \item The estimated hourly wage paid to participants and the total amount spent on participant compensation. [Not Applicable]
 \end{enumerate}

 \end{enumerate}

\clearpage
\appendix
\renewcommand*{\proofname}{Proof}
\onecolumn

\aistatstitle{Leveraging Ensemble Diversity for Robust Self-Training in the Presence of Sample Selection Bias: Supplementary Materials}

\paragraph{Roadmap.} We provide related work in Section~\ref{app:related_work}, the experimental setup in Section~\ref{app:exp_setup}, additional experiments in Section~\ref{app:add_exp}, the ablation study and sensitivity analysis in Section~\ref{app:ablation_study_sensitivity} and the proofs in Section~\ref{app:proofs}.

\addtocontents{toc}{\protect\setcounter{tocdepth}{2}}

\renewcommand*\contentsname{\Large Table of Contents}

\tableofcontents
\clearpage

\section{Related work}
\label{app:related_work}
\subsection{Wrapper self-training}
\label{app:self_training_algorithm}
In Algorithm~\ref{alg:self_training_algorithm}, we present a general pseudo-code for wrapper self-training algorithms that are based on different confidence estimators $\phi_h$ and pseudo-labeling policies $\psi$. For neural networks, the conventional choice of $\phi_h$ is the \texttt{softmax} function, so $\hat{\mathbf{P}}_u\in(\Delta_C)^{n_j}$.

\begin{algorithm}[H] 
\caption{Wrapper Self-Training}
\label{alg:self_training_algorithm}
    \textbf{Input:} Labeled training set $(\mathbf{X}_l, \mathbf{y}_\ell)$, unlabeled training set $\mathbf{X}_u$, base classifier $h$ \\
    \textbf{Parameters:} Pseudo-labeling policy $\psi$, confidence estimator $\phi_h$ based on $h$, maximum number of iterations $N$ \\
    \textbf{Initialization:} Iteration $t=1$ \\
    \While {$t \leq N$ \text{ and } $\mathbf{X}_u \neq \emptyset$}{
        \textbf{1. Training} \\
        \hspace{.5cm} (Re-)Train the base classifier $h$ on $(\mathbf{X}_\ell, \mathbf{y}_\ell)$\\
        \textbf{2. Confidence estimation} \\
        \hspace{.5cm} 
        $\hat{\mathbf{P}}_u = \{\phi_h(\mathbf{x})\}_{\mathbf{x} \in \mathbf{X}_u}$ \\
        \textbf{3. Pseudo-labeling} \\
        \hspace{.5cm} Determine subset $\mathbf{X}_{pl}\subset\mathbf{X}_u$ for pseudo-labeling $\mathbf{X}_{pl} = \psi(\mathbf{X}_u,\hat{\mathbf{P}}_u, t)$\\
        \hspace{.5cm} Compute pseudo-labels $\hat{\mathbf{y}}_{pl} = \{\argmax h(\mathbf{x})\}_{\mathbf{x} \in \mathbf{X}_{pl}}$ \\
        \textbf{4. Update of training sets} \\
        \hspace{.5cm} $(\mathbf{X}_\ell, \mathbf{y}_\ell) \gets (\mathbf{X}_\ell, \mathbf{y}_\ell) \cup (\mathbf{X}_{pl}, \hat{\mathbf{y}}_{pl})$ \\
        \hspace{.5cm} $\mathbf{X}_u \gets \mathbf{X}_u \setminus \mathbf{X}_{pl}$\\
        \hspace{.5cm} $t \gets t+1$
    }
    \textbf{Output} Final classifier $h$ 
\end{algorithm}

\subsection{Self-training policies}
\label{app:policies}

\paragraph{Fixed Threshold \texttt{PL}$_{\theta}$.} It is the most standard policy \citep{Yarowsky:1995,lee_pl,Sohn:2020:FixMatch} that fixes the threshold to a certain value $\theta$. In the case of $\phi_h\equiv\texttt{softmax}$,  we have $[\hat{\mathbf{P}}_u]_{j}\in\Delta_C$, and then the pseudo-labeling policy outputs: 
\begin{align}
\label{eq:pl-policy}
    \psi(\mathbf{X}_u,\hat{\mathbf{P}}_u, t)=\{\mathbf{x}_j|\max_c[\hat{\mathbf{P}}_u]_{j,c} > \theta\}_{j=1}^{n_u}.
\end{align}

\paragraph{Curriculum \texttt{CSTA}$_{\Delta}$.} We follow the implementation of \cite{cascante2021curriculum} that finds a new threshold $\theta^{(t)}$ at every iteration $t$ as the $(1-t\cdot\Delta)$-th quantile of the distribution of the prediction confidence $\{\max_c[\hat{\mathbf{P}}_u]_{j,c}\}_{j=1}^{n_u}$, that is assumed to follow a Pareto distribution. Then, the final policy is Eq.~\eqref{eq:pl-policy}, where $\theta$ is replaced by $\theta^{(t)}$.

\paragraph{Transductive policy \texttt{MSTA}}
This policy is based on the upper-bound of the transductive error on the unlabeled examples that have a confidence score larger than a threshold $\bm{\theta}$, denoted by $R_{u,\geq\bm{\theta}}$ \citep{amini2008transductive}. We use the multi-class implementation of \cite{Feofanov:2019} that employs a threshold vector $\bm{\theta}^{(t)}=\left(\theta^{(t)}_c\right)_{c=1}^C$, where $\theta^{(t)}_c$ is a threshold for class $c$ at iteration $t$:
\begin{align*}
    \psi(\mathbf{X}_u,\hat{\mathbf{P}}_u, t)=\left\{\mathbf{x}_j|[\hat{\mathbf{P}}_u]_{j,\hat{y}_j} > \theta^{(t)}_{\hat{y}_j}\right\}_{j=1}^{n_u},
\end{align*}
where $\hat{y}_j=\argmax_c [\hat{\mathbf{P}}_u]_{j,c}$ is the prediction for $\mathbf{x}_j$.
The threshold is found by solving the following minimization problem:
\begin{align*}
\bm{\theta}^{(t)}=\argmin_{\bm{\theta}\in[0,1]^C}\frac{R_{u,\geq\bm{\theta}}}{(1/n_u)\sum_{j=1}^{n_u} \mathbb{I}({[\hat{\mathbf{P}}_u]_{j,\hat{y}_j} > \theta_{\hat{y}_j}})},
\end{align*}
where $\mathbb{I}$ denotes the indicator function. Thus, the threshold at each iteration is chosen by minimizing the ratio between the upper bound on the error and the number of examples to be pseudo-labeled.

\section{Experimental Setup}
\label{app:exp_setup}
\subsection{Datasets}
\label{app:datasets}
In all experiments, the only pre-processing step is to standardize the features. Table~\ref{tab:dataset_description} sums up the characteristics of the datasets used in our experiments and the corresponding values of hyperparameter $r$ used in the \texttt{SSB} labeling procedure (Algorithm~\ref{alg:pca_labeling_procedure}). We considered $13$ publicly available SSL datasets with various data modalities:
\begin{itemize}
\item[-] Biological data for \texttt{Cod-RNA} \citep{chang2011libsvm}, \texttt{DNA} \citep{chang2011libsvm}, \texttt{Protein} \citep{dua_2019}, \texttt{Splice} \citep{dua_2019}
\item[-] Images for \texttt{COIL-20} \citep{Nene1996}, \texttt{Digits} \citep{scikit-learn}, \texttt{Mnist} \citep{726791}
\item[-] Tabular data for \texttt{DryBean} \citep{dua_2019}, \texttt{Mushrooms} \citep{dua_2019}, \texttt{Phishing} \citep{chang2011libsvm}, \texttt{Rice} \citep{dua_2019}, \texttt{Svmguide1} \citep{chang2011libsvm}
\item[-] Time series for \texttt{HAR} \citep{dua_2019}
\end{itemize}

\begin{table}[htbp]
    \centering
    \caption{Characteristics of the datasets and corresponding values of hyperparameter $r$.}
    \label{tab:dataset_description}
    \begin{tabular}{l|ccccc}
        \hline
         Dataset & Size & $\#$ of lab. examples $n_\ell$ & Dimension $d$ & $\#$ classes $C$ & \texttt{SSB} hyperparameter $r$\\
         \hline
         \texttt{Cod-RNA} & $59535$ & $99$ & $8$ & $2$ & $2$\\
         \texttt{COIL-20} & $1440$ & $200$ & $1024$ & $20$ & $0.33$\\
         \texttt{Digits} & $1797$ & $99$ & $64$ & $10$ & $0.5$\\
         \texttt{DNA}& $3186$ & $149$ & $180$ & $6$ & 25\\
         \texttt{DryBean}& $13543$ & $104$ & $16$ & $7$ & $2$\\
         \texttt{HAR}& $10299$ & $299$ & $561$ & $6$ & $0.33$\\
         \texttt{Mnist} & $70000$ & $100$ & $784$ & $10$ & $0.33$\\
         \texttt{Mushrooms}& $8124$ & $79$ & $112$ & $2$ & $2$\\
         \texttt{Phishing}& $11055$ & $99$ & $68$ & $2$ & $2$\\
         \texttt{Protein}& $1080$ & $80$ & $77$ & $8$ & $0.6$\\
         \texttt{Rice}& $3810$ & $29$ & $7$ & $2$ & 2\\
         \texttt{Splice} & $3175$ &  $39$ & $60$ & $2$ & $2$\\
         \texttt{Svgmguide1}& $3089$ & $39$ & $4$ & $2$ & $2$\\
         \hline
    \end{tabular}
\end{table}

\subsection{More details on the labeling procedure}
\label{app:labeling_procedure}
\paragraph{Labeling procedure.} A classical strategy in SSL benchmarks is to use an i.i.d. sampling to select the same number of labeled data in each class. It can be achieved by applying i.i.d. sampling in a class-wise manner. In our work, we refer to this labeling procedure as \texttt{IID}. Inspired by \citet{NIPS2006_a2186aa7, Zadrozny:2004}, we consider another strategy that simulates sample selection bias to select training examples in each class. We make sure that the original class proportion is preserved. In our work, we refer to this labeling procedure as \texttt{SSB}. The pseudo-code of the \texttt{SSB} labeling procedure is outlined in Algorithm~\ref{alg:pca_labeling_procedure} (values of $r$ are given in the last column of Table~\ref{tab:dataset_description}).

\paragraph{Visualization.} As a picture is worth a thousand words, we illustrate the sample selection bias on the \texttt{Mushrooms} dataset \citep{dua_2019}. We select $80$ labeled examples out of the $6093$ training examples with the \texttt{IID} and the \texttt{SSB} procedures. Inspired by \citep{adapt}, we plot in the first row of Figure~\ref{fig:visu_bias_mushrooms} the distribution of the projection values on the first principal component (PC1) of the labeled training set (\textcolor{rightblue}{blue}) and of the whole training set (\textcolor{rightorange}{orange}). Contrary to the \texttt{IID} sampling, we can see that the \texttt{SSB} sampling injects a clear bias between distributions. We also visualize in the second row of Figure~\ref{fig:visu_bias_mushrooms} the projection values on the first two principal components (PC1 and PC2) of the labeled training examples (\textcolor{rightblue}{blue}) and of all the training samples (\textcolor{rightorange}{orange}).  We can see that the \texttt{IID} procedure samples in all regions of the space while the \texttt{SSB} concentrates in specific areas. 
\begin{algorithm}[!h]
    \caption{\texttt{SSB} Labeling Procedure}
    \label{alg:pca_labeling_procedure}
    \textbf{Input:} Training examples $\{(\mathbf{x}_1, y_1), \dots, (\mathbf{x}_n, y_m)\}$, class proportions $\{p_1, \dots, p_C\}$ \\
    \textbf{Parameters:} number of training examples to label $n_\ell$, hyperparameter $r > 0$. \\
    \textbf{Initialize} labeled training set $(\mathbf{X}_\ell, \mathbf{y}_\ell) = \emptyset$ \\
    \For{$c \in [C]$}{
    $\mathcal{T}_c = \{(\mathbf{x}_i, y_i) | y_i = c\}$ set of training examples of class $c$\\
    \textbf{Compute projection values} \\
    Apply PCA on features of $\mathcal{T}_c$ \\
    Recover for each $\mathbf{x}_i$ the projection value on the first principal component $\mathrm{proj}_1(\mathbf{x}_i)$ \\
    Compute $\beta = \sum_{\mathbf{x}_i} \exp(r \times \lvert \mathrm{proj}_1(\mathbf{x}_i) \rvert)$ \\
    \textbf{Draw without replacement in $\mathcal{T}_k$} \\
    \While {$\lvert \mathbf{X}_\ell \rvert < p_k n_\ell$}{
    Draw $\mathbf{x}_i$ with probability $P(s=1|\mathbf{x}_i, y_i=c) = \frac{1}{\beta} \exp(r \times \lvert \mathrm{proj}_1(\mathbf{x}_i) \rvert)$ \\
    $\mathbf{X}_\ell \gets \mathbf{X}_\ell \cup \{(\mathbf{x}_i, y_i)\}$
    }
    }
    \textbf{Create the unlabeled training set} \\
    $\mathbf{X}_u = \{\mathbf{x}_j | \mathbf{x}_j\notin \mathbf{X}_\ell\}$ \\
    \textbf{Output} labeled training set $(\mathbf{X}_\ell, \mathbf{y}_\ell)$, unlabeled training set $\mathbf{X}_u$
\end{algorithm}

\subsection{Baselines}
\label{app:baselines}

We performed experiments with three different pseudo-labeling policies: \texttt{PL}, \texttt{CSTA}, and \texttt{MSTA}. For a fair comparison of the methods, we tested manually different values of hyperparameters. We kept those that gave good results on average, namely, $\theta=0.8$ for \texttt{PL} and $\Delta=0.4$ for \texttt{CSTA}. The maximum number of self-training iterations $N$ is set to $5$. 

\paragraph{Implementation of \texttt{MSTA}} In the original implementation of \citep{Feofanov:2019}, the authors estimate the posterior probability $P(y|\mathbf{x})$ by the votes of the majority vote classifier. To apply the policy \texttt{MSTA} in Algorithm~\ref{alg:self_training_algorithm}, the posterior probabilities in the upper-bound expression has to be estimated, and they have empirically shown that the prediction probabilities given by the supervised baseline allow to have a good proxy for the bound \citep{feofanov2021multi}. We have tested several ways to estimate these probabilities for our network (majority vote of the ensemble, using $\mathcal{T}$-similarities), and the use of predicted probabilities by the prediction head gives the most stable results.

\section{Additional Experiments}
\label{app:add_exp}
In this section, we provide additional experiments to showcase the effectiveness of our method.

\subsection{Failure cases of self-training}
\label{app:app_failure_self}
The classification performance of the self-training algorithms is shown in Table~\ref{tab:bias_no_bias_comparative_performance}. For each method, on all datasets, we observe a huge drop in performance when sample selection bias is applied (labeling with \texttt{SSB}) compared to when it is not (labeling with \texttt{IID}). Self-training even turns out to be harmful in some situations, for instance on \texttt{Mushrooms}, \texttt{Phishing}, or \texttt{Protein}. These results confirm the fact that a biased confidence measure, due to the sample selection bias, diminishes the quality of the pseudo-labeling and in some cases, the benefit of using unlabeled data becomes a disadvantage. It should be noted that the performance of the supervised baselines \texttt{ERM} also declines when sample selection bias is applied. 
\setlength{\tabcolsep}{0.4em}
\begin{table}[!h] 
\centering
    \caption{Classification performance of the different baselines on the datasets described in Table~\ref{tab:dataset_description}. We display the average and the standard deviation of the test accuracy (both in \%) over the $9$ trials. For each baseline, the best result between \texttt{IID} and \texttt{SSB} is in \textbf{bold}.}
    \label{tab:bias_no_bias_comparative_performance}
\scalebox{0.8}{
    \begin{tabular}{lccccc}
    \toprule
    \multirow{2}{*}{Dataset} & \multirow{2}{*}{Bias} & \multicolumn{4}{c}{Baselines} \\ 
    \cmidrule(r{10pt}l{5pt}){3-6}
     & & \texttt{ERM} & $\texttt{PL}_{\theta=0.8}$ & $\texttt{CSTA}_{\Delta=0.4}$ & \texttt{MSTA} \\
    \midrule
\rowcolor{celadon!25}  \multirow{2}{*}{\cellcolor{white} \texttt{Cod-RNA}} & \texttt{IID} & $89.28 \pm 2.13$ & $89.91 \pm 2.03$ & $89.09 \pm 2.37$ & $89.89 \pm 1.89$ \\
 & \texttt{SSB} & $74.51 \pm 8.86$ & $74.21 \pm 7.76$ & $73.39 \pm 7.36$ & $75.28 \pm 8.79$ \\
 \midrule
\rowcolor{celadon!25} \multirow{2}{*}{\cellcolor{white} \texttt{COIL-20}} & \texttt{IID} & $93.18 \pm 1.5$ & 	$94.32 \pm 1.13$ & 	$93.09 \pm 1.73$ &  $93.21 \pm 1.57$  \\
 & \texttt{SSB} & $84.54 \pm 2.19$ & $84.6 \pm 3.86$ & $84.38 \pm 3.05$ & $84.32 \pm 2.34$  \\
  \midrule
\rowcolor{celadon!25} \multirow{2}{*}{\cellcolor{white} \texttt{Digits}} & \texttt{IID} & $81.38 \pm 2.45$ & $84.27 \pm 2.98$ & $81.78 \pm 2.51$ & $83.04 \pm 2.13$  \\
 & \texttt{SSB} & $75.68 \pm 4.59$ & $80.86 \pm 4.11$ & $78.4 \pm 3.28$ & $78.02 \pm 5.15$ \\
  \midrule
\rowcolor{celadon!25} \multirow{2}{*}{\cellcolor{white} \texttt{DNA}} & \texttt{IID} & $81.28 \pm 2.27$ & $83.45 \pm 2.01$ & $81.54 \pm 2.44$ &  $83.19 \pm 1.96$ \\
 & \texttt{SSB} & $78.82 \pm 2.31$ & $82.28 \pm 2.5$ & $80.12 \pm 2.08$ & $80.89 \pm 2.64$ \\
 \midrule
\rowcolor{celadon!25} \multirow{2}{*}{\cellcolor{white} \texttt{DryBean}} & \texttt{IID} & $86.85 \pm 1.68$ & $88.02 \pm 1.49$ & $86.98 \pm 1.61$ & $87.72 \pm 1.54$ \\
& \texttt{SSB} & $64.6 \pm 3.89$ & $66.12 \pm 4.35$ & $64.91 \pm 3.72$ & $66.24 \pm 4.31$ \\
  \midrule
\rowcolor{celadon!25} \multirow{2}{*}{\cellcolor{white} \texttt{HAR}} & \texttt{IID} & $91.16 \pm 0.54$ & $91.82 \pm 0.4$ & $91.36 \pm 0.24$ & $89.29 \pm 1.24$ \\
 & \texttt{SSB} & $82.57 \pm 1.96$ & $84.02 \pm 2.61$  & $82.19 \pm 2.61$  & $81.35 \pm 2.54$  \\
  \midrule
\rowcolor{celadon!25} \multirow{2}{*}{\cellcolor{white} \texttt{Mnist}} & \texttt{IID} & $73.98 \pm 1.46$ & $75.37 \pm 1.57$ & $75.24 \pm 1.48$ & $74.6 \pm 1.76$ \\
& \texttt{SSB} & $50.74 \pm 2.25$ & $51.07 \pm 2.2$  & $51.7 \pm 3.52$ & $51.6 \pm 2.58$ \\
 \midrule
\rowcolor{celadon!25} \multirow{2}{*}{\cellcolor{white} \texttt{Mushrooms}} & \texttt{IID} & $96.48 \pm 1.57$ & $96.94 \pm 1.4$ & $96.3 \pm 1.32$ & $96.68 \pm 1.31$ \\
 & \texttt{SSB} & $69.45 \pm 7.29$ & $54.08 \pm 5.56$ & $62.98 \pm 7.25$ & $72.16 \pm 7.59$ \\
  \midrule
\rowcolor{celadon!25} \multirow{2}{*}{\cellcolor{white} \texttt{Phishing}} & \texttt{IID} & $88.51 \pm 1.51$ & $89.41 \pm 1.44$ & $88.96 \pm 1.37$ & $88.82 \pm 1.7$ \\
 & \texttt{SSB} & $67.42 \pm 3.55$ & $65.34 \pm 7.86$ & $66.88 \pm 5.64$ & $69.48 \pm 4.37$ \\
  \midrule
\rowcolor{celadon!25} \multirow{2}{*}{\cellcolor{white} \texttt{Protein}} & \texttt{IID} & $73.74 \pm 4.78$ & $75.51 \pm 2.83$ & $74.73 \pm 3.01$ & $73.99 \pm 5.6$ \\
& \texttt{SSB} & $57.57 \pm 6.33$ & $56.87 \pm 5.79$ & $56.09 \pm 5.61$ & $58.81 \pm 6.54$ \\
 \midrule
\rowcolor{celadon!25} \multirow{2}{*}{\cellcolor{white} \texttt{Rice}} & \texttt{IID} & $88.24 \pm 3.63$ & $88.5 \pm 3.39$ & $88.15 \pm 3.46$ & $88.69 \pm 3.49$ \\
& \texttt{SSB} & $79.19 \pm 5.12$ & $80.87 \pm 4.43$ & $79.88 \pm 4.48$ & $80.35 \pm 4.89$ \\
 \midrule
\rowcolor{celadon!25} \multirow{2}{*}{\cellcolor{white} \texttt{Splice}} & \texttt{IID} & $69.09 \pm 4.09$ & $70.64 \pm 4.52$  & 	$70.46 \pm 4.32$ & $69.65 \pm 4.27$ \\
 & \texttt{SSB} & $66.13 \pm 4.47$ & $67.02 \pm 2.11$ & $67.28 \pm 2.07$ & $66.08 \pm 4.98$ \\
  \midrule
\rowcolor{celadon!25} \multirow{2}{*}{\cellcolor{white} \texttt{Svmguide1}} & \texttt{IID} & $93.01 \pm 1.63$ & 	$93.22 \pm 1.66$ & $92.77 \pm 1.77$ & $93.4 \pm 1.21$  \\
 & \texttt{SSB} & $70.89 \pm 10.98$ & $70.22 \pm 11.64$ & $69.84 \pm 11.06$ & $71.04 \pm 11.11$ \\
    \bottomrule
    \end{tabular}
    }
\end{table}

\begin{figure}[!b]
\centering
\subfloat[Mnist.]{\includegraphics[width=0.29\textwidth]{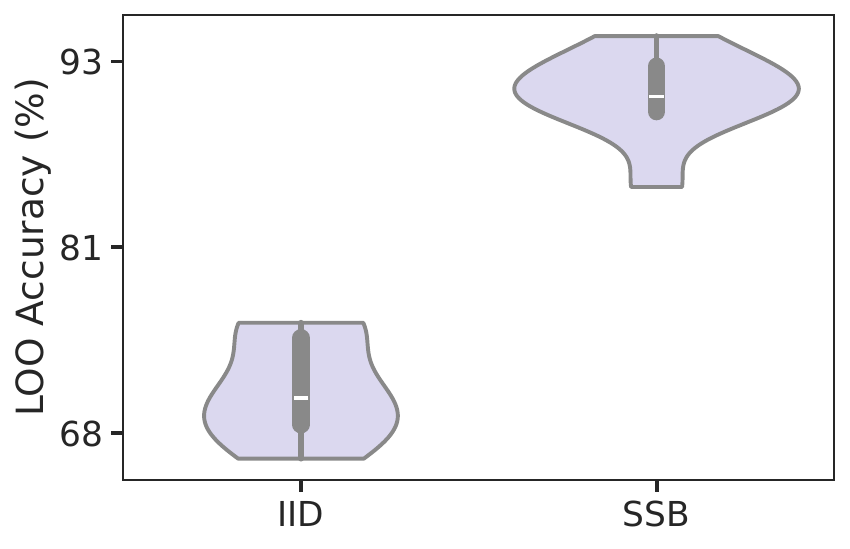}} 
\label{fig:unreliable_model_selection_mnist} \qquad
\subfloat[Mushrooms.]{\includegraphics[width=0.29\textwidth]{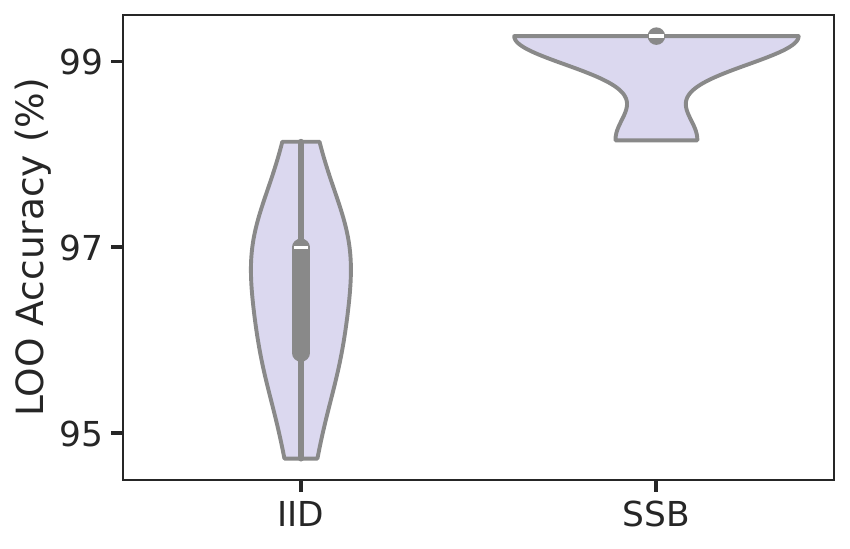}} 
\label{fig:unreliable_model_selection_mushrooms} \qquad
\subfloat[Protein.]{\includegraphics[width=0.29\textwidth]{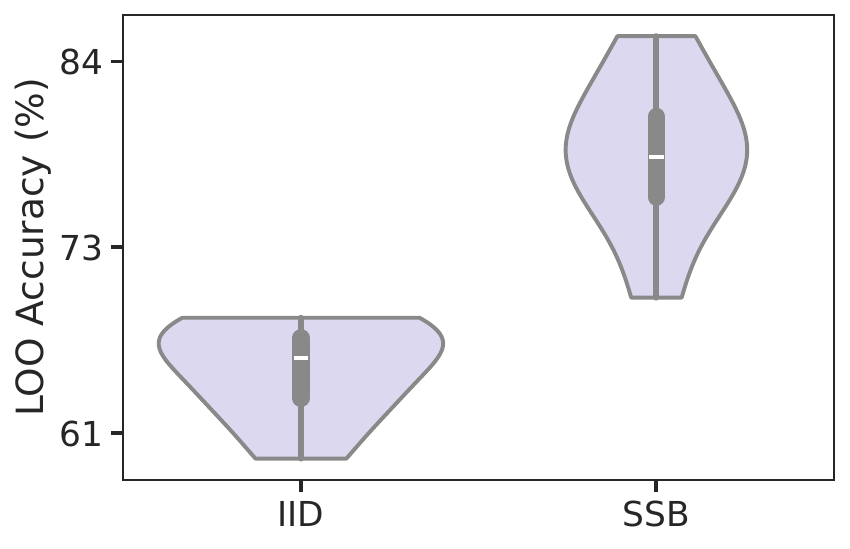}}
\label{fig:unreliable_model_selection_protein} \qquad
\caption{Leave-one-out on the labeled set of \texttt{Mnist}, \texttt{Mushrooms} and \texttt{Protein}. We display the distribution of the leave-one-out accuracy (\%) over the $9$ trials.}
 \label{fig:unreliable_model_selection}
\end{figure}

\subsection{Unreliable model selection}
\label{app:app_model_selection}
To further motivate the importance of the framework considered in this paper, we now look at it from a model selection perspective. Traditionally, values of hyperparameters are searched by cross-validation, where the performance of each candidate model is evaluated on a separate validation set (in average over different train/validation splits). In semi-supervised learning, as labeled data are scarce, it is reasonable to do leave-one-out~\citep{lachenbruch1967almost}: each labeled example is used as a validation set while training is performed on the remaining examples. We will refer to the corresponding average validation accuracy as the leave-one-out accuracy, whose maximization eventually determines the chosen values of hyperparameters. Although cross-validation is usually considered to be an unbiased estimator \citep{Hastie:2009}, it is often not the case for \texttt{SSL} \citep{madani2004co,feofanov2022wrapper,Feofanov:2023}. We show that in the presence of sample selection bias, this problem is further exacerbated. For this, we evaluate the leave-one-out accuracy in both \texttt{SSB} and \texttt{IID} settings, comparing these two values. We perform the experiment on \texttt{Mushrooms}, \texttt{Protein}, and \texttt{Mnist} using the same architectures and training parameters as before. On each dataset, we repeat the experiment with $9$ different seeds and display the distribution of the leave-one-out accuracy in Figure~\ref{fig:unreliable_model_selection}. We can see that on all datasets, the leave-one-out accuracy under \texttt{SSB} is always higher than under \texttt{IID}. The analysis of these results is two-fold: \textbf{(1)} it supports our intuition that the failure of self-training methods is due to a biased confidence measure and not to a potentially uninformative labeled set, \textbf{(2)} it highlights the risk of performing cross-validation as a model selection tool in the case of \texttt{SSB}, as the scores are overly optimistic and do not correspond to the real generalization performance (Table ~\ref{tab:bias_no_bias_comparative_performance}).

\def\figlength{0.45}
\begin{figure}[!b]
\centering
\subfloat[\texttt{IID}]{\includegraphics[width=\figlength\textwidth]{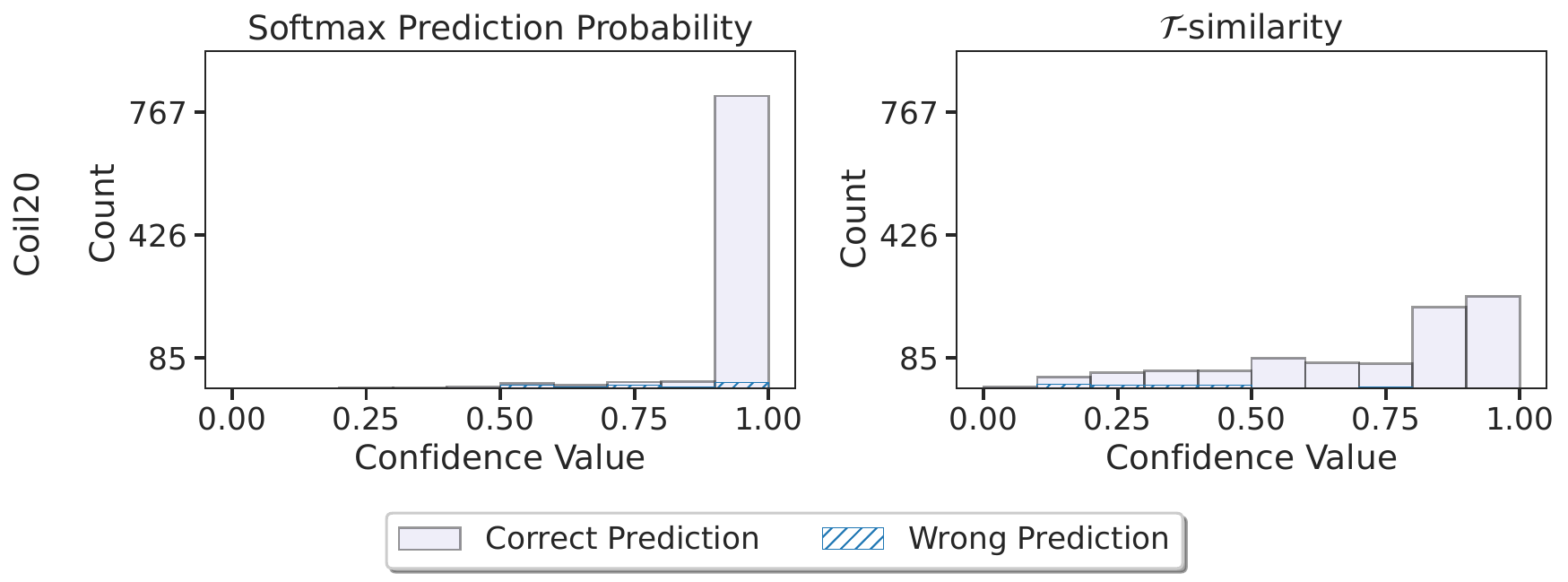}}
\label{fig:calibrated_similarity_coil20_balanced} \qquad
\subfloat[\texttt{SSB}]{\includegraphics[width=\figlength\textwidth]{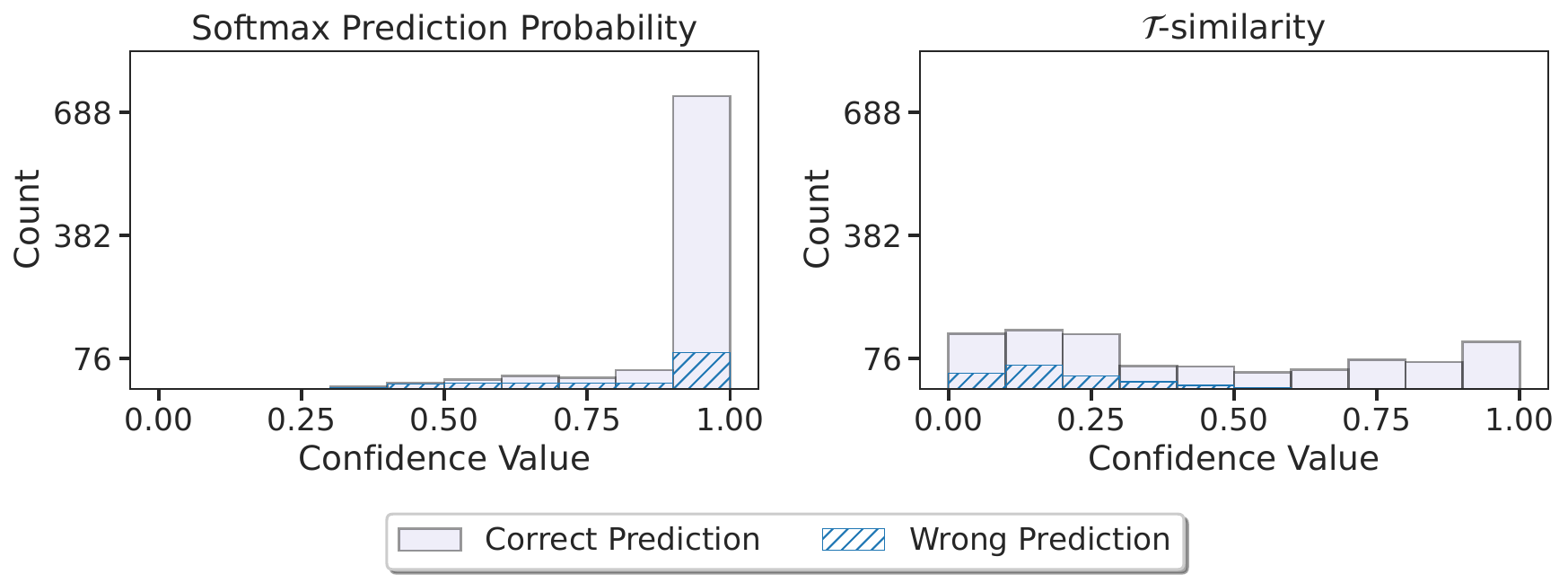}}
\label{fig:calibrated_similarity_coil20_pca} \qquad

\subfloat[\texttt{IID}]{\includegraphics[width=\figlength\textwidth]{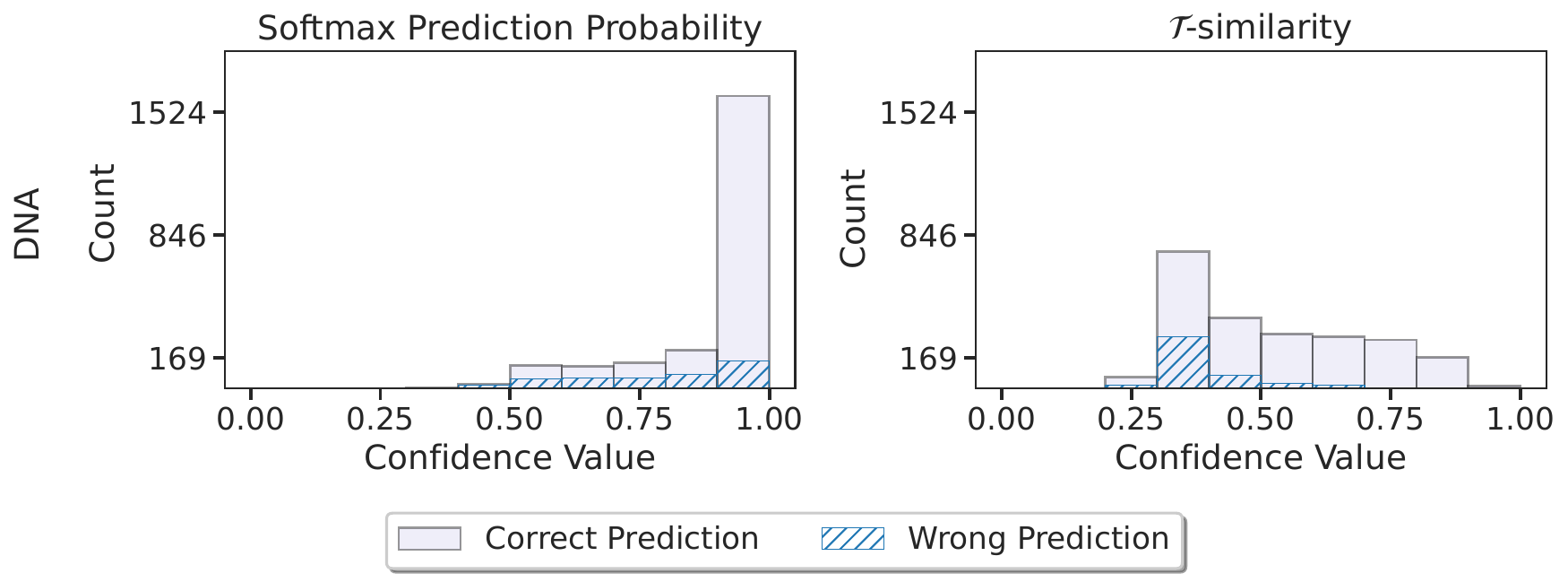}}
\label{fig:calibrated_similarity_dna_balanced} \qquad
\subfloat[\texttt{SSB}]{\includegraphics[width=\figlength\textwidth]{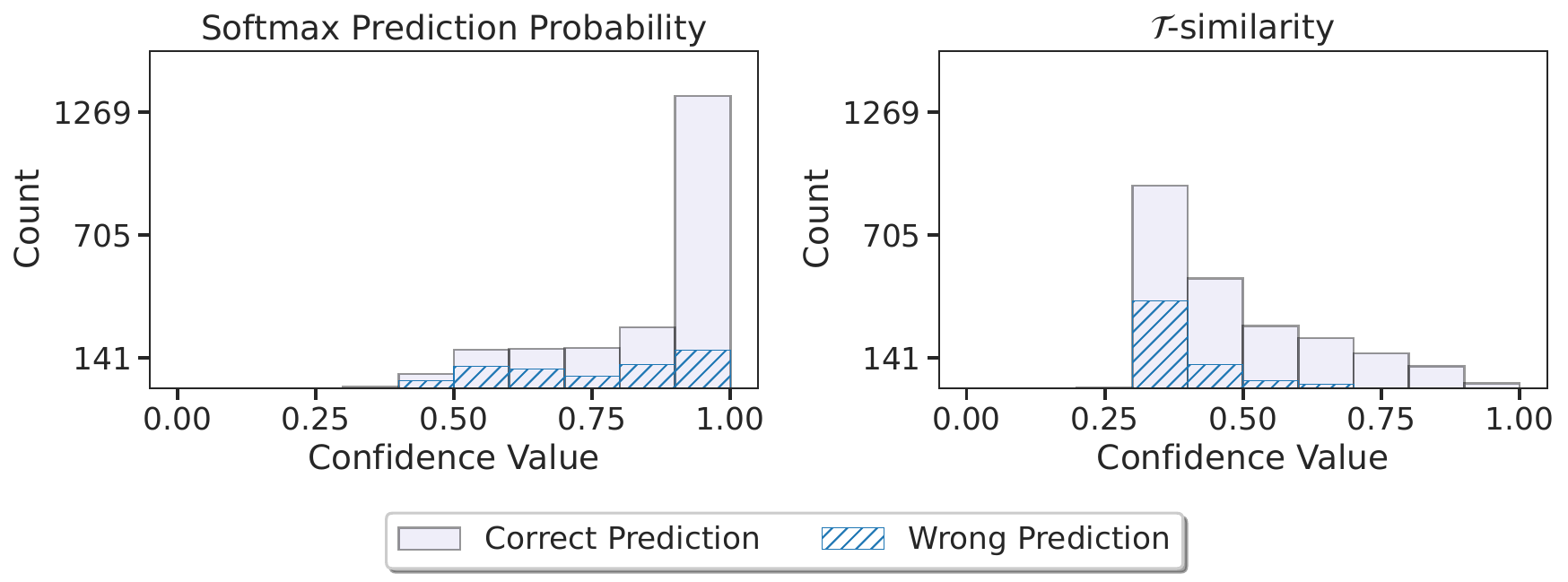}}
\label{fig:calibrated_similarity_dna_pca} \qquad

\subfloat[\texttt{IID}]{\includegraphics[width=\figlength\textwidth]{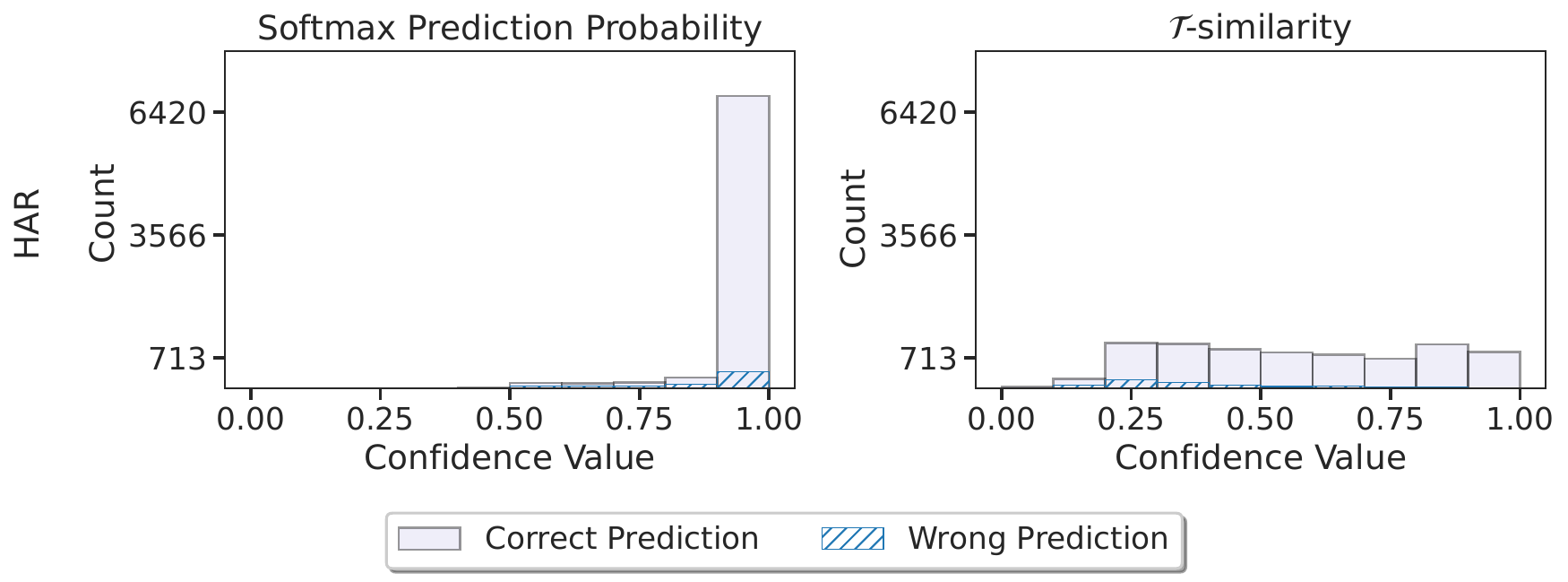}} 
\label{fig:calibrated_similarity_har_balanced} \qquad
\subfloat[\texttt{SSB}]{\includegraphics[width=\figlength\textwidth]{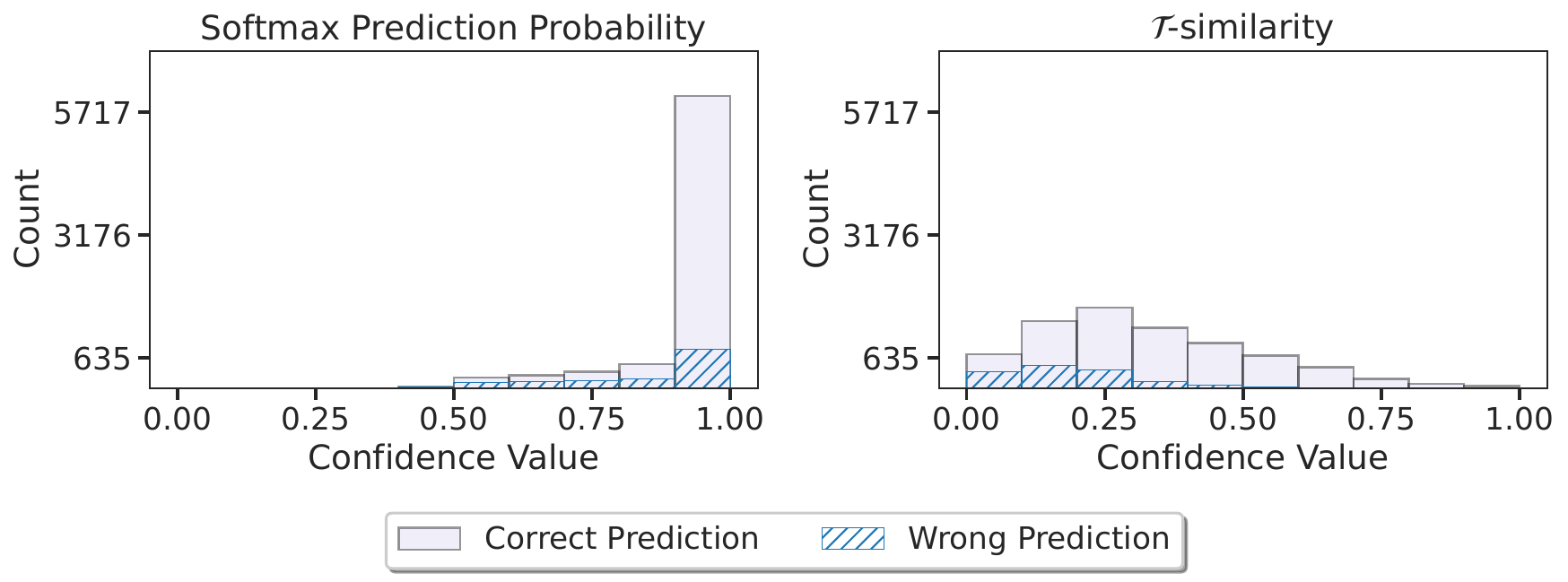}}
\label{fig:calibrated_similarity_har_pca} \qquad

\subfloat[\texttt{IID}]{\includegraphics[width=\figlength\textwidth]{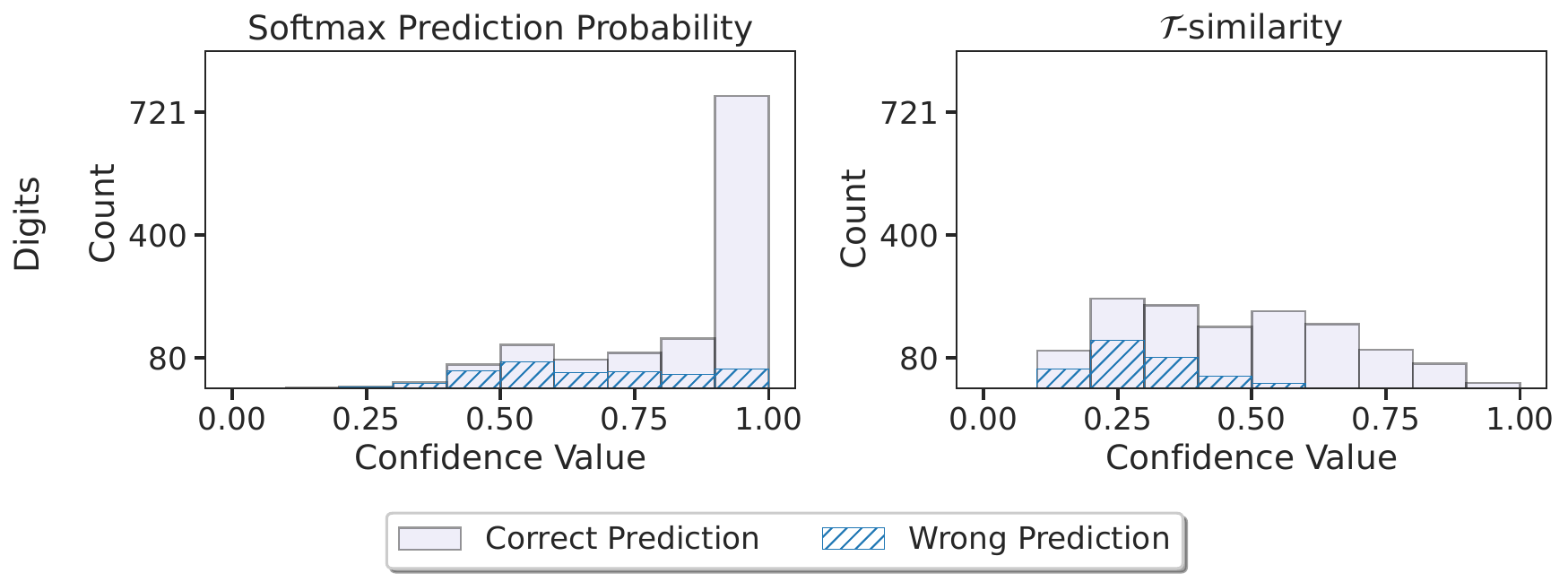}} 
\label{fig:calibrated_similarity_digits_balanced} \qquad
\subfloat[\texttt{SSB}]{\includegraphics[width=\figlength\textwidth]{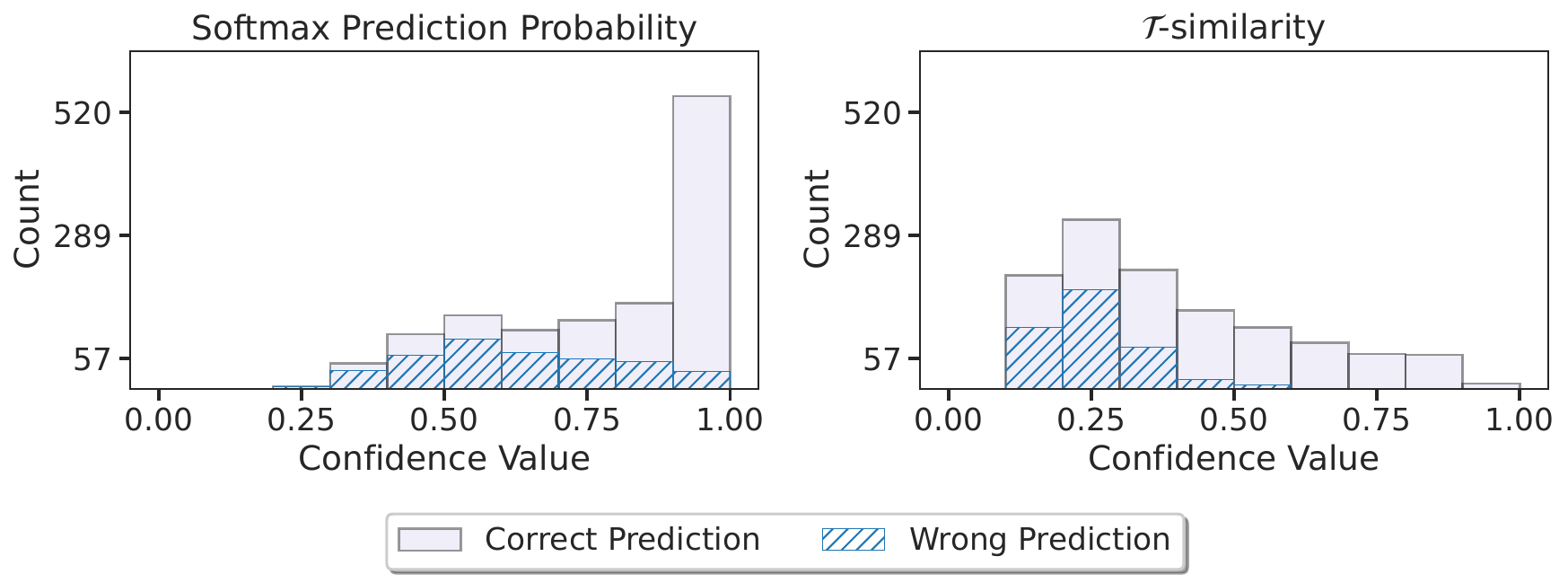}}
\label{fig:calibrated_similarity_digits_pca} \qquad

\subfloat[\texttt{IID}]{\includegraphics[width=\figlength\textwidth]{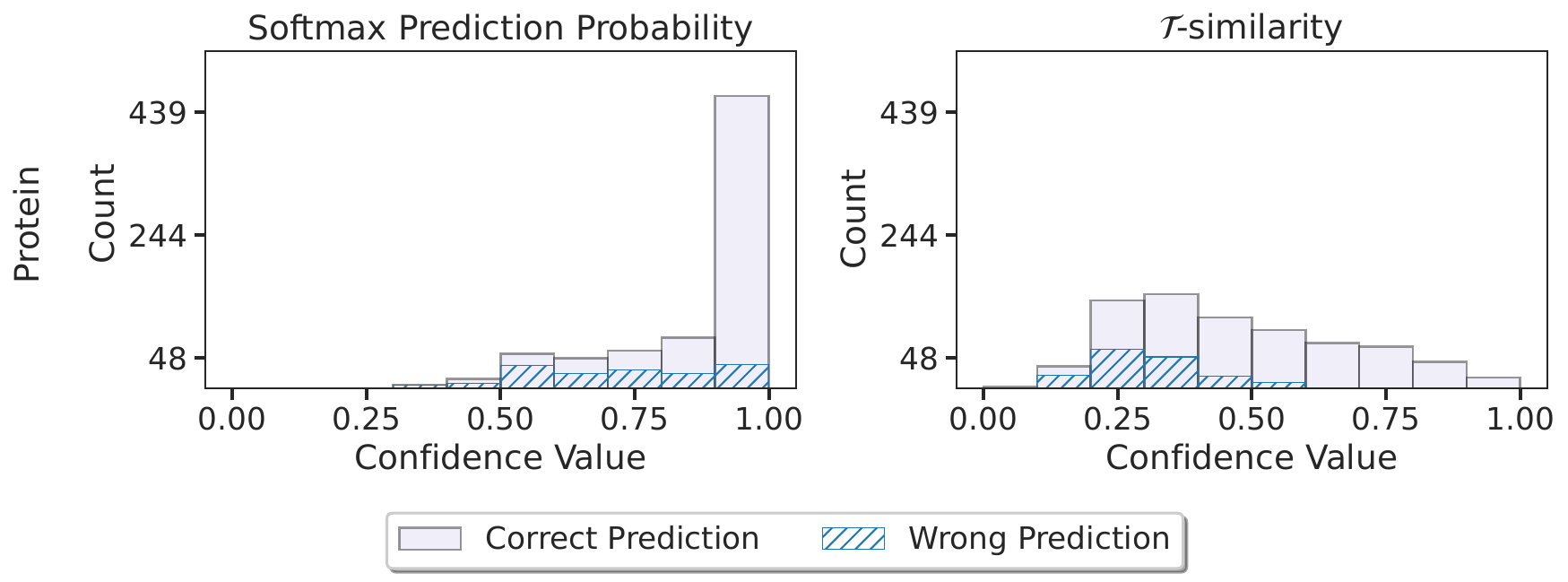}}
\label{fig:calibrated_similarity_protein_balanced} \qquad
\subfloat[\texttt{SSB}]{\includegraphics[width=\figlength\textwidth]{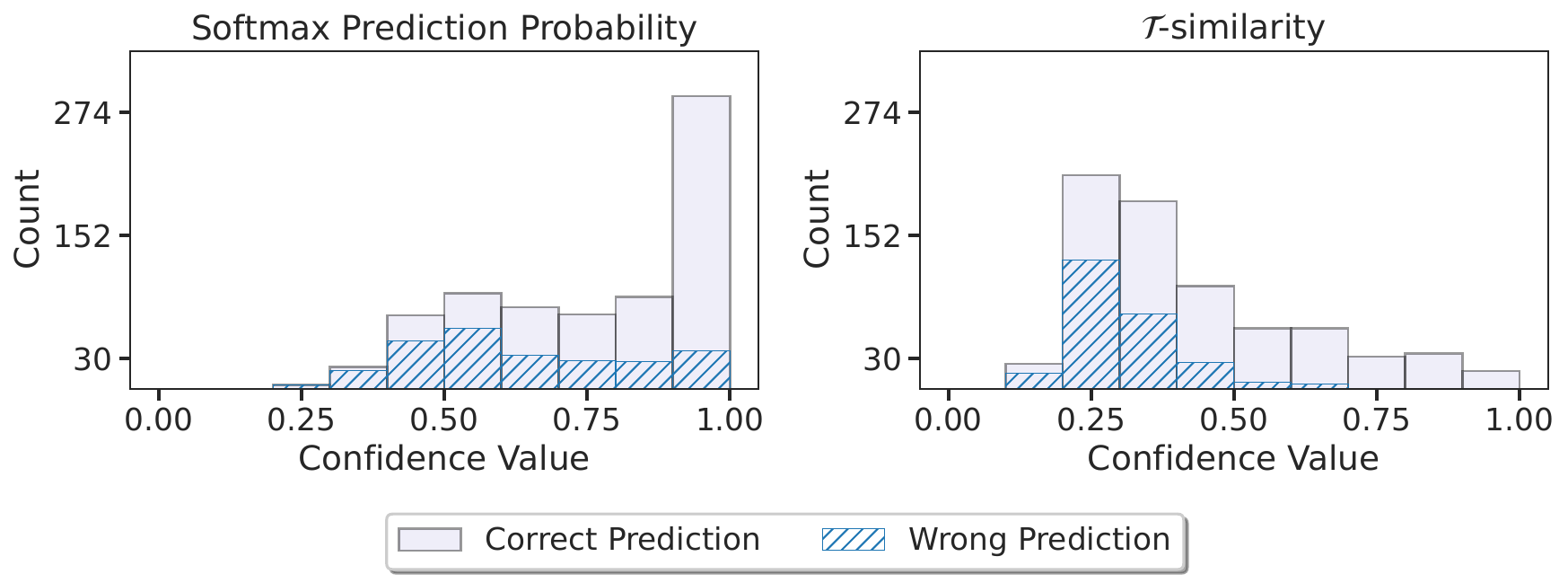}}
\label{fig:calibrated_similarity_protein_pca} \qquad
\caption{We display the distribution over the \texttt{softmax} and the $\mathcal{T}$-similarity on the unlabeled examples correctly classified by the base classifier and on the unlabeled examples misclassified by the base classifier.}
 \label{fig:calibrated_uncertainty}
\end{figure}

\subsection{Correcting the \texttt{softmax} overconfidence}
\label{app:calibration}
As in the main paper, for this experiment, we artificially use the true labels on $\mathbf{X}_u$ to compute the distribution of $s_\mathcal{T}$ on the examples for which the prediction of the model is accurate (Correct Prediction) and on the examples for which the prediction is incorrect (Wrong Prediction). We compute the \texttt{softmax} prediction probabilities and the $\mathcal{T}$-similarity values on the unlabeled examples. Again, we artificially use the true labels to plot the distribution of the confidence value (\texttt{softmax} or $\mathcal{T}$-similarity) on the examples for which the prediction is accurate (Correct Prediction) and on the examples for which the prediction is incorrect (Wrong Prediction). We display the plots obtained in Figure ~\ref{fig:calibrated_uncertainty} for various datasets. We can see that the obtained results are very similar to those presented in the main paper and our conclusions about them hold.

\setlength{\tabcolsep}{0.4em}
\begin{table}[!t] 
\centering
\caption{Classification performance of the different baselines on the datasets described in Table~\ref{tab:dataset_description} when labeling is done with \texttt{IID}. We display the average and the standard deviation of the test accuracy (both in \%) over the $9$ trials. The \texttt{softmax} corresponds to the usual self-training which uses the \texttt{softmax} prediction probability as a confidence estimate while the $\mathcal{T}$-similarity corresponds to our proposed method in Algorithm~\ref{alg:self_training_algorithm}. For each baseline, the best result between \texttt{softmax} and $\mathcal{T}$-similarity is in \textbf{bold}.}
\label{tab:comparative_performance_no_bias}
\scalebox{0.8}{
\begin{tabular}{l||c||cc||cc||cc}
\toprule
\multicolumn{1}{c}{\multirow{2}{*}{Dataset}} & \multicolumn{1}{c}{\multirow{2}{*}{\texttt{ERM}}} & \multicolumn{2}{c}{$\texttt{PL}_{\theta=0.8}$} & \multicolumn{2}{c}{$\texttt{CSTA}_{\Delta=0.4}$}  & \multicolumn{2}{c}{\texttt{MSTA}}\\ 
\cmidrule(r{10pt}l{5pt}){3-4} \cmidrule(r{10pt}l{5pt}){5-6} \cmidrule(r{10pt}l{5pt}){7-8}
\multicolumn{1}{c}{} & \multicolumn{1}{c}{} & \multicolumn{1}{c}{\texttt{softmax}} & \multicolumn{1}{c}{$\mathcal{T}$-similarity} & \multicolumn{1}{c}{\texttt{softmax}} & \multicolumn{1}{c}{$\mathcal{T}$-similarity} & \multicolumn{1}{c}{\texttt{softmax}} & \multicolumn{1}{c}{$\mathcal{T}$-similarity} \\
\midrule
\texttt{Cod-RNA} & $89.28 \pm 2.13$ & $\mathbf{89.34 \pm 2.44}$ & $84.87 \pm 3.54$ & $\mathbf{89.09 \pm 2.37}$ & $87.85 \pm 2.24$ & $\mathbf{89.89 \pm 1.89}$ & $89.65 \pm 2.19$ \\
\texttt{COIL-20} & $93.18 \pm 1.5$ & $\mathbf{93.77 \pm 1.19}$ & $93.49 \pm 1.97$ & 	$93.09 \pm 1.73$ & $\mathbf{93.3 \pm 1.77}$ & $93.21 \pm 1.57$ & $\mathbf{94.17 \pm 1.99}$ \\
\texttt{Digits} & $81.38 \pm 2.45$ & $\mathbf{83.58 \pm 3.26}$ & $81.36 \pm 2.75$ & $81.78 \pm 2.51$ & $\mathbf{81.88 \pm 3.02}$& $\mathbf{83.04 \pm 2.13}$ & 	$82.62 \pm 3.03$ \\
\texttt{DNA} & $81.28 \pm 2.27$ & $\mathbf{81.25 \pm 2.5}$ & $79.49 \pm 2.53$ & $81.54 \pm 2.44$ & $\mathbf{81.64 \pm 2.3}$ & $83.19 \pm 1.96$ & $\mathbf{84.72 \pm 2.3}$ \\
\texttt{DryBean} & $86.85 \pm 1.68$ & $\mathbf{87.59 \pm 1.6}$ & $86.6 \pm 1.85$ & $\mathbf{86.98 \pm 1.61}$ & $86.74 \pm 1.75$ & $\mathbf{87.72 \pm 1.54}$ & $87.35 \pm 2.19$ \\
\texttt{HAR} & $91.16 \pm 0.54$ & 	$91.24 \pm 1.03$ & $\mathbf{91.24 \pm 0.38}$ & $\mathbf{91.36 \pm 0.24}$ & $91.29 \pm 0.39$ & $\mathbf{89.29 \pm 1.24}$ & $89.25 \pm 1.26$ \\
\texttt{Mnist} & $73.98 \pm 1.46$ & $\mathbf{74.61 \pm 1.85}$ & $72.16 \pm 3.1$ & $\mathbf{75.24 \pm 1.48}$ & $73.42 \pm 1.6$ & $\mathbf{74.6 \pm 1.76}$ & $73.36 \pm 1.33$ \\
\texttt{Mushrooms} & $96.48 \pm 1.57$ & $\mathbf{96.56 \pm 1.26}$ & $96.23 \pm 1.57$ & $\mathbf{96.3 \pm 1.32}$ & $96.25 \pm 1.38$ & $\mathbf{96.68 \pm 1.31}$ & $96.44 \pm 1.33$ \\
\texttt{Phishing} & $88.51 \pm 1.51$ & 	$\mathbf{89.02 \pm 1.37}$ & $88.15 \pm 1.4$ & $\mathbf{88.96 \pm 1.37}$ & $88.82 \pm 1.06$ & $88.82 \pm 1.7$ & $\mathbf{88.94 \pm 1.85}$ \\
\texttt{Protein} & $73.74 \pm 4.78$ & $74.86 \pm 3.48$ & $\mathbf{75.43 \pm 2.65}$ & $74.73 \pm 3.01$ & $\mathbf{75.68 \pm 2.93}$ & $73.99 \pm 5.6$ & $\mathbf{75.1 \pm 4.98}$ \\
\texttt{Rice} & $88.24 \pm 3.63$ & $\mathbf{88.34 \pm 3.18}$ & 	$88.32 \pm 3.29$ & $88.15 \pm 3.46$ & $\mathbf{88.7 \pm 2.92}$ & $88.69 \pm 3.49$ & $\mathbf{89.75 \pm 2.1}$ \\
\texttt{Splice} & $69.09 \pm 4.09$ & 	$\mathbf{70.26 \pm 4.08}$ & $69.65 \pm 3.94$ & 	$\mathbf{70.46 \pm 4.32}$ & $70.32 \pm 4.09$ & $69.65 \pm 4.27$ & $\mathbf{70.51 \pm 4.26}$ \\
\texttt{Svmguide1} & $93.01 \pm 1.63$ & $\mathbf{92.94 \pm 1.91}$ & $91.92 \pm 2.26$ & $\mathbf{92.77 \pm 1.77}$ & $92.31 \pm 2.04$ & $\mathbf{93.4 \pm 1.21}$ & $92.83 \pm 1.48$ \\
\bottomrule
\end{tabular}
}
\end{table}

\subsection{The $\mathcal{T}$-similarity is comparable to the \texttt{softmax} when there is no sample selection bias}
\label{app:comparative_performance_no_bias}
We perform the same experiment as in Table \ref{tab:comparative_performance_pca_bias} but when the labeling is done with \texttt{IID}. The obtained results are in Table~\ref{tab:comparative_performance_no_bias}. It should be noted that in this setting, even if the \texttt{softmax} is subject to overconfidence, the pseudo-labeling can still be of good quality as the distributions of labeled and unlabeled samples do not differ. We can observe that the $\mathcal{T}$-similarity induces a slight decrease in performance on most datasets. However, it manages to remain competitive, notably for $\texttt{CSTA}_{\Delta=0.4}$ and \texttt{MSTA}, where it even manages 
\begin{wrapfigure}[11]{r}{0.35\textwidth}
\centering
    \includegraphics[width=0.9\linewidth]{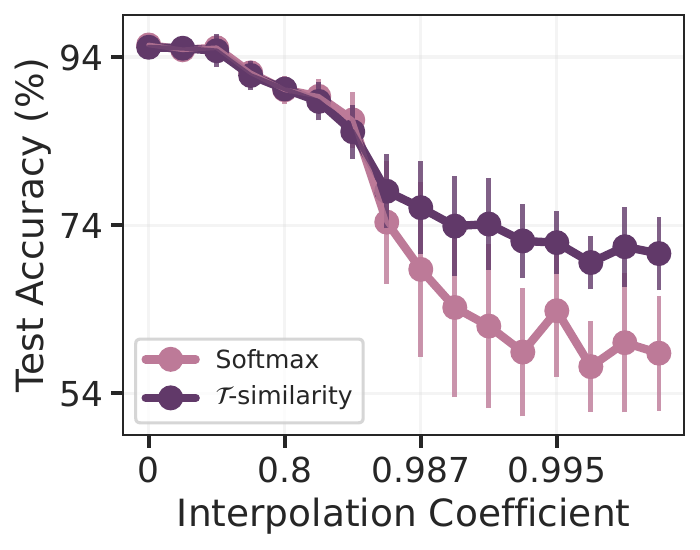}
    \caption{The $\mathcal{T}$-similarity is more robust to distribution shift than \texttt{softmax}.}
    \label{fig:interpolation_pca}
\end{wrapfigure}
to be better or similar on $4$ datasets out of $13$.
\subsection{Robustness to the strength of the bias}
\label{app:app_ssb_stength}
In this experiment, we study the robustness of our method to the strength of the bias. More specifically, we make the labeling procedure gradually evolve from \texttt{IID} to \texttt{SSB}. We consider for each class $c$ and for each data $\mathbf{x}$ with label $y=c$ 
\begin{equation*}
    P_{\alpha}(s=1|\mathbf{x}, y=c) = (1-\alpha) \cdot \underbrace{\frac{1}{\mathrm{card}(\mathcal{T}_c)}}_{\texttt{IID}} \;+\; \alpha \cdot \underbrace{\frac{1}{\beta}\exp(r \times \lvert \mathrm{proj}_1(\mathbf{x}) \rvert)}_{\texttt{SSB}},
\end{equation*}
where $\alpha \in [0, 1]$ is the interpolation coefficient and $\mathcal{T}_c = \{(\mathbf{x}, y) | y=c\}$ is the set of training data of class $c$. Increasing $\alpha$ corresponds to imposing more sample selection bias in the labeling procedure. In Figure \ref{fig:interpolation_pca}, we display the test accuracy on \texttt{Mushrooms} when the labeling procedure is done with $P_{\alpha}$, where $\alpha$ varies in $[0,1]$. We note that, up to some point, increasing $\alpha$ does not lead to a strong distribution mismatch, and both the \texttt{softmax} prediction probabilities and the proposed $\mathcal{T}$-similarity give similar results. However, when the value of $\alpha$ exceeds $0.8$, test accuracy suffers from a sharp drop for both methods, with a more drastic decrease in the case of $\texttt{softmax}$. We empirically demonstrate that our proposed $\mathcal{T}$-similarity is more robust to distribution shift than the \texttt{softmax}. 

\section{Ablation study and sensitivity analysis}
\label{app:ablation_study_sensitivity}
\subsection{Ablation study with different number of labeled examples}
\label{app:ablation_study}
Intuitively, when the number of labeled examples increases, the impact of the \texttt{SSB} labeling procedure should decrease as the base classifier has more labeled data to learn from and hence relies less on the unlabeled data. To study the robustness of our method when the impact of sample selection bias decreases, we observe the performance of each pseudo-labeling policy with the \texttt{softmax} and the $\mathcal{T}$-similarity when the number of labeled data $n_\ell$ increases from $20$ to $2000$. We display the results in Figure~\ref{fig:ablation_study_lab_size}. We observe that the two methods behave similarly in \texttt{IID} setting and the performance gradually increases with the number of available labeled points. However, in \texttt{SSB} setting, we see that $\mathcal{T}$-similarity leads to much better results for realistic scenarios commonly considered in \texttt{SSL} when the number of samples is small. This improvement is consistent across all pseudo-labeling policies considered.

\def\figlength{0.45}
\begin{figure}[htbp]
\centering
\subfloat[$\texttt{PL}_{\theta=0.8}$]{\includegraphics[width=\figlength\textwidth]{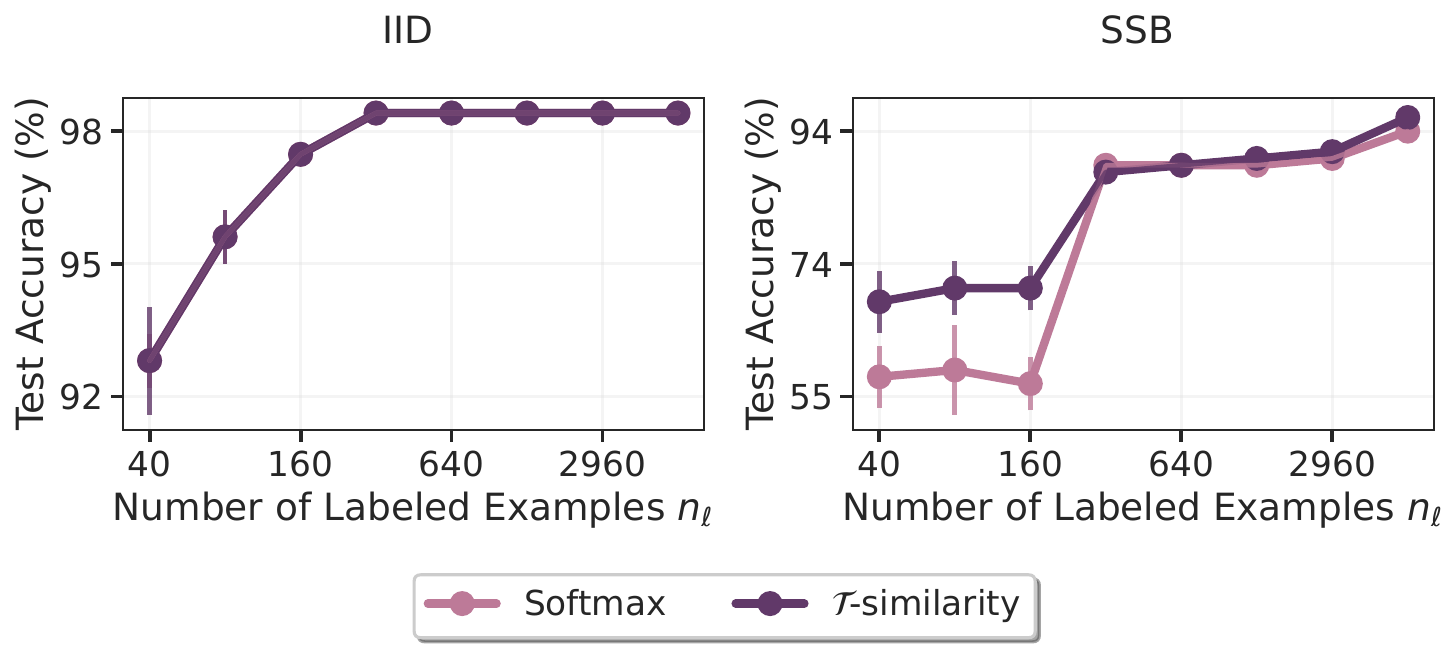}}
\label{fig:ablation_study_lab_size_Fixed_mushrooms}
\subfloat[$\texttt{CSTA}_{\Delta=0.4}$]{\includegraphics[width=\figlength\textwidth]{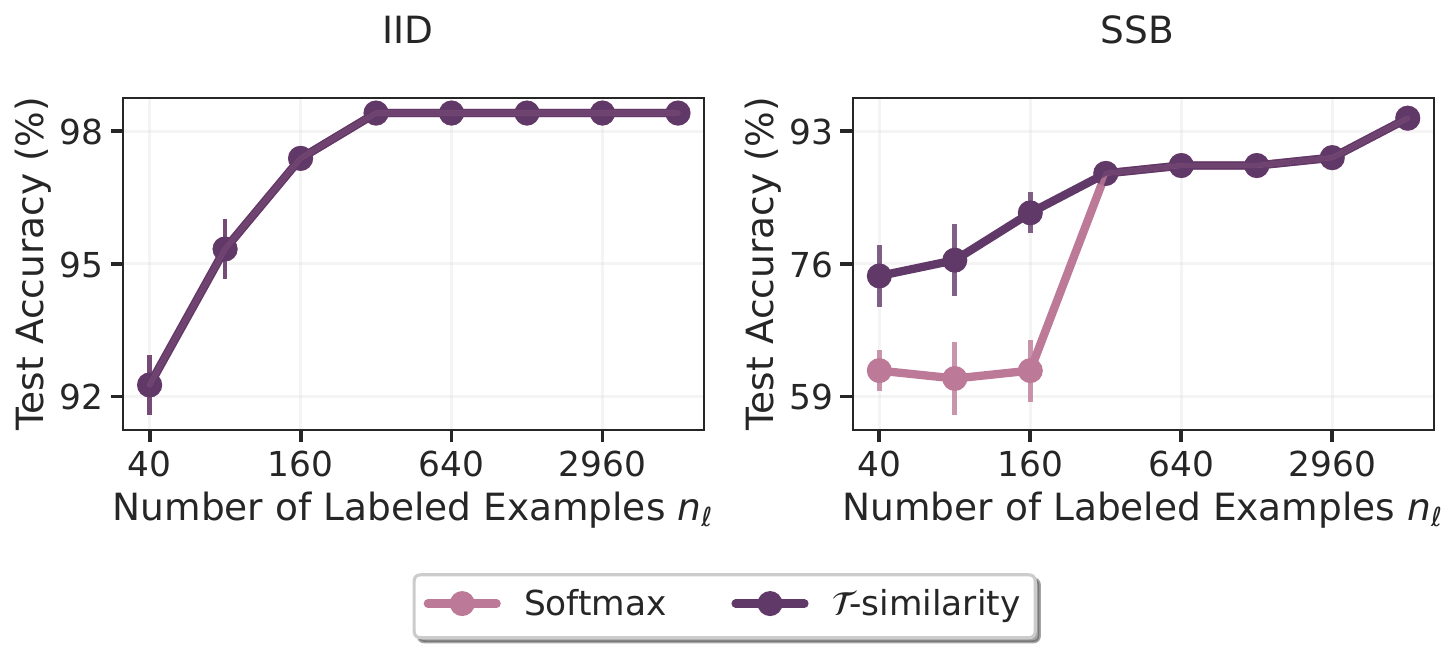}}
\label{fig:ablation_study_lab_size_Curriculum_mushrooms} 
\subfloat[MSTA]{\includegraphics[width=\figlength\textwidth]{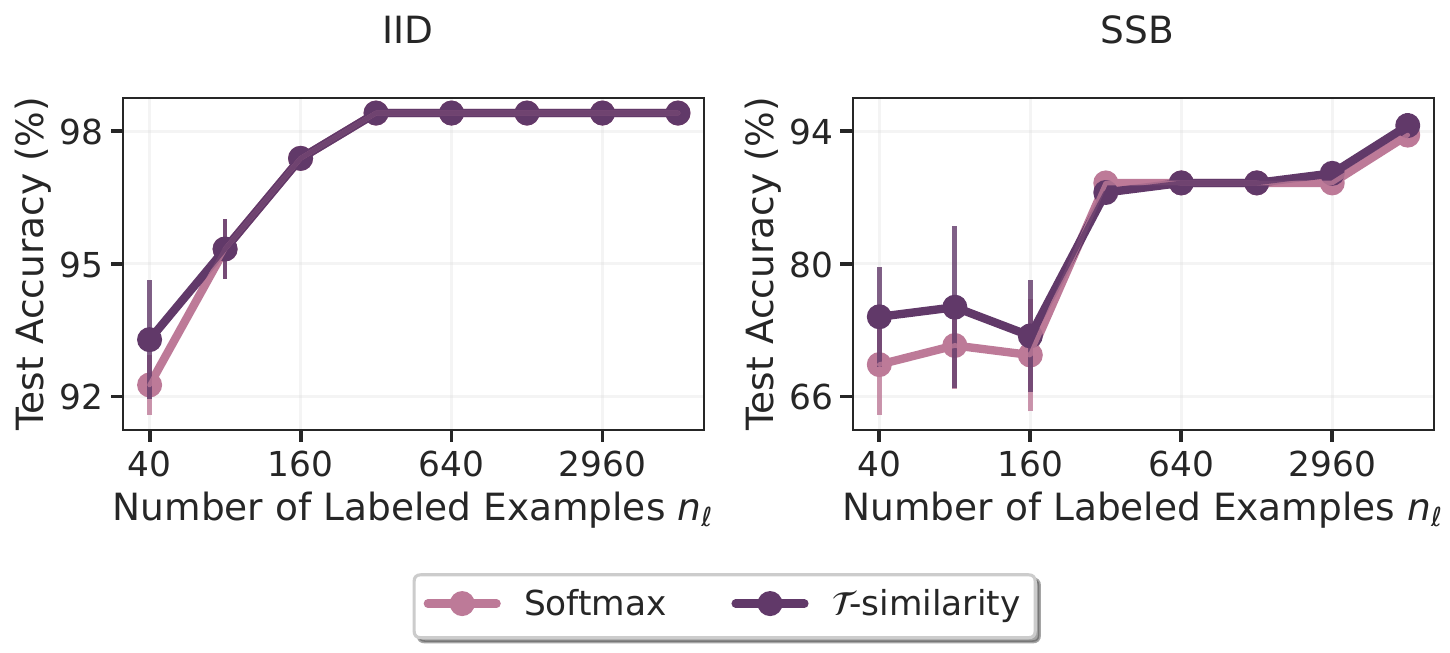}}
\label{fig:ablation_study_lab_size_MSTA_mushrooms} \qquad
\caption{Ablation study on the number of labeled examples $n_\ell$ on \texttt{Mushrooms} for the three pseudo-labeling policies. }
 \label{fig:ablation_study_lab_size}
\end{figure}

\subsection{Sensitivity to hyperparameters}
\label{app:sensitivity_analysis}
We conduct a sensitivity analysis on \texttt{Mushrooms} to see how self-training with the proposed $\mathcal{T}$-similarity behaves under different choices of hyperparameters. For each pseudo-labeling policy, we make the diversity strength $\gamma$ vary in $\{0, 0.5, 1, 1.5, 2\}$ and display the results in Figure~\ref{fig:ablation_study_diversity_tradeoff}. We observe that in \texttt{SSB} setting all pseudo-labeling policies benefit from the diversity almost for all positive values of $\gamma$. Similarly, we see that in \texttt{IID}, the diversity does not hurt the performance. In the same fashion, we make the number of classifiers $M$ vary in $\{2, 5, 10\}$ and display the results Figure~\ref{fig:ablation_study_n_classifiers}. The behavior is similar to the previous experiment and is consistent with our observations made so far.

\def\figlength{0.45}
\begin{figure}[htbp]
\centering
\subfloat[$\texttt{PL}_{\theta=0.8}$]{\includegraphics[width=\figlength\textwidth]{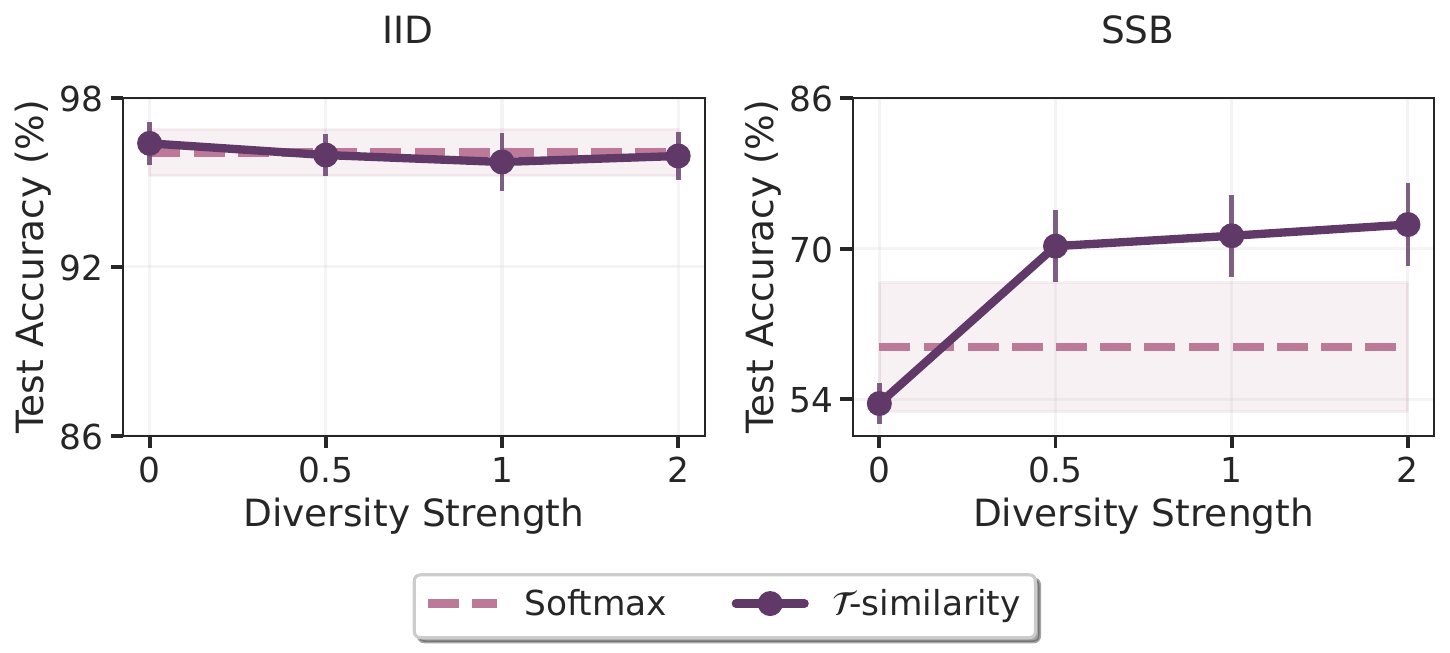}}
\label{fig:ablation_study_diversity_tradeoff_Fixed_mushrooms}
\subfloat[$\texttt{CSTA}_{\Delta=0.4}$]{\includegraphics[width=\figlength\textwidth]{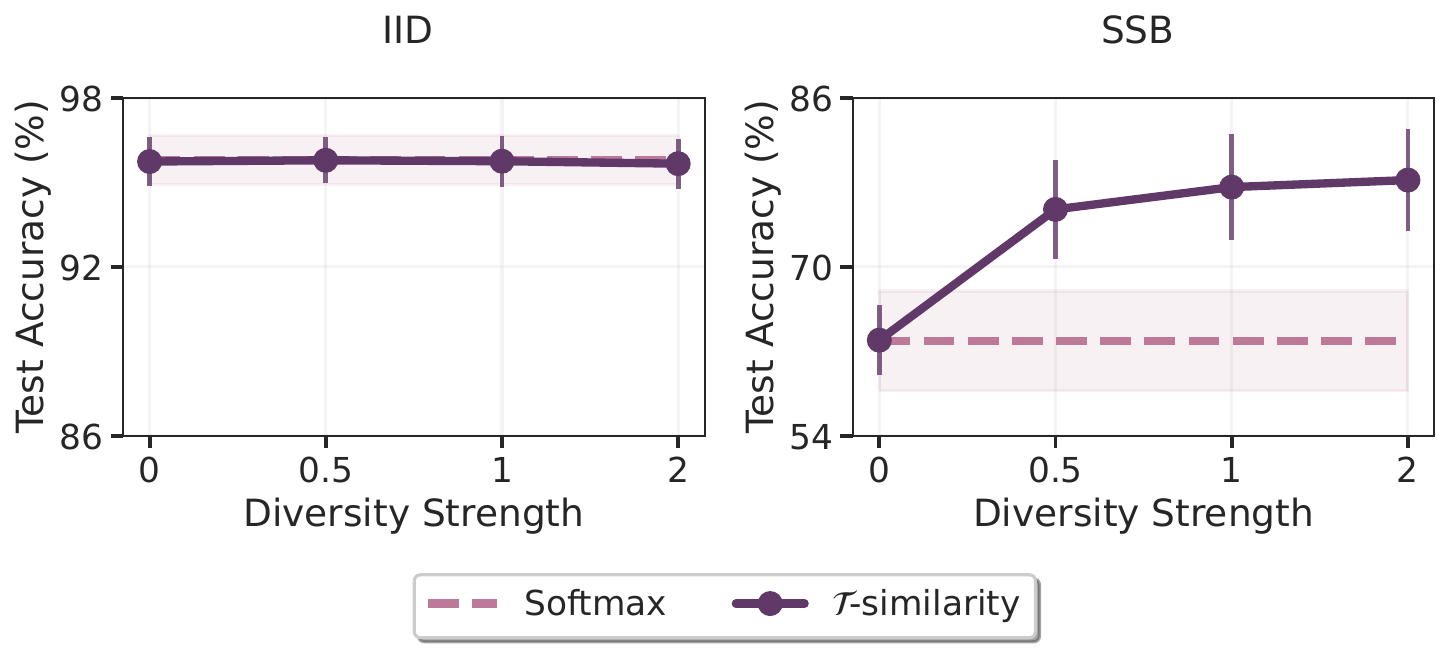}}
\label{fig:ablation_study_diversity_tradeoff_Curriculum_mushrooms} 
\subfloat[MSTA]{\includegraphics[width=\figlength\textwidth]{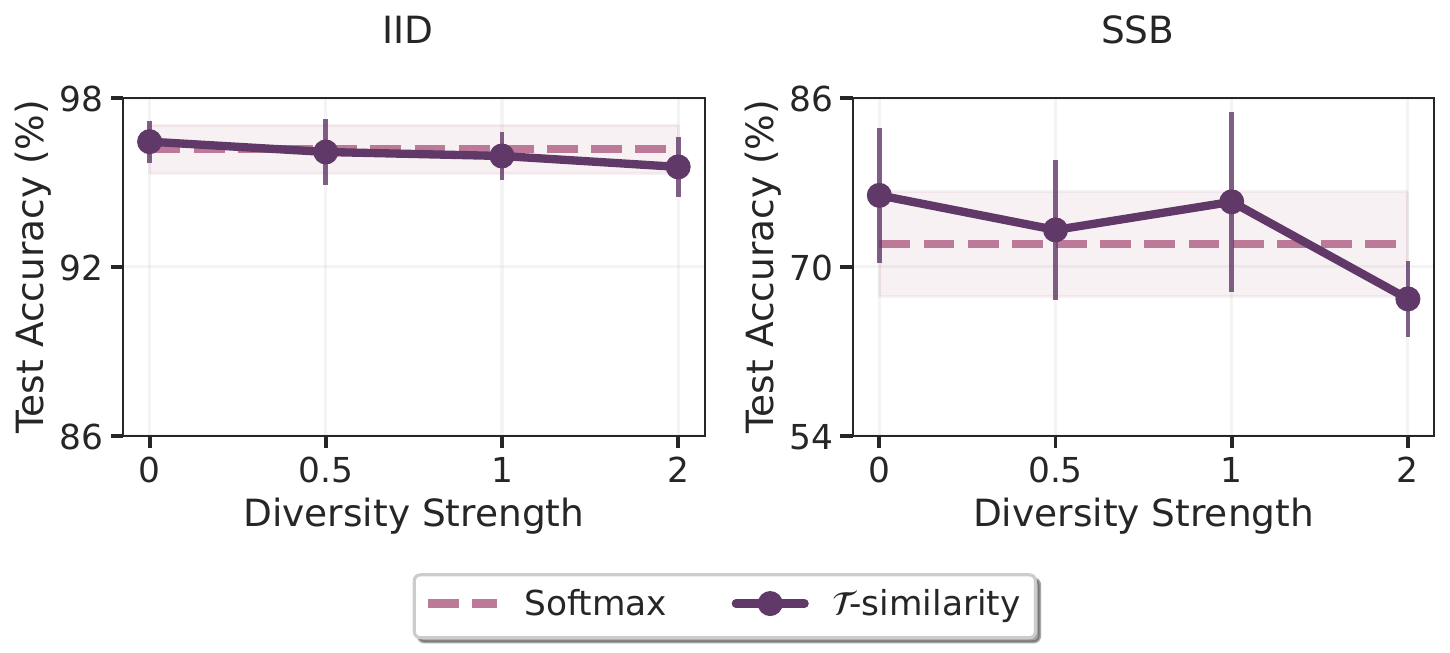}}
\label{fig:ablation_study_diversity_tradeoff_MSTA_mushrooms} \qquad
\caption{Sensitivity analysis of the diversity strength parameter $\gamma$ on \texttt{Mushrooms} for the three pseudo-labeling policies.}
 \label{fig:ablation_study_diversity_tradeoff}
\end{figure}

\def\figlength{0.45}
\begin{figure}[htbp]
\centering
\subfloat[$\texttt{PL}_{\theta=0.8}$]{\includegraphics[width=\figlength\textwidth]{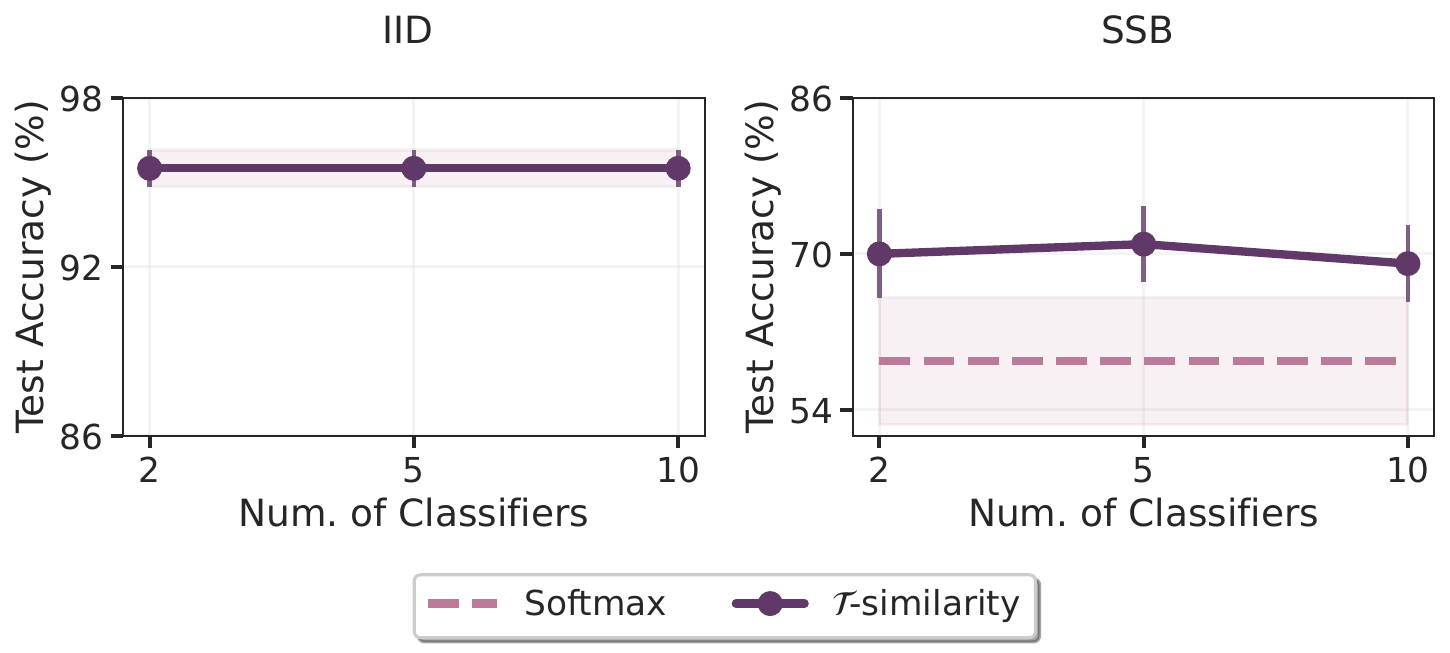}}
\label{fig:ablation_study_n_classifiers_Fixed_mushrooms}
\subfloat[$\texttt{CSTA}_{\Delta=0.4}$]{\includegraphics[width=\figlength\textwidth]{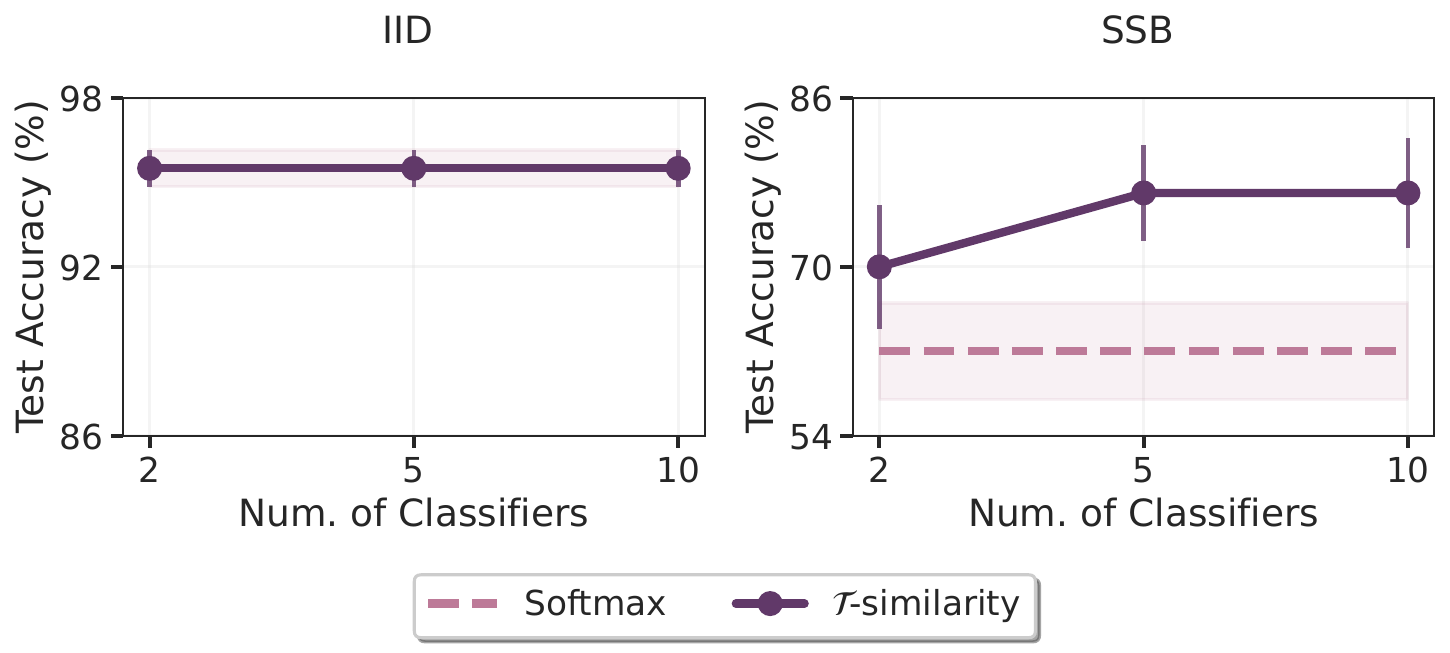}}
\label{fig:ablation_study_n_classifiers_Curriculum_mushrooms} 
\subfloat[MSTA]{\includegraphics[width=\figlength\textwidth]{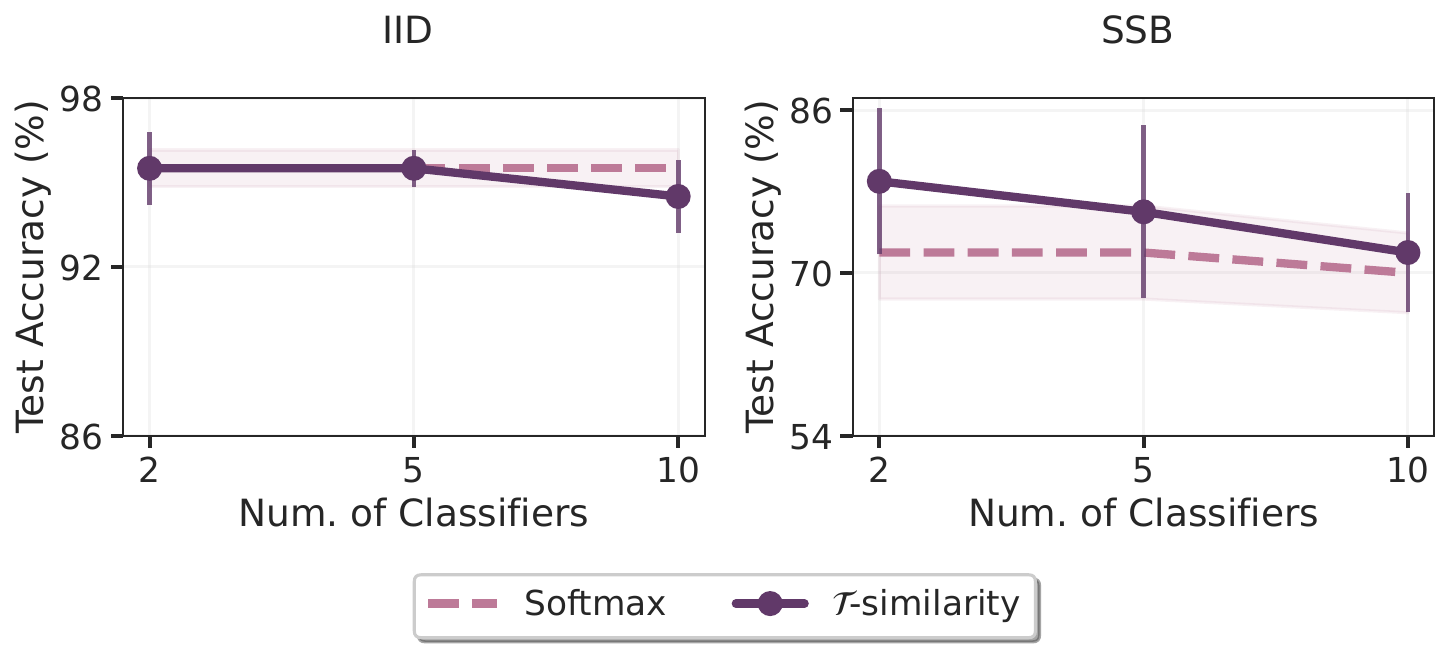}}
\label{fig:ablation_study_n_classifiers_MSTA_mushrooms} \qquad
\caption{Sensitivity analysis of the number of classifiers $M$ on \texttt{Mushrooms} for the three pseudo-labeling policy. We display the average and standard deviation of the test accuracy over $9$ seeds.}
 \label{fig:ablation_study_n_classifiers}
\end{figure}

\newpage
\section{Proofs} 
\label{app:proofs}
In this section, we detail the proofs of our theoretical results. 

\subsection{Notations} 
To ease the readability of the proofs, we recall the following notations. Scalar values are denoted by regular letters (e.g., parameter $\lambda$), vectors are represented in bold lowercase letters (e.g., vector $\mathbf{x}$) and matrices are represented by bold capital letters (e.g., matrix $\mathbf{A}$). The $i$-th row of the matrix $\mathbf{A}$ is denoted by $\mathbf{A}_{i}$ and its $j$-th column is denoted by $\mathbf{A}_{\cdot, j}$. The trace of a matrix $\mathbf{A}$ is denoted $\mathrm{Tr}(\mathbf{A})$ and its transpose by $\mathbf{A}^\top$. The identity matrix of size $n$ is denoted by $\mathbf{I}_n \in \mathbb{R}^{n \times n}$. The vector of size $n$ with each entry equal to $1$ is denoted by $\mathbb{1}_n$. The matrix of size $n$ with each entry equal to $1$ is denoted by $\mathbf{U}^{[n]} \in \mathbb{R}^{n \times n}$. We have $\mathbf{U}^{[n]} = \mathbb{1}_n \mathbb{1}_n^\top$. We denote by $\lambda_{\mathrm{min}}(\mathbf{A})$ and $\lambda_{\mathrm{max}}(\mathbf{A})$ the minimum and maximum eigenvalues of a matrix $\mathbf{A}$, respectively. For ease of notation, we define the following quantities:
\begin{equation}
\label{eq:notations}
    \mathbf{S}_u = \frac{\gamma (M+1)}{n_u(M-1)} \mathbf{X}_u^\top \mathbf{X}_u \text{ and } \alpha_u = \frac{\gamma}{2n_u(M-1)}.
\end{equation}

\subsection{Proof of Proposition~\ref{prop:similarity_0_1}}
\label{app:similarity_0_1}
The proof of Proposition~\ref{prop:similarity_0_1} is detailed below.
\begin{proof}
    For ease of notation, we will denote by $h_m^c$ the $c$-th entry of $h_m(\mathbf{x})$. Using the fact that each $h_m(\mathbf{x})$ is in the simplex $\Delta_C$, we have that
    \begin{equation*}
        0 \leq s_\mathcal{T}(\mathbf{x}) = \frac{1}{M(M-1)} \sum_{m \neq k} h_m(\mathbf{x})^\top h_k(\mathbf{x}) = \frac{1}{M(M-1)} \sum_{m \neq k} \sum_{c=1}^C h_m^c \underbrace{h_k^c}_{\leq 1} \leq \frac{1}{M(M-1)} \sum_{m \neq k} \sum_{c=1}^C h_m^c = 1.
    \end{equation*}
\end{proof}

\subsection{Closed-form expression of the gradient of $\mathcal{L}$}
\label{app:closed_form_gradient}
In the following proposition, we provide a closed-form expression of the gradient of the loss $\mathcal{L}$.
\begin{boxprop}[Closed-form gradient]
    \label{prop:closed_form_gradient}
    Let $\mathcal{L}$ be the loss function of Problem~\eqref{eq:neg_corr_linear_ensemble_with_reg}. For any $\mathbf{W} \in \mathbb{R}^{d \times M}$, the gradient of $\mathcal{L}$ in $\mathbf{W}$ writes
        \begin{equation}
    \label{eq:grad_formulation}
        \nabla \mathcal{L}(\mathbf{W}) = \frac{2}{M} \left[ \left( \Lambda + \frac{\mathbf{X}_\ell^\top \mathbf{X}_\ell}{n_\ell} \right)\mathbf{W} + 2\alpha_u \mathbf{X}_u^\top\mathbf{X}_u \mathbf{W} \left(\mathbf{U}^{[M]} - \mathbf{I}_M \right) -\frac{\mathbf{X}_\ell^\top \mathbf{Y}}{n_\ell}\right].
    \end{equation}
\end{boxprop}
\begin{proof}
        To compute $\nabla \mathcal{L} \colon \mathbb{R}^{d \times M} \to \mathbb{R}^{d \times M}$, we will rewrite the loss function $\mathcal{L}$ using the Frobenius inner product, that is the usual inner product on matrix spaces. We will use this formulation to write its Taylor expansion of order $1$ and identify $\nabla \mathcal{L}$. Using the formulation of Problem~\eqref{eq:neg_corr_linear_ensemble_with_reg} and the notations introduced in Eq.~\eqref{eq:notations}, we have:
        \begin{align*}
            \label{eq:loss_frobenius}
            \mathcal{L}(\mathbf{W}) &= \frac{1}{M} \sum_{m=1}^M \frac{1}{n_\ell} \sum_{i=1}^{n_\ell} \left (y_i - \boldsymbol{\omega}_m^\top\mathbf{x}_i\right) ^ 2 + \frac{1}{M} \sum_{m=1}^M \lambda_m \lVert \boldsymbol{\omega}_m \rVert_2^2 + \frac{2\alpha_u}{M} \sum_{m \neq k} \sum_{i=n_\ell+1}^{n_\ell +n_u} w_k^\top\mathbf{x}_i w_\ell^\top\mathbf{x}_i \\
            &= \frac{1}{n_\ell M} \sum_{m=1}^M \lVert \mathbf{y}_\ell - \mathbf{X}_\ell  \boldsymbol{\omega}_m \rVert _2^2 + \frac{1}{M} \sum_{m=1}^M \lambda_m \lVert \boldsymbol{\omega}_m \rVert_2^2  + \frac{2\alpha_u}{M} \sum_{m \neq k} \left(\mathbf{X}_u w_m \right) ^\top \left(\mathbf{X}_u w_k \right) \\ 
            &= \frac{1}{n_\ell M} \sum_{m=1}^M \left(\mathbf{y}_\ell - \mathbf{X}_\ell  \boldsymbol{\omega}_m \right)^\top\left(\mathbf{y}_\ell - \mathbf{X}_\ell  \boldsymbol{\omega}_m \rVert\right) + \frac{1}{M} \sum_{m=1}^M \left(\sqrt{\lambda_m} \boldsymbol{\omega}_m\right)^\top \left(\sqrt{\lambda_m} \boldsymbol{\omega}_m\right) \\
            & \quad+ \frac{2\alpha_u}{M} \sum_{m=1}^M \sum_{k=1}^M \left(\mathbf{X}_u w_m \right) ^\top \left(\mathbf{X}_u w_k \right) - 2\alpha_u \sum_{m=1}^M \left(\mathbf{X}_u w_m \right) ^\top \left(\mathbf{X}_u w_m \right) \\ 
            &= \frac{1}{n_\ell M} \lVert \mathbf{Y} - \mathbf{X}_\ell \mathbf{W} \rVert_\mathrm{F}^2 + \frac{1}{M} \lVert \mathbf{\Lambda}^{1/2} \mathbf{W}\rVert_\mathrm{F}^2 + \frac{2\alpha_u}{M} \mathbb{1}_M^\top \left(\mathbf{X}_u \mathbf{W} \right)^\top \left(\mathbf{X}_u \mathbf{W} \right) \mathbb{1}_M  - \frac{2\alpha_u}{M} \lVert \mathbf{X}_u \mathbf{W} \rVert_\mathrm{F}^2,
        \end{align*}
    where $\mathbf{Y} \in \mathbb{R}^{n_\ell \times M}$ is the matrix with the vector $\mathbf{y}_\ell$ repeated $M$ times as columns and $\mathbf{\Lambda} \in \mathbb{R}^{M \times M}$ is a diagonal matrix with entries $\lambda_m$. For the last equality, we used the fact that for any matrices $\mathbf{A}, \mathbf{B} \in \mathbb{R}^{d \times M}$ with columns $\mathbf{A}_{\cdot,m}, \mathbf{B}_{\cdot,m} \in \mathbb{R}^d$, the following property of the Frobenius inner product holds:
\begin{equation}
\label{eq:frobenius_reformulation}
    \langle \mathbf{A}, \mathbf{B} \rangle_\mathrm{F} = \mathrm{Tr}(\mathbf{A}^\top\mathbf{B}) = \sum_{m=1}^M \sum_{k=1}^d \mathbf{A}_{km}\mathbf{B}_{km} = \sum_{m=1}^M \mathbf{A}_{\cdot,m}^\top \mathbf{B}_{\cdot,m}.
\end{equation}
 Then, we write the first order Taylor expansion $\mathcal{L}$ near $\mathbf{W}$ by considering a small displacement $\mathbf{H} \in \mathbb{R}^{d \times M}$ and obtain that
\begin{align}
\label{eq:taylor_exp_loss_func}
    \mathcal{L}(\mathbf{W} + \mathbf{H}) &= \frac{1}{n_\ell M} \lVert \mathbf{Y} - \mathbf{X}_\ell \left( \mathbf{W} + \mathbf{H} \right) \rVert_\mathrm{F}^2 + \frac{1}{M} \lVert \mathbf{\Lambda}^{1/2} \mathbf{\left( \mathbf{W} + \mathbf{H} \right)}\rVert_\mathrm{F}^2 \notag + \frac{2\alpha_u}{M} \mathbb{1}_M^\top \left(\mathbf{X}_u \left(\mathbf{W} + \mathbf{H} \right) \right)^\top \left(\mathbf{X}_u \left(\mathbf{W} + \mathbf{H} \right) \right) \mathbb{1}_M \notag \\
    & \qquad - \frac{2\alpha_u}{M} \lVert \mathbf{X}_u \mathbf{\left( \mathbf{W} + \mathbf{H} \right)} \rVert_\mathrm{F}^2 \notag \\
    &= \frac{1}{n_\ell M} \lVert \mathbf{Y} - \mathbf{X}_\ell \mathbf{W} \rVert_\mathrm{F}^2 + \frac{1}{M} \lVert \mathbf{\Lambda}^{1/2} \mathbf{W}\rVert_\mathrm{F}^2 + \frac{2\alpha_u}{M} \mathbb{1}_M^\top \left(\mathbf{X}_u \mathbf{W} \right)^\top \left(\mathbf{X}_u \mathbf{W} \right) \mathbb{1}_M  \notag \\
    & \qquad - \frac{2\alpha_u}{M} \lVert \mathbf{X}_u \mathbf{W} \rVert_\mathrm{F}^2 - \langle \frac{2}{n_lM} \left( \mathbf{Y} - \mathbf{X}_\ell \mathbf{W} \right), \mathbf{X}_\ell \mathbf{H}\rangle_\mathrm{F} \notag \\
    & \qquad + \langle \frac{2}{M} \mathbf{\Lambda}^{1/2} \mathbf{W}, \mathbf{\Lambda}^{1/2} \mathbf{H}\rangle_\mathrm{F} + \frac{4\alpha_u}{M}  \mathbb{1}_M^\top \left(\mathbf{X}_u \mathbf{W} \right)^\top \left(\mathbf{X}_u \mathbf{H} \right) \mathbb{1}_M \notag \\
    & \qquad - \frac{4\alpha_u}{M} \langle \mathbf{X}_u \mathbf{W}, \mathbf{X}_u \mathbf{H}\rangle_\mathrm{F} \notag \\
    & \qquad + \underbrace{\frac{1}{n_\ell M} \lVert \mathbf{X}_\ell \mathbf{H} \rVert_\mathrm{F}^2 + \frac{1}{M} \lVert \mathbf{\Lambda}^{1/2} \mathbf{H}\rVert_\mathrm{F}^2 + \frac{2\alpha_u}{M} \mathbb{1}_M^\top \left(\mathbf{X}_u \mathbf{H} \right)^\top \left(\mathbf{X}_u \mathbf{H} \right) \mathbb{1}_M  - \frac{2\alpha_u}{M} \lVert \mathbf{X}_u \mathbf{H} \rVert_\mathrm{F}^2}_{o(\lVert \mathbf{H} \rVert_\mathrm{F})} \notag \\
    & = \mathcal{L}(\mathbf{W}) + \frac{2}{M} \langle \mathbf{\Lambda} \mathbf{W} - \frac{1}{n_\ell} \mathbf{X}_\ell^\top \left( \mathbf{Y} - \mathbf{X}_\ell \mathbf{W} \right) - 2\alpha_u \mathbf{X}_u^\top\mathbf{X}_u \mathbf{W}, \mathbf{H} \rangle_\mathrm{F} \notag \\
    & \qquad + \frac{4\alpha_u}{M}  \underbrace{\mathbb{1}_M^\top \left(\mathbf{X}_u \mathbf{W} \right)^\top \left(\mathbf{X}_u \mathbf{H} \right)\mathbb{1}_M}_{d(\mathbf{H})} + o(\lVert \mathbf{H} \rVert_\mathrm{F}).
\end{align}
Moreover, we have that
\begin{align*}
    d(\mathbf{H}) = \mathbb{1}_M^\top \left(\mathbf{X}_u \mathbf{W} \right)^\top \left(\mathbf{X}_u \mathbf{H} \right)\mathbb{1}_M &= \sum_{m=1}^M \sum_{k=1}^M \mathbf{W}_{\cdot, m}^\top \mathbf{X}_u^\top \mathbf{X}_u \mathbf{H}_{\cdot, k} \\
    &= \sum_{k=1}^M \left(\mathbf{X}_u^\top \mathbf{X}_u \sum_{m=1}^M \mathbf{W}_{\cdot, m} \right)^\top \mathbf{H}_{\cdot, k}  \\
    &= \sum_{k=1}^M \mathbf{L}_{\cdot, k}^\top \mathbf{H}_{\cdot, k} \\
    &= \langle \mathbf{L}, \mathbf{H} \rangle_\mathrm{F},
\end{align*}
where $\mathbf{L} \in \mathbb{R}^{d \times d}$ is the matrix with the vector $\mathbf{X}_u^\top \mathbf{X}_u \sum_{m=1}^M \mathbf{W}_{\cdot, m}$ repeated $d$ times as columns. Another way to write columns of $\mathbf{L}$ is the following:
\begin{equation*}
    \mathbf{L}_{\cdot, m} = \mathbf{X}_u^\top \mathbf{X}_u \sum_{k=1}^M \mathbf{W}_{\cdot, k} = \mathbf{X}_u^\top \mathbf{X}_u \mathbf{W} \mathbb{1}_M.
\end{equation*}
Hence, by introducing $\mathbf{U}^{[M]}$, the matrix of size $M$ full of ones, we obtain:
\begin{equation*}
    \mathbf{L} = \mathbf{X}_u^\top \mathbf{X}_u \mathbf{W} \mathbb{1}_M \mathbb{1}_M^\top = \mathbf{X}_u^\top \mathbf{X}_u \mathbf{W} \mathbf{U}^{[M]}.
\end{equation*}
It leads to:
    \begin{equation*}
        d(\mathbf{H}) = \langle \mathbf{X}_u^\top \mathbf{X}_u \mathbf{W} \mathbf{U}^{[M]}, \mathbf{H} \rangle_\mathrm{F}.
    \end{equation*}
    By injecting $d(\mathbf{H})$ into Eq.~\eqref{eq:taylor_exp_loss_func}, we obtain that
    \begin{equation*}
        \begin{split}
        \mathcal{L}(\mathbf{W} + \mathbf{H}) &= \mathcal{L}(\mathbf{W}) \\
        & \qquad + \langle \underbrace{\frac{2}{M} \left(\mathbf{\Lambda} \mathbf{W} - \frac{1}{n_\ell} \mathbf{X}_\ell^\top \left( \mathbf{Y} - \mathbf{X}_\ell \mathbf{W} \right) - 2\alpha_u \mathbf{X}_u^\top\mathbf{X}_u \mathbf{W} + 2\alpha_u \mathbf{X}_u^\top \mathbf{X}_u \mathbf{W} \mathbf{U}^{[M]}\right)}_{\nabla \mathcal{L}(\mathbf{W})}, \mathbf{H} \rangle_\mathrm{F} \\
        & \qquad + o(\lVert \mathbf{H} \rVert_\mathrm{F}).
        \end{split}
    \end{equation*}
    Finally, the gradient of $\mathcal{L}$ writes:
    \begin{equation*}
        \nabla \mathcal{L}(\mathbf{W}) = \frac{2}{M} \left[ \left( \Lambda + \frac{\mathbf{X}_\ell^\top \mathbf{X}_\ell}{n_\ell} \right)\mathbf{W} + 2\alpha_u \mathbf{X}_u^\top\mathbf{X}_u \mathbf{W} \left(\mathbf{U}^{[M]} - \mathbf{I}_M \right) -\frac{\mathbf{X}_\ell^\top \mathbf{Y}}{n_\ell}\right].
    \end{equation*}
\end{proof}

\subsection{Stationary points of $\mathcal{L}$ are solutions of a linear problem}
\label{app:stationary_point_euler}
In the following proposition, we show that the stationary points of $\mathcal{L}$ are the solutions of a linear problem in $\mathbf{W}$ and provide a closed-form expression of the columns of $\mathbf{W}$.
\begin{boxprop}[Stationary points of $\mathcal{L}$]
    \label{prop:sol_P_linear_problem}
    Let $\mathbf{Y} \in \mathbb{R}^{n_\ell \times M}$ be the matrix with the vector $\mathbf{y}_\ell$ repeated $M$ times as columns and $\mathbf{\Lambda} \in \mathbb{R}^{M \times M}$ be a diagonal matrix with entries $\lambda_m$. Solving Eq.~\eqref{eq:stationary_point_euler} amounts to solving a linear problem in $\mathbf{W}$ and the columns $\boldsymbol{\omega}_m$ of any stationary point $\mathbf{W}$ of $\mathcal{L}$ verify
\begin{equation}
    \label{eq:sol_P_linear_problem_columns}
    \forall m \in \llbracket 1,M \rrbracket, \quad \left( \lambda_m \mathbf{I}_d + \frac{\mathbf{X}_\ell^\top \mathbf{X}_\ell}{n_\ell} \right)\boldsymbol{\omega}_m = \frac{\mathbf{X}_\ell^\top \mathbf{y}_\ell}{n_\ell} - \frac{\gamma}{n_u(M-1)} \mathbf{X}_u^\top\mathbf{X}_u \sum_{k=1 | k \neq m}^M \boldsymbol{\omega}_k.
\end{equation}
\end{boxprop}
\begin{proof}
    We first recall that stationary points of $\mathcal{L}$ verify the Euler equation
        \begin{equation}
        \label{eq:recall_euler_eq}
        \nabla \mathcal{L}(\boldsymbol{W}) = 0.
    \end{equation}
Using Proposition~\ref{prop:closed_form_gradient}, we combine Eq.~\eqref{eq:grad_formulation} and Eq.~\eqref{eq:recall_euler_eq}, recalling that $\alpha_u = \frac{\gamma}{2n_u(M-1)}$, and deduce that any stationary point $\mathbf{W}$ is characterized by the following linear problem:
    \begin{equation*}
        \left( \Lambda + \frac{\mathbf{X}_\ell^\top \mathbf{X}_\ell}{n_\ell} \right)\mathbf{W} + \frac{\gamma}{n_u(M-1)} \mathbf{X}_u^\top\mathbf{X}_u \mathbf{W} \left(\mathbf{U}^{[M]} - \mathbf{I}_M \right) = \frac{\mathbf{X}_\ell^\top \mathbf{Y}}{n_\ell}.
    \end{equation*}
Moreover, this matrix equality holds if and only if it holds at the column level. Hence, we have
    \begin{align*}
    & \left( \Lambda + \frac{\mathbf{X}_\ell^\top \mathbf{X}_\ell}{n_\ell} \right)\mathbf{W} + \frac{\gamma}{n_u(M-1)} \mathbf{X}_u^\top\mathbf{X}_u \mathbf{W} \left(\mathbf{U}^{[M]} - \mathbf{I}_M \right) = \frac{\mathbf{X}_\ell^\top \mathbf{Y}}{n_\ell} \\
        \iff & \forall m \in \llbracket 1,M \rrbracket, \quad \left( \lambda_m \mathbf{I}_d + \frac{\mathbf{X}_\ell^\top \mathbf{X}_\ell}{n_\ell} \right)\boldsymbol{\omega}_m + \frac{\gamma}{n_u(M-1)} \mathbf{X}_u^\top\mathbf{X}_u \left( \sum_{k=1}^k \boldsymbol{\omega}_k - \boldsymbol{\omega}_m \right) = \frac{\mathbf{X}_\ell^\top \mathbf{y}_\ell}{n_\ell} \\
        \iff &  \forall m \in \llbracket 1,M \rrbracket, \quad \left( \lambda_m \mathbf{I}_d + \frac{\mathbf{X}_\ell^\top \mathbf{X}_\ell}{n_\ell} \right)\boldsymbol{\omega}_m = \frac{\mathbf{X}_\ell^\top \mathbf{y}_\ell}{n_\ell} - \frac{\gamma}{n_u(M-1)} \mathbf{X}_u^\top\mathbf{X}_u \sum_{k=1 | k \neq m}^M \boldsymbol{\omega}_k. 
    \end{align*}
\end{proof}

\subsection{Proof of Proposition~\ref{prop:loss_func_properties}}
\label{app:prop_loss_func}
We start by recalling the definition of coercivity.
\begin{boxdef}[Coercivity]
\label{def:coercivity}
A bilinear form $a \colon \mathbf{H} \times \mathbf{H} \mapsto \mathbb{R}$, where $\mathbf{H}$ is a Hilbert space, is called coercive if there exists $\gamma > 0$ such that:
\begin{equation}
    \label{eq:coercivity_def}
    \forall \mathbf{x} \in \mathbf{H}, a(\mathbf{x}, \mathbf{x}) \geq \gamma \lVert \mathbf{x} \rVert ^2.
\end{equation}
\end{boxdef}

Then, we prove the following technical lemmas.

\begin{boxlem}{\citep[see][chap. 3, p. 74]{citeulike:163662}.}
\label{lemma:real_line_cvx}
    Let $\mathbf{E}$ be a vector space. A function $f \colon \mathbf{E} \to \mathbb{R}$ is convex if and only if for all $\mathbf{x} \in \mathrm{dom}(f)$ and all $\mathbf{v} \in E$, the function $g \colon \mathbb{R} \to \mathbb{R}, t \mapsto f(\mathbf{x} + t\mathbf{v})$ is convex on its domain $\{t \in \mathbb{R} | \mathbf{x} + t\mathbf{v} \in \mathrm{dom}(f)\}$.
\end{boxlem}
\begin{rmk}
    By construction, $\mathrm{dom(f)}$ is a convex set if and only if all the $\mathrm{dom(g)}$ are convex sets.
\end{rmk}

\begin{proof}
The implication part is straightforward as composing by an affine function preserves the convexity and that the convexity of $\mathrm{dom}(f)$ induces the convexity of all the $\mathrm{dom}(g)$. We will now prove the converse. We assume that for all $\mathbf{x} \in \mathrm{dom}(f)$ and all $\mathbf{v} \in E$, the function $g \colon t \mapsto f(\mathbf{x} + t\mathbf{v})$ is convex on its domain i.e., for all $\mathbf{x} \in \mathrm{dom}(f), v \in \mathbf{E}$, we have for all $\alpha \in [0, 1]$ and for all $t, t' \in \mathrm{dom}(g)$:
\begin{align}
\label{eq:g_cvx}
    & g(\alpha t + (1- \alpha)t') \leq \alpha g(t) + (1-\alpha)g(t') \notag \\
    & \iff f(\mathbf{x} + (\alpha t + (1- \alpha)t')\mathbf{v}) \leq \alpha f(\mathbf{x} + tv) + (1-\alpha)f(\mathbf{x} + t'\mathbf{v}).
\end{align}
Let $\theta \in [0,1]$ and $ \mathbf{u}, \mathbf{y} \in \mathrm{dom}(f)$. By assumption, all the $\mathrm{dom}(g)$ are convex sets, so $\mathrm{dom}(f)$ is a convex set. We can apply Eq.~\eqref{eq:g_cvx} with $\mathbf{x} = \mathbf{u} \in \mathrm{dom}(f), \mathbf{v}=\mathbf{y}-\mathbf{u} \in \mathbf{E}, \alpha = \theta \in [0, 1], t =0, t' = 1$. Indeed, $\mathbf{x} + t\mathbf{v} = \mathbf{u} \in \mathrm{dom}(f), \mathbf{x} + t'\mathbf{v} = y \in \mathrm{dom}(f)$, ensuring that $t, t' \in \mathrm{dom}(g)$. We obtain:
\begin{align*}
    & f(\mathbf{x} + (\alpha t + (1- \alpha)t')\mathbf{v}) \leq \alpha f(\mathbf{x} + t\mathbf{v}) + (1-\alpha)f(\mathbf{x} + t'\mathbf{v}) \\
    & \iff f(\theta \mathbf{u} + (1-\theta)\mathbf{y}) \leq \theta f(\mathbf{x}) + (1-\theta)f(\mathbf{y}) .
\end{align*}
Hence, $f$ is convex.
\end{proof}

\begin{boxlem}
\label{lemma:eigenvalues_psd}
    For any positive semi-definite matrix $\mathbf{S} \in \mathbb{R}^{d \times d}$ with minimum and maximum eigenvalues $\lambda_{\mathrm{min}}, \lambda_{\mathrm{max}}$ respectively, we have:
    \begin{equation}
        \label{eq:eigenvalues_psd_ineq}
        \forall \boldsymbol{\omega} \in \mathbb{R}^d, \lambda_{\mathrm{min}} \lVert \boldsymbol{\omega}\rVert_2^2  \leq \boldsymbol{\omega}^\top \mathbf{S} \boldsymbol{\omega} \leq \lambda_{\mathrm{max}} \lVert \boldsymbol{\omega}\rVert_2^2.
    \end{equation}
\end{boxlem}
\begin{proof}
    Let $\mathbf{S} \in \mathbb{R}^{d \times d}$ be a positive semi-definite matrix. Let $\lambda_{\mathrm{max}} = \lambda_1 \geq \dots \geq \lambda_d = \lambda_{\mathrm{min}} \geq 0$ be the eigenvalues sorted in decreasing order. The spectral theorem ensures the existence of $\mathbf{U} \in \mathbb{R}^{d \times d}$ orthogonal and $\mathbf{D} \in \mathbb{R}^{d \times d}$ diagonal with entries $\mathbf{D}_{ii} = \lambda_i$ such that:
    \begin{equation*}
        \mathbf{S} = \mathbf{U}^\top \mathbf{D} \mathbf{U}.
    \end{equation*}

Let $\boldsymbol{\omega} \in \mathbb{R}^d$. We have:
\begin{align*}
    \boldsymbol{\omega}^\top\mathbf{S}\boldsymbol{\omega} &= \boldsymbol{\omega}^\top\mathbf{U}^\top \mathbf{D} \mathbf{U}\boldsymbol{\omega} \\
    &= \left(\mathbf{U}\boldsymbol{\omega}\right)^\top \mathbf{D} \mathbf{U}\boldsymbol{\omega} \\
    &= \sum_{k=1}^d \underbrace{\lambda_k}_{\geq 0}\underbrace{\left(\mathbf{U}\boldsymbol{\omega}\right)_k^2}_{\geq 0}.
\end{align*}
Hence, we deduce:
\begin{equation}
\label{eq:eigen_ineq}
    \lambda_{\mathrm{min}} \lVert \mathbf{U}\boldsymbol{\omega} \rVert ^2_2 = \lambda_{\mathrm{min}} \sum_{k=1}^d \left(\mathbf{U}\boldsymbol{\omega}\right)_k^2 \leq \boldsymbol{\omega}^\top\mathbf{S}\boldsymbol{\omega} \leq \lambda_{\mathrm{max}} \sum_{k=1}^d \left(\mathbf{U}\boldsymbol{\omega}\right)_k^2 = \lambda_{\mathrm{max}} \lVert \mathbf{U}\boldsymbol{\omega} \rVert ^2_2.
\end{equation}

Using the fact that $\mathbf{U}$ is orthogonal, we know that: 
\begin{equation}
\label{eq:U_orthogonal}
    \lVert \mathbf{U}\boldsymbol{\omega} \rVert ^2_2 = \boldsymbol{\omega}^\top \underbrace{\mathbf{U}^\top \mathbf{U}}_{= \mathbf{I}_d} \boldsymbol{\omega} = \boldsymbol{\omega}^\top\boldsymbol{\omega} = \lVert \boldsymbol{\omega} \rVert ^2_2.
\end{equation}
Finally, we combine Eq.~\eqref{eq:eigen_ineq} and Eq.~\eqref{eq:U_orthogonal} to obtain the desired inequalities:
\begin{equation*}
    \lambda_{\mathrm{min}}  \lVert \boldsymbol{\omega} \rVert ^2_2 \leq \boldsymbol{\omega}^\top\mathbf{S}\boldsymbol{\omega} \leq \lambda_{\mathrm{max}}  \lVert \boldsymbol{\omega} \rVert ^2_2.
\end{equation*}
\end{proof}
Then, we show that the loss function in Problem~\eqref{eq:neg_corr_linear_ensemble_with_reg} can be reformulated as
\begin{equation} 
\label{eq:loss_reformulation}
\begin{split}
    \mathcal{L}(\mathbf{W}) &= \frac{1}{M} \sum_{m=1}^M \left[ \frac{1}{n_\ell} \lVert \mathbf{y}_\ell - \mathbf{X}_\ell  \boldsymbol{\omega}_m \rVert _2^2 + \boldsymbol{\omega}_m^\top \left( \lambda_m\mathbf{I}_d - \frac{\gamma(M+1)}{n_u(M-1)} \mathbf{X}_u^\top \mathbf{X}_u \right) \boldsymbol{\omega}_m \right]\\
    & \qquad + \frac{\gamma}{2n_uM(M-1)} \sum_{m=1}^M\sum_{k=1}^M \left (\boldsymbol{\omega}_m + \boldsymbol{\omega}_k \right)^\top \mathbf{X}_u^\top \mathbf{X}_u \left( \boldsymbol{\omega}_m + \boldsymbol{\omega}_k \right).
\end{split}
\end{equation}
\begin{proof}
We have that
\begin{align}
\label{eq:proof_compact_loss}
    \mathcal{L}(\mathbf{W}) &= \frac{1}{M} \sum_{m=1}^M \frac{1}{n_\ell} \sum_{i=1}^{n_\ell} \left (y_i - \boldsymbol{\omega}_m^\top\mathbf{x}_i\right) ^ 2 + \frac{1}{M} \sum_{m=1}^M \lambda_m \lVert \boldsymbol{\omega}_m \rVert_2^2  \notag \\
    & \qquad + \frac{\gamma}{n_uM(M-1)} \sum_{m \neq k} \sum_{i=n_\ell+1}^{n_\ell +n_u} \boldsymbol{\omega}_m^\top\mathbf{x}_i \boldsymbol{\omega}_k^\top\mathbf{x}_i \notag \\ 
    &= \frac{1}{M} \sum_{m=1}^M \frac{1}{n_\ell} \lVert \mathbf{y}_\ell - \mathbf{X}_\ell  \boldsymbol{\omega}_m \rVert _2^2 + \frac{1}{M} \sum_{m=1}^M \lambda_m \boldsymbol{\omega}_m^\top\boldsymbol{\omega}_m \notag \\
    & \qquad + \frac{\gamma}{n_uM(M-1)} \sum_{m \neq k} \left(\mathbf{X}_u \boldsymbol{\omega}_m \right)^\top \left(\mathbf{X}_u \boldsymbol{\omega}_k \right) \notag \\ 
    &= \frac{1}{M} \sum_{m=1}^M \frac{1}{n_\ell} \lVert \mathbf{y}_\ell - \mathbf{X}_\ell  \boldsymbol{\omega}_m \rVert _2^2 + \frac{1}{M} \sum_{m=1}^M \lambda_m \boldsymbol{\omega}_m^\top\boldsymbol{\omega}_m \notag \\
    & \qquad + \frac{\gamma}{n_uM(M-1)} \sum_{m \neq k} \boldsymbol{\omega}_m^\top \mathbf{X}_u^\top \mathbf{X}_u \boldsymbol{\omega}_k.
\end{align}
Using the fact that $\boldsymbol{\omega}_m^\top \mathbf{X}_u^\top \mathbf{X}_u \boldsymbol{\omega}_k$ is a real number, it is equal to its transpose term. We deduce
\begin{align*}
    \boldsymbol{\omega}_m^\top \mathbf{X}_u^\top \mathbf{X}_u \boldsymbol{\omega}_k &= \frac{1}{2}\left[\boldsymbol{\omega}_m^\top \mathbf{X}_u^\top \mathbf{X}_u \boldsymbol{\omega}_k + \boldsymbol{\omega}_k^\top \mathbf{X}_u^\top \mathbf{X}_u \boldsymbol{\omega}_m \right] \\
    &= \frac{1}{2}\left[ \left (\boldsymbol{\omega}_m + \boldsymbol{\omega}_k \right)^\top \mathbf{X}_u^\top \mathbf{X}_u \left( \boldsymbol{\omega}_m + \boldsymbol{\omega}_k \right) - \boldsymbol{\omega}_m^\top \mathbf{X}_u^\top \mathbf{X}_u \boldsymbol{\omega}_m - \boldsymbol{\omega}_k^\top \mathbf{X}_u^\top \mathbf{X}_u \boldsymbol{\omega}_k\right].
\end{align*}
Thus, by summing over $\{m \neq k\}$, we obtain:
\begin{align*}
    \sum_{m \neq k} \boldsymbol{\omega}_m^\top \mathbf{X}_u^\top \mathbf{X}_u \boldsymbol{\omega}_k 
    &= \sum_{m \neq k} \frac{1}{2}\left[ \left (\boldsymbol{\omega}_m + \boldsymbol{\omega}_k \right)^\top \mathbf{X}_u^\top \mathbf{X}_u \left( \boldsymbol{\omega}_m + \boldsymbol{\omega}_k \right) - \boldsymbol{\omega}_m^\top \mathbf{X}_u^\top \mathbf{X}_u \boldsymbol{\omega}_m - \boldsymbol{\omega}_k^\top \mathbf{X}_u^\top \mathbf{X}_u \boldsymbol{\omega}_k\right] \\ 
    &= \frac{1}{2} \sum_{m \neq k} \left (\boldsymbol{\omega}_m + \boldsymbol{\omega}_k \right)^\top \mathbf{X}_u^\top \mathbf{X}_u \left( \boldsymbol{\omega}_m + \boldsymbol{\omega}_k \right) - (M-1) \sum_{m=1}^M \boldsymbol{\omega}_m^\top \mathbf{X}_u^\top \mathbf{X}_u \boldsymbol{\omega}_m \\ 
    &= \frac{1}{2} \sum_{m, k} \left (\boldsymbol{\omega}_m + \boldsymbol{\omega}_k \right)^\top \mathbf{X}_u^\top \mathbf{X}_u \left( \boldsymbol{\omega}_m + \boldsymbol{\omega}_k \right) \\
    & \qquad - \frac{1}{2} \sum_{m=1}^M (2\boldsymbol{\omega}_m)^\top \mathbf{X}_u^\top \mathbf{X}_u (2\boldsymbol{\omega}_m) - (M-1) \sum_{m=1}^M \boldsymbol{\omega}_m^\top \mathbf{X}_u^\top \mathbf{X}_u \boldsymbol{\omega}_m \\
    &= \frac{1}{2} \sum_{m, k} \left (\boldsymbol{\omega}_m + \boldsymbol{\omega}_k \right)^\top \mathbf{X}_u^\top \mathbf{X}_u \left( \boldsymbol{\omega}_m + \boldsymbol{\omega}_k \right) - (M+1) \sum_{m=1}^M \boldsymbol{\omega}_m^\top \mathbf{X}_u^\top \mathbf{X}_u \boldsymbol{\omega}_m.
\end{align*}
We now inject this term in Eq.~\eqref{eq:proof_compact_loss} and gather quadratic and cross terms to obtain
\begin{align*}
    \mathcal{L}(\mathbf{W}) &= \frac{1}{M} \sum_{m=1}^M \frac{1}{n_\ell} \lVert \mathbf{y}_\ell - \mathbf{X}_\ell  \boldsymbol{\omega}_m \rVert _2^2 + \frac{1}{M} \sum_{m=1}^M \lambda_m \boldsymbol{\omega}_m^\top\boldsymbol{\omega}_m \\
    & \qquad + \frac{\gamma}{n_uM(M-1)} \left [ \frac{1}{2} \sum_{m, k} \left (\boldsymbol{\omega}_m + \boldsymbol{\omega}_k \right)^\top \mathbf{X}_u^\top \mathbf{X}_u \left( \boldsymbol{\omega}_m + \boldsymbol{\omega}_k \right) - (M+1) \sum_{m=1}^M \boldsymbol{\omega}_m^\top \mathbf{X}_u^\top \mathbf{X}_u \boldsymbol{\omega}_m \right] \\
   &=  \frac{1}{M} \sum_{m=1}^M \frac{1}{n_\ell} \lVert \mathbf{y}_\ell - \mathbf{X}_\ell  \boldsymbol{\omega}_m \rVert _2^2 \\
     & \qquad + \frac{1}{M} \sum_{m=1}^M \boldsymbol{\omega}_m^\top \left(\lambda_m \mathbf{I}_d - \frac{\gamma(M+1)}{n_u(M-1)} \mathbf{X}_u^\top \mathbf{X}_u \right) \boldsymbol{\omega}_m \\
     & \qquad + \frac{\gamma}{2n_uM(M-1)} \sum_{m=1}^M\sum_{k=1}^M \left (\boldsymbol{\omega}_m + \boldsymbol{\omega}_k \right)^\top \mathbf{X}_u^\top \mathbf{X}_u \left( \boldsymbol{\omega}_m + \boldsymbol{\omega}_k \right).
\end{align*}
\end{proof}
We now proceed to the proof of Proposition~\ref{prop:loss_func_properties}.
\begin{proof}
We assume that assumption~\ref{assumption:pd} holds.
Using the formulation of the loss in Eq.~\eqref{eq:loss_reformulation} and the quantities $\mathbf{S}_u, \alpha_u$ introduced in Eq.~\eqref{eq:notations}, we can decompose it as follows:
    \begin{align*}
    \mathcal{L}(\mathbf{W}) &= \underbrace{\frac{1}{M} \sum_{m=1}^M \frac{1}{n_\ell} \lVert \mathbf{y}_\ell - \mathbf{X}_\ell  \boldsymbol{\omega}_m \rVert _2^2}_{\ell_1(\mathbf{W})} + \underbrace{\frac{1}{M} \sum_{m=1}^M \boldsymbol{\omega}_m^\top \left( \lambda_m\mathbf{I}_d - \mathbf{S}_u \right) \boldsymbol{\omega}_m}_{\ell_2(\mathbf{W})} \\
    & \qquad + \underbrace{\frac{\alpha_u}{M} \sum_{m=1}^M  \sum_{k=1}^M \left (\boldsymbol{\omega}_m + \boldsymbol{\omega}_k \right)^\top \mathbf{X}_u^\top \mathbf{X}_u \left( \boldsymbol{\omega}_m + \boldsymbol{\omega}_k \right)}_{\ell_3(\mathbf{W})}.
    \end{align*}

\paragraph{Continuity and differentiability of $\mathcal{L}$.} The losses $\ell_1, \ell_2$ and $\ell_3$ are differentiable w.r.t $\mathbf{W}$, leading to $\mathcal{L}$ being differentiable. Thus, $\mathcal{L}$ is continuous on $\mathbb{R}^{d \times M}$.

\paragraph{Strict Convexity of $\mathcal{L}$.}
We will show that $\ell_1, \ell_2, \ell_3$ are convex and that, in addition, $\ell_2$ is strictly convex. This will lead to the strict convexity of $\mathcal{L}$ on the convex set $\mathbb{R}^{d \times M}$.

\begin{itemize}
    \item[-] $\lVert \cdot \rVert _2^2\colon\mathbb{R}^d \to \mathbb{R}$ is convex as it is a norm function. For all $m \in \llbracket 1, M \rrbracket$, we define the affine function
    \begin{align*}
      a^m \colon \mathbb{R}^{d \times M} &\to \mathbb{R}^{d}\\
      \mathbf{W} & \mapsto \mathbf{y}_\ell - \mathbf{X}_\ell\boldsymbol{\omega}_m.
    \end{align*}
As composing by an affine function preserves the convexity, we deduce that for all $m \in \llbracket 1, M \rrbracket$, 
    \begin{align*}
      \ell_1^m \colon \mathbb{R}^{d \times M} &\to \mathbb{R}\\
      \mathbf{W} & \mapsto \lVert \mathbf{y}_\ell - \mathbf{X}_\ell\boldsymbol{\omega}_m \rVert ^2_2.
    \end{align*}
is convex. By non-negative weighted summation, we obtain that $\ell_1\colon\mathbb{R}^{d \times M} \to \mathbb{R}$ is convex.

    \item[-] For a given matrix $\mathbf{P} \in \mathbb{R}^{d \times d}$, we define 
    \begin{align*}
      \lVert \cdot \rVert ^2_{\mathbf{P}} \colon \mathbb{R}^{d} &\to \mathbb{R}\\
      \boldsymbol{\omega} & \mapsto \boldsymbol{\omega}^\top \mathbf{P} \boldsymbol{\omega}.
    \end{align*}
Such a function is twice differentiable with Hessian $\mathbf{P}$. Using the property of differentiable convex functions, $\lVert \cdot \rVert ^2_{\mathbf{P}}$ is convex (respectively strictly convex) if and only if $\mathbf{P}$ is positive semi-definite  (respectively positive definite). Using Assumption~\ref{assumption:pd}, we know that for all $m \in \llbracket 1,M \rrbracket$, $\lVert \cdot \rVert ^2_{\lambda_m\mathbf{I}_d - \mathbf{S}_u}$ is strictly convex. As before, for all $m \in \llbracket 1, M \rrbracket$, we can define the linear function 
    \begin{align*}
      e^m \colon \mathbb{R}^{d \times M} &\to \mathbb{R}^{d}\\
      \mathbf{W} & \mapsto \boldsymbol{\omega}_m.
    \end{align*}
As composing by a linear function preserves the strict convexity (provided that this linear function is not identically equal to zero), we deduce that for all $m \in \llbracket 1, M \rrbracket$,
    \begin{align*}
      \ell_2^m \colon \mathbb{R}^{d \times M} &\to \mathbb{R}\\
      \mathbf{W} & \mapsto \boldsymbol{\omega}_m^\top \left( \lambda_m\mathbf{I}_d - \mathbf{S}_u \right)\boldsymbol{\omega}_m.
    \end{align*}
is strictly convex. By non-negative weighted summation of strictly convex functions, we obtain that $\ell_2\colon\mathbb{R}^{d \times M} \to \mathbb{R}$ is strictly convex.

\item[-] Let $\mathbf{W}, \mathbf{H} \in \mathbb{R}^{d \times M}$. We consider
    \begin{align*}
      f \colon \mathbb{R} &\to \mathbb{R}\\
      t & \mapsto \ell_3(\mathbf{W} + t\mathbf{H}).
    \end{align*}
We will show that $f$ is convex. Following Lemma~\ref{lemma:real_line_cvx}, as $\mathbf{W}$ and $\mathbf{H}$ are taken arbitrary in $\mathbb{R}^{d \times M}$, it will induce the convexity of $\ell_3$. Let $t \in \mathbb{R}$. We have:
\begin{align*}
    f(t) &= \ell_3(\mathbf{W} + t\mathbf{H}) \\
         &= \frac{\alpha_u}{M} \sum_{m=1}^M\sum_{k=1}^M \left (\boldsymbol{\omega}_m + \boldsymbol{\omega}_k + t(\mathbf{h}_m + \mathbf{h}_k) \right)^\top \mathbf{X}_u^\top \mathbf{X}_u \left(\boldsymbol{\omega}_m + \boldsymbol{\omega}_k + t(\mathbf{h}_m + \mathbf{h}_k)\right) \\
         &= t^2 \times \frac{\alpha_u}{M} \sum_{m=1}^M\sum_{k=1}^M \left(\mathbf{h}_m + \mathbf{h}_k \right)^\top \mathbf{X}_u^\top \mathbf{X}_u \left(\mathbf{h}_m + \mathbf{h}_k\right) \\
         & \qquad + 2t \times \frac{\alpha_u}{M} \sum_{m=1}^M\sum_{k=1}^M \left(\boldsymbol{\omega}_m + \boldsymbol{\omega}_k \right)^\top \mathbf{X}_u^\top \mathbf{X}_u \left(\mathbf{h}_m + \mathbf{h}_k\right) \\
         & \qquad +
         \frac{\alpha_u}{M} \sum_{m=1}^M\sum_{k=1}^M \left(\boldsymbol{\omega}_m + \boldsymbol{\omega}_k \right)^\top \mathbf{X}_u^\top \mathbf{X}_u \left(\boldsymbol{\omega}_m + \boldsymbol{\omega}_k\right) \\
         &= at^2 + bt + c.
\end{align*}
where 
\begin{equation*}
    \begin{cases}
    a = \frac{\alpha_u}{M}\sum_{m=1}^M\sum_{k=1}^M \left(\mathbf{h}_m + \mathbf{h}_k \right)^\top \mathbf{X}_u^\top \mathbf{X}_u \left(\mathbf{h}_m + \mathbf{h}_k\right) \\
    b = \frac{2\alpha_u}{M}\sum_{m=1}^M\sum_{k=1}^M \left(\boldsymbol{\omega}_m + \boldsymbol{\omega}_k \right)^\top \mathbf{X}_u^\top \mathbf{X}_u \left(\mathbf{h}_m + \mathbf{h}_k\right) \\
    c = \frac{\alpha_u}{M}\sum_{m=1}^M\sum_{k=1}^M \left(\boldsymbol{\omega}_m + \boldsymbol{\omega}_k \right)^\top \mathbf{X}_u^\top \mathbf{X}_u \left(\boldsymbol{\omega}_m + \boldsymbol{\omega}_k\right).
\end{cases}
\end{equation*}

We see that $f$ is a second-order polynomial function of the real line. It is a convex function if and only if $a \geq 0$. It should be noted that $\mathbf{X}_u^\top \mathbf{X}_u$ is symmetric positive semi-definite by construction. Indeed, we have: 
$$ \forall \boldsymbol{\omega} \in \mathbb{R}^d, \boldsymbol{\omega}^\top \mathbf{X}_u^\top \mathbf{X}_u \boldsymbol{\omega} = \lVert \mathbf{X}_u \boldsymbol{\omega} \rVert ^2_2 \geq 0.$$
Hence, we deduce:
\begin{equation*}
    a = \sum_{m=1}^M\sum_{k=1}^M \underbrace{\left(\mathbf{h}_m + \mathbf{h}_k \right)^\top \mathbf{X}_u^\top \mathbf{X}_u \left(\mathbf{h}_m + \mathbf{h}_k\right)}_{\geq 0} \geq 0.
\end{equation*}
We have shown that $f$ is a convex function of the real line. As $\mathbf{W}$ and $\mathbf{H}$ were taken arbitrarily, we obtain that $\ell_3\colon\mathbb{R}^{d \times M} \to \mathbb{R}$ is convex.
 
By summation of convex functions and a strictly convex function, we finally obtain that $\mathcal{L}\colon\mathbb{R}^{d \times M} \to \mathbb{R}$ is strictly convex.
\end{itemize}

\paragraph{Coercivity of $\mathcal{L}$.}
We will show that under assumptions~\ref{assumption:pd}, $\mathcal{L}$ is lower-bounded by a coercive function, implying the coercivity of $\mathcal{L}$ i.e 
\begin{equation*}
    \lim_{\lVert \mathbf{W} \rVert_\mathrm{F} \to +\infty} \mathcal{L}(\mathbf{W}) = +\infty.
\end{equation*}
where $\lVert \cdot \rVert_\mathrm{F}$ is the norm associated to the Frobenius inner product 
    \begin{align*}
      \langle \cdot, \cdot \rangle_\mathrm{F} \colon \mathbb{R}^{d \times M} \times  \mathbb{R}^{d \times M} &\to \mathbb{R}\\
      (\mathbf{A}, \mathbf{B}) & \mapsto \mathrm{Tr}(\mathbf{A}^\top \mathbf{B}).
    \end{align*}
\begin{itemize}
    \item[-] The function $\ell_1$ is non-negative as the sum of non-negative functions.
    \item[-] By construction, $\mathbf{X}_u^\top\mathbf{X}_u$ is symmetric positive semi-definite. Hence, we have:
    $$
    \forall m, k \in \llbracket 1,M \rrbracket, \quad
    \left(\boldsymbol{\omega}_m + \boldsymbol{\omega}_k \right)^\top \mathbf{X}_u^\top \mathbf{X}_u \left( \boldsymbol{\omega}_m + \boldsymbol{\omega}_k \right) \geq 0.
    $$
    It leads to $\ell_3$ being non-negative as the sum of non-negative functions.
    \item[-] By lower bounding, we obtain:
    \begin{equation}
    \label{eq:loss_func_ineq}
        \mathcal{L}(\mathbf{W}) = \underbrace{\ell_1(\mathbf{W})}_{\geq 0} + \ell_2(\mathbf{W}) + \underbrace{\ell_3(\mathbf{W})}_{\geq 0} \geq \ell_2(\mathbf{W}) = \frac{1}{M} \sum_{m=1}^M \boldsymbol{\omega}_m^\top \left( \lambda_m\mathbf{I}_d - \mathbf{S}_u \right)\boldsymbol{\omega}_m.
    \end{equation}
Under Assumption \ref{assumption:pd}, we know that for all $m \in \llbracket 1, M \rrbracket$, $\lambda_m\mathbf{I}_d - \mathbf{S}_u$ is positive definite. It ensures that $\lambda_{\mathrm{min}}(\lambda_m\mathbf{I}_d - \mathbf{S}_u)$, the minimum eigenvalue of $\lambda_m\mathbf{I}_d - \mathbf{S}_u$, is positive. Hence, we have:

\begin{align*}
    \ell_2(\mathbf{W}) &= \frac{1}{M} \sum_{m=1}^M \underbrace{\boldsymbol{\omega}_m^\top \left( \lambda_m\mathbf{I}_d - \mathbf{S}_u \right)\boldsymbol{\omega}_m}_{\geq \lambda_{\mathrm{min}}(\lambda_m\mathbf{I}_d - \mathbf{S}_u) \boldsymbol{\omega}_m^\top \boldsymbol{\omega}_m} \tag{from Lemma~\ref{lemma:eigenvalues_psd}} \\
    & \geq \frac{1}{M} \sum_{m=1}^M \underbrace{\lambda_{\mathrm{min}}(\lambda_m\mathbf{I}_d - \mathbf{S}_u)}_{> 0} \underbrace{\boldsymbol{\omega}_m^\top \boldsymbol{\omega}_m}_{\geq 0} \\
    & \geq \underbrace{\frac{1}{M} \min_{m \in \llbracket 1, M \rrbracket} \{ \lambda_{\mathrm{min}}(\lambda_m\mathbf{I}_d - \mathbf{S}_u)\}}_{ > 0} \sum_{m=1}^M \boldsymbol{\omega}_m^\top \boldsymbol{\omega}_m \\
    & \geq \gamma \sum_{m=1}^M \boldsymbol{\omega}_m^\top \boldsymbol{\omega}_m \text{ where } \gamma = \frac{1}{M} \min_{m \in \llbracket 1, M \rrbracket} \{ \lambda_{\mathrm{min}}(\lambda_m\mathbf{I}_d - \mathbf{S}_u)\} > 0.
\end{align*}

As we know that the columns of $\mathbf{W} \in \mathbb{R}^{d \times M}$ are the $\boldsymbol{\omega}_m \in \mathbb{R}^d$, we can use Eq.~\eqref{eq:frobenius_reformulation} and have that
\begin{equation*}
    \sum_{m=1}^M \boldsymbol{\omega}_m^\top \boldsymbol{\omega}_m = \langle \mathbf{W}, \mathbf{W} \rangle_\mathrm{F}.
\end{equation*}
By using the previous lower bound on $\ell_2$ in Eq.~\eqref{eq:loss_func_ineq}, we obtain:
\begin{equation*}
    \mathcal{L}(\mathbf{W}) \geq \gamma \langle \mathbf{W}, \mathbf{W} \rangle_\mathrm{F} \coloneqq a(\mathbf{W}, \mathbf{W}) \text{ where } a(\mathbf{A}, \mathbf{B}) = \gamma \langle \mathbf{A}, \mathbf{B} \rangle_\mathrm{F}. 
\end{equation*}

Following Definition~\ref{def:coercivity}, it is straightforward that $a$ is a coercive bilinear form on the Hilbert $\mathbb{R}^{d \times M}$:
\begin{equation*}
    a(\mathbf{W}, \mathbf{W}) = \gamma \langle \mathbf{W}, \mathbf{W} \rangle_\mathrm{F} = \gamma \lVert \mathbf{W} \rVert^2_\mathrm{F} \to +\infty.
\end{equation*}
Hence by lower bounding, we obtain:
\begin{equation*}
    \lim_{\lVert \mathbf{W} \rVert_\mathrm{F} \to +\infty} \mathcal{L}(\mathbf{W}) = +\infty.
\end{equation*}
\end{itemize}
We have proved that under Assumption~\ref{assumption:pd}, the loss function $\mathcal{L}$ is strictly convex and coercive w.r.t $\mathbf{W}$.

\paragraph{Convergence } As a continuous, strictly convex, and coercive function on the convex $\mathbb{R}^{d \times M}$, $\mathcal{L}$ admits a unique global minimizer. As $\mathcal{L}$ is differentiable, this minimizer is a stationary point of $\mathcal{L}$, i.e., must verify the Euler equation Eq.~\eqref{eq:stationary_point_euler}. Hence, Problem~\eqref{eq:neg_corr_linear_ensemble_with_reg} converges towards the unique stationary point of $\mathcal{L}$.
\end{proof}

\subsection{Proof of Theorem~\ref{thm:diversity_characterization}}
\label{app:diversity_characterization}
The proof of Theorem~\ref{thm:diversity_characterization} is detailed below.
\begin{proof}
    Let $\tilde{\mathbf{W}}$ be a stationary point of $\mathcal{L}$, i.e., $\tilde{\mathbf{W}}$ is solution of Eq.~\eqref{eq:stationary_point_euler}. From Appendix~\ref{app:stationary_point_euler}, the columns $\tilde{\boldsymbol{\omega}}_m$ of $\tilde{\mathbf{W}}$, verify Eq.~\eqref{eq:sol_P_linear_problem_columns} and we have for all $m \in \llbracket 1, M \rrbracket$:
    \begin{align*}
        & \left( \lambda_m \mathbf{I}_d + \frac{\mathbf{X}_\ell^\top \mathbf{X}_\ell}{n_\ell} \right) \tilde{\boldsymbol{\omega}}_m = \frac{\mathbf{X}_\ell^\top \mathbf{y}_\ell}{n_\ell} - \frac{\gamma}{n_u(M-1)} \mathbf{X}_u^\top\mathbf{X}_u \sum_{k=1 | k \neq m}^M \tilde{\boldsymbol{\omega}}_k \\
        \iff & \frac{\gamma}{n_u(M-1)} \mathbf{X}_u^\top\mathbf{X}_u \sum_{k=1 | k \neq m}^M \tilde{\boldsymbol{\omega}}_k = \frac{\mathbf{X}_\ell^\top \mathbf{y}_\ell}{n_\ell} - \left( \lambda_m \mathbf{I}_d + \frac{\mathbf{X}_\ell^\top \mathbf{X}_\ell}{n_\ell} \right) \tilde{\boldsymbol{\omega}}_m \\
        \iff & \frac{\gamma}{n_uM(M-1)} \mathbf{X}_u^\top\mathbf{X}_u \sum_{k=1 | k \neq m}^M \tilde{\boldsymbol{\omega}}_k = \frac{1}{M} \left[ \frac{\mathbf{X}_\ell^\top \mathbf{y}_\ell}{n_\ell} - \left( \lambda_m \mathbf{I}_d + \frac{\mathbf{X}_\ell^\top \mathbf{X}_\ell}{n_\ell} \right) \tilde{\boldsymbol{\omega}}_m \right].
    \end{align*}
By taking the left inner product with $\tilde{\boldsymbol{\omega}}_m$ and summing over all $m \in \llbracket 1, M \rrbracket$, we obtain:
\begin{equation*}
\label{eq:div_emp_loss_intermediary}
    \underbrace{\frac{\gamma}{n_uM(M-1)} \sum_{m=1}^M \tilde{\boldsymbol{\omega}}_m \mathbf{X}_u^\top\mathbf{X}_u \sum_{k=1 | k \neq m}^M \tilde{\boldsymbol{\omega}}_k}_{\mathrm{(LHS)}} = \underbrace{\frac{1}{M} \sum_{m=1}^M \tilde{\boldsymbol{\omega}}_m^\top \left[ \frac{\mathbf{X}_\ell^\top \mathbf{y}_\ell}{n_\ell} - \left( \lambda_m \mathbf{I}_d + \frac{\mathbf{X}_\ell^\top \mathbf{X}_\ell}{n_\ell} \right) \tilde{\boldsymbol{\omega}}_m \right]}_{\mathrm{(RHS)}}.
\end{equation*}
The left-hand side (LHS) term can be rewritten as
\begin{align*}
    \mathrm{(LHS)} &= \frac{\gamma}{n_uM(M-1)} \sum_{m=1}^M \tilde{\boldsymbol{\omega}}_m \mathbf{X}_u^\top\mathbf{X}_u \sum_{k=1 | k \neq m}^M \tilde{\boldsymbol{\omega}}_k \\
    &= \frac{1}{n_uM(M-1)} \sum_{m=1}^M \sum_{k=1 | k \neq m}^M \tilde{\boldsymbol{\omega}}_m \mathbf{X}_u^\top\mathbf{X}_u  \tilde{\boldsymbol{\omega}}_k \\
    &= \frac{\gamma}{n_uM(M-1)} \sum_{m \neq k} \tilde{\boldsymbol{\omega}}_m \mathbf{X}_u^\top\mathbf{X}_u  \tilde{\boldsymbol{\omega}}_k \\
    &= \frac{\gamma}{n_uM(M-1)} \sum_{m \neq k} \sum_{i=n_\ell+1}^{n_\ell +n_u} w_m^\top\mathbf{x}_i w_k^\top\mathbf{x}_i \\
    &= - \gamma \ell_{\mathrm{div}}(\tilde{\mathbf{W}}, \mathbf{X}_u),
\end{align*}
where $\ell_{\mathrm{div}}$ is the diversity introduced in Eq.~\eqref{def:diversity_ensemble}. Moreover, using the decomposition of the Euclidean norm, we have:
\begin{align*}
    & \lVert \mathbf{y}_\ell - \mathbf{X}_\ell \tilde{\boldsymbol{\omega}}_m \rVert_2^2 = \mathbf{y}_\ell^\top \mathbf{y}_\ell -2 \mathbf{y}_\ell^\top \mathbf{X}_\ell \tilde{\boldsymbol{\omega}}_m + \tilde{\boldsymbol{\omega}}_m ^\top \mathbf{X}_\ell^\top \mathbf{X}_\ell \tilde{\boldsymbol{\omega}}_m \\
    \iff & \mathbf{y}_\ell^\top \mathbf{X}_\ell \tilde{\boldsymbol{\omega}}_m = \frac{1}{2} \left[ \mathbf{y}_\ell^\top \mathbf{y}_\ell + \tilde{\boldsymbol{\omega}}_m ^\top \mathbf{X}_\ell^\top \mathbf{X}_\ell \tilde{\boldsymbol{\omega}}_m -  \lVert \mathbf{y}_\ell - \mathbf{X}_\ell \tilde{\boldsymbol{\omega}}_m \rVert_2^2\right].
\end{align*}
As we are in binary classification, we know that each entry of $\mathbf{y}_\ell$ is in $\{-1,+1\}$. Thus, we have:
\begin{equation*}
    \mathbf{y}_\ell^\top \mathbf{y}_\ell = \Vert \mathbf{y}_\ell \rVert _2^2 = \sum_{i=1}^{n_\ell} \lvert y_{\ell, i} \rvert^2 = n_\ell.
 \end{equation*}
We inject the new formula for $\mathbf{y}_\ell^\top \mathbf{X}_\ell \tilde{\boldsymbol{\omega}}_m$ in the right-hand side (RHS) part of Eq.~\eqref{eq:div_emp_loss_intermediary} and obtain:
\begin{align*}
   \mathrm{(RHS)} &= \frac{1}{M} \sum_{m=1}^M \tilde{\boldsymbol{\omega}}_m^\top \left[ \frac{\mathbf{X}_\ell^\top \mathbf{y}_\ell}{n_\ell} - \left( \lambda_m \mathbf{I}_d + \frac{\mathbf{X}_\ell^\top \mathbf{X}_\ell}{n_\ell} \right) \tilde{\boldsymbol{\omega}}_m \right] \\
   &= \frac{1}{M} \left[ \frac{1}{n_l}\sum_{m=1}^M \mathbf{y}_\ell^\top \mathbf{X}_\ell \tilde{\boldsymbol{\omega}}_m - \sum_{m=1}^M \tilde{\boldsymbol{\omega}}_m^\top \left( \lambda_m \mathbf{I}_d + \frac{\mathbf{X}_\ell^\top \mathbf{X}_\ell}{n_\ell} \right) \tilde{\boldsymbol{\omega}}_m  \right] \\
    &= \frac{1}{M} \left( \frac{1}{2n_l}\sum_{m=1}^M \left[n_\ell + \tilde{\boldsymbol{\omega}}_m ^\top \mathbf{X}_\ell^\top \mathbf{X}_\ell \tilde{\boldsymbol{\omega}}_m -  \lVert \mathbf{y}_\ell - \mathbf{X}_\ell \tilde{\boldsymbol{\omega}}_m \rVert_2^2\right] \right) \\
    &\quad - \frac{1}{M} \sum_{m=1}^M \tilde{\boldsymbol{\omega}}_m^\top \left( \lambda_m \mathbf{I}_d + \frac{\mathbf{X}_\ell^\top \mathbf{X}_\ell}{n_\ell} \right) \tilde{\boldsymbol{\omega}}_m \\
    &= \frac{1}{M} \left(\frac{M}{2} - \frac{1}{2n_\ell} \sum_{m=1}^M \lVert \mathbf{y}_\ell - \mathbf{X}_\ell \tilde{\boldsymbol{\omega}}_m \rVert_2^2 \right) \\
    & \qquad + \frac{1}{M} \left(\sum_{m=1}^M \tilde{\boldsymbol{\omega}}_m^\top \frac{\mathbf{X}_\ell^\top \mathbf{X}_\ell}{2n_\ell} \tilde{\boldsymbol{\omega}}_m - \tilde{\boldsymbol{\omega}}_m^\top \left( \lambda_m \mathbf{I}_d + \frac{\mathbf{X}_\ell^\top \mathbf{X}_\ell}{n_\ell} \right) \tilde{\boldsymbol{\omega}}_m \right) \\
    &= \frac{1}{2} \left[1 - \frac{1}{n_\ell M} \sum_{m=1}^M \lVert \mathbf{y}_\ell - \mathbf{X}_\ell \tilde{\boldsymbol{\omega}}_m \rVert_2^2  - \frac{2}{M} \sum_{m=1}^M \tilde{\boldsymbol{\omega}}_m^\top \left( \lambda_m \mathbf{I}_d + \frac{\mathbf{X}_\ell^\top \mathbf{X}_\ell}{2n_\ell} \right) \tilde{\boldsymbol{\omega}}_m \right]\\
    &= \frac{1}{2} \left[1 - \frac{1}{n_\ell M} \sum_{m=1}^M \lVert \mathbf{y}_\ell - \mathbf{X}_\ell \tilde{\boldsymbol{\omega}}_m \rVert_2^2  - \frac{1}{M}\sum_{m=1}^M \tilde{\boldsymbol{\omega}}_m^\top \left( 2 \lambda_m \mathbf{I}_d + \frac{\mathbf{X}_\ell^\top \mathbf{X}_\ell}{n_\ell} \right) \tilde{\boldsymbol{\omega}}_m \right]\\
    &= \frac{1}{2} - \frac{1}{2n_\ell M} \sum_{m=1}^M \lVert \mathbf{y}_\ell - \mathbf{X}_\ell \tilde{\boldsymbol{\omega}}_m \rVert_2^2  - \frac{1}{2M} \sum_{m=1}^M \lambda_m \tilde{\boldsymbol{\omega}}_m^\top \tilde{\boldsymbol{\omega}}_m \\
    & \qquad - \frac{1}{2M} \sum_{m=1}^M \tilde{\boldsymbol{\omega}}_m^\top \left( \lambda_m \mathbf{I}_d + \frac{\mathbf{X}_\ell^\top \mathbf{X}_\ell}{n_\ell} \right) \tilde{\boldsymbol{\omega}}_m \\
    &= \frac{1}{2} - \frac{1}{2n_\ell M} \sum_{m=1}^M \lVert \mathbf{y}_\ell - \mathbf{X}_\ell \tilde{\boldsymbol{\omega}}_m \rVert_2^2  - \frac{1}{2M} \sum_{m=1}^M \lambda_m \lVert \tilde{\boldsymbol{\omega}}_m \rVert_2^2 \\
    & \qquad - \frac{1}{2M} \sum_{m=1}^M \tilde{\boldsymbol{\omega}}_m^\top \left( \lambda_m \mathbf{I}_d + \frac{\mathbf{X}_\ell^\top \mathbf{X}_\ell}{n_\ell} \right) \tilde{\boldsymbol{\omega}}_m.
\end{align*}
We finally obtain from $\mathrm{(LHS)} = \mathrm{(RHS)}$, after multiplying by $-1$, that
\begin{equation*}
    \gamma \ell_{\mathrm{div}}(\tilde{\mathbf{W}}, \mathbf{X}_u) = \frac{1}{2n_\ell M} \sum_{m=1}^M \lVert \mathbf{y}_\ell - \mathbf{X}_\ell \tilde{\boldsymbol{\omega}}_m \rVert_2^2  + \frac{1}{2M} \sum_{m=1}^M \tilde{\boldsymbol{\omega}}_m^\top \left(\lambda_m \mathbf{I}_d + \frac{\mathbf{X}_\ell^\top \mathbf{X}_\ell}{n_\ell} \right) \tilde{\boldsymbol{\omega}}_m + \frac{1}{2M} \sum_{m=1}^M \lambda_m \lVert \tilde{\boldsymbol{\omega}}_m \rVert_2^2 - \frac{1}{2}.
\end{equation*}

Under the hypothesis that $\frac{1}{M} \sum_{m=1}^M \lambda_{m} \lVert \tilde{\boldsymbol{\omega}_{m}} \rVert _2^2 \geq 1$, we have:
\begin{equation*}
\frac{1}{2M} \sum_{m=1}^M \lambda_m \lVert \tilde{\boldsymbol{\omega}}_m \rVert_2^2 - \frac{1}{2} = \frac{1}{2} \left[ \frac{1}{M} \sum_{m=1}^M \lambda_m \lVert \tilde{\boldsymbol{\omega}}_m \rVert _2^2 -1 \right] \geq 0.
\end{equation*} 
Taking this fact into account, we obtain:
\begin{align*}
    \gamma \ell_{\mathrm{div}}(\tilde{\mathbf{W}}, \mathbf{X}_u) &= \frac{1}{2n_\ell M} \sum_{m=1}^M \lVert \mathbf{y}_\ell - \mathbf{X}_\ell \tilde{\boldsymbol{\omega}}_m \rVert_2^2  + \frac{1}{2M} \sum_{m=1}^M \tilde{\boldsymbol{\omega}}_m^\top \left(\lambda_m \mathbf{I}_d + \frac{\mathbf{X}_\ell^\top \mathbf{X}_\ell}{n_\ell} \right) \tilde{\boldsymbol{\omega}}_m + \underbrace{\frac{1}{2M} \sum_{m=1}^M \lambda_m \lVert \tilde{\boldsymbol{\omega}}_m \rVert_2^2 - \frac{1}{2}}_{\geq 0} \\
    &\geq \frac{1}{2n_\ell M} \sum_{m=1}^M \lVert \mathbf{y}_\ell - \mathbf{X}_\ell \tilde{\boldsymbol{\omega}}_m \rVert_2^2  + \frac{1}{2M} \sum_{m=1}^M \tilde{\boldsymbol{\omega}}_m^\top \left(\lambda_m \mathbf{I}_d + \frac{\mathbf{X}_\ell^\top \mathbf{X}_\ell}{n_\ell} \right) \tilde{\boldsymbol{\omega}}_m.
\end{align*}
\end{proof}

\subsection{Proof of Corollary~\ref{cor:contrastive_learning}}
\label{app:contrastive_learning}
The proof of Corollary~\ref{cor:contrastive_learning} is detailed below.
\begin{proof}
Assuming $\frac{1}{M} \sum_{m=1}^M \lambda_m \lVert \tilde{\boldsymbol{\omega}}_m \rVert_2^2 \geq 1$, Theorem~\ref{thm:diversity_characterization} holds. Using the fact that the $\lambda_m$ are all equal to some $\lambda$, it leads to:
\begin{align*}
\gamma \ell_{\mathrm{div}}(\tilde{\mathbf{W}}, \mathbf{X}_u) &\geq \underbrace{\frac{1}{2n_\ell M} \sum_{m=1}^M \lVert \mathbf{y}_\ell - \mathbf{X}_\ell \tilde{\boldsymbol{\omega}}_m \rVert_2^2}_{\geq 0}  + \frac{1}{2M} \sum_{m=1}^M \tilde{\boldsymbol{\omega}}_m^\top \left(\lambda \mathbf{I}_d + \frac{\mathbf{X}_\ell^\top \mathbf{X}_\ell}{n_\ell} \right) \tilde{\boldsymbol{\omega}}_m \\
&\geq \frac{1}{2M} \left( \lambda \sum_{m=1}^M \lVert \tilde{\boldsymbol{\omega}}_m \rVert _2^2 + \frac{1}{n_\ell}\sum_{m=1}^M \tilde{\boldsymbol{\omega}}_m^\top \mathbf{X}_\ell^\top \mathbf{X}_\ell \tilde{\boldsymbol{\omega}}_m\right)\\
&\geq \frac{1}{2M} \left( \lambda + \frac{1}{n_\ell}\lambda_{min}\left(\mathbf{X}_\ell^\top \mathbf{X}_\ell\right)\right) \sum_{m=1}^M \lVert \tilde{\boldsymbol{\omega}}_m \rVert _2^2 \tag{from Lemma~\ref{lemma:eigenvalues_psd}} \\
&= \frac{1}{2M} \left( \lambda + \frac{1}{n_\ell}\lambda_{min}\left(\mathbf{X}_\ell^\top \mathbf{X}_\ell\right)\right) \lVert \tilde{\mathbf{W}}\rVert^2_\mathrm{F} \tag{from Eq.~\eqref{eq:frobenius_reformulation}}.
\end{align*}
\end{proof}

\vfill

\end{document}